\documentclass{article}

\usepackage{microtype}
\usepackage{graphicx}
\usepackage{subfigure}
\usepackage{booktabs} 

\usepackage{hyperref}
\usepackage{enumitem}
\usepackage{multirow}
\usepackage{bm}

\usepackage[accepted]{icml2023}

\usepackage{amsmath}
\usepackage{amssymb}
\usepackage{mathtools}
\usepackage{amsthm}

\usepackage[capitalize,noabbrev]{cleveref}

\theoremstyle{plain}

\theoremstyle{definition}

\theoremstyle{remark}

\usepackage[textsize=tiny]{todonotes}

\newcommand{\dssectionheader}[1]{%
   \noindent\framebox[\columnwidth]{%
      {\fontfamily{phv}\selectfont \textbf{\textcolor{blue}{#1}}}
   }
}

\newcommand{\dsquestion}[1]{%
   {\noindent \scriptsize {\fontfamily{phv}\selectfont \textcolor{blue}{\textbf{#1}}}}
}

\newcommand{\dsquestionex}[2]{%
   {\noindent \scriptsize {\fontfamily{phv}\selectfont \textcolor{blue}{\textbf{#1} #2}}}
}

\newcommand{\dsanswer}[1]{%
   {\noindent \footnotesize {#1} \medskip}
}

\icmltitlerunning{MABe22: Multi-Species Multi-Task Benchmark for Learned Representations of Behavior}

\begin{document}

\twocolumn[
\icmltitle{MABe22: A Multi-Species Multi-Task Benchmark \\ for Learned Representations of Behavior}



\icmlsetsymbol{equal}{*}

\begin{icmlauthorlist}
\icmlauthor{Jennifer J. Sun}{equal,yyy}
\icmlauthor{Markus Marks}{equal,yyy}
\icmlauthor{Andrew W. Ulmer}{comp}
\icmlauthor{Dipam Chakraborty}{sch}
\icmlauthor{Brian Geuther}{jax}
\icmlauthor{Edward Hayes}{} 
\icmlauthor{Heng Jia}{zhe}
\icmlauthor{Vivek Kumar}{jax}
\icmlauthor{Sebastian Oleszko}{irl}
\icmlauthor{Zachary Partridge}{unsw}
\icmlauthor{Milan Peelman}{gent}
\icmlauthor{Alice Robie}{jan}
\icmlauthor{Catherine E. Schretter}{jan}
\icmlauthor{Keith Sheppard}{jax}
\icmlauthor{Chao Sun}{zhe}
\icmlauthor{Param Uttarwar}{saa}
\icmlauthor{Julian M. Wagner}{yyy}
\icmlauthor{Erik Werner}{irl}
\icmlauthor{Joseph Parker}{yyy}
\icmlauthor{Pietro Perona}{yyy}
\icmlauthor{Yisong Yue}{yyy}
\icmlauthor{Kristin Branson}{jan}
\icmlauthor{Ann Kennedy}{comp}
\icmlauthor{{\small Website: \url{https://sites.google.com/view/computational-behavior/our-datasets/mabe2022-dataset}} }{}
\end{icmlauthorlist}

\icmlaffiliation{yyy}{Caltech}
\icmlaffiliation{comp}{Northwestern University}
\icmlaffiliation{sch}{AICrowd}
\icmlaffiliation{jax}{JAX Labs}
\icmlaffiliation{jan}{Janelia}
\icmlaffiliation{zhe}{Zhejiang University}
\icmlaffiliation{irl}{IRLAB Therapeutics}
\icmlaffiliation{unsw}{University of New South Wales
}
\icmlaffiliation{gent}{Ghent University
}
\icmlaffiliation{saa}{Saarland University}

\icmlcorrespondingauthor{Ann Kennedy}{ann.kennedy@northwestern.edu}

\icmlkeywords{Machine Learning, ICML, Representation learning, datasets and benchmarks, animal behavior, behavior analysis}

\vskip 0.3in
]



\printAffiliationsAndNotice{\icmlEqualContribution} 

\begin{abstract}
We introduce \textit{MABe22}, a large-scale, multi-agent video and trajectory benchmark to assess the quality of learned behavior representations. 
This dataset is collected from a variety of biology experiments, and includes triplets of interacting mice (4.7 million frames video+pose tracking data, 10 million frames pose only), symbiotic beetle-ant interactions (10 million frames video data), and groups of interacting flies (4.4 million frames of pose tracking data).
Accompanying these data, we introduce a panel of real-life downstream analysis tasks to assess the quality of learned representations by evaluating how well they preserve information about the experimental conditions (e.g. strain, time of day, optogenetic stimulation) and animal behavior.
We test multiple state-of-the-art self-supervised video and trajectory representation learning methods to demonstrate the use of our benchmark, revealing that methods developed using human action datasets do not fully translate to animal datasets.  
We hope that our benchmark and dataset encourage a broader exploration of behavior representation learning methods across species and settings.
\end{abstract}





\vspace{-0.1in}
\section{Introduction}
\label{sec:intro}

\begin{figure}[t!]
    \centering
    \includegraphics[width=0.95\linewidth]{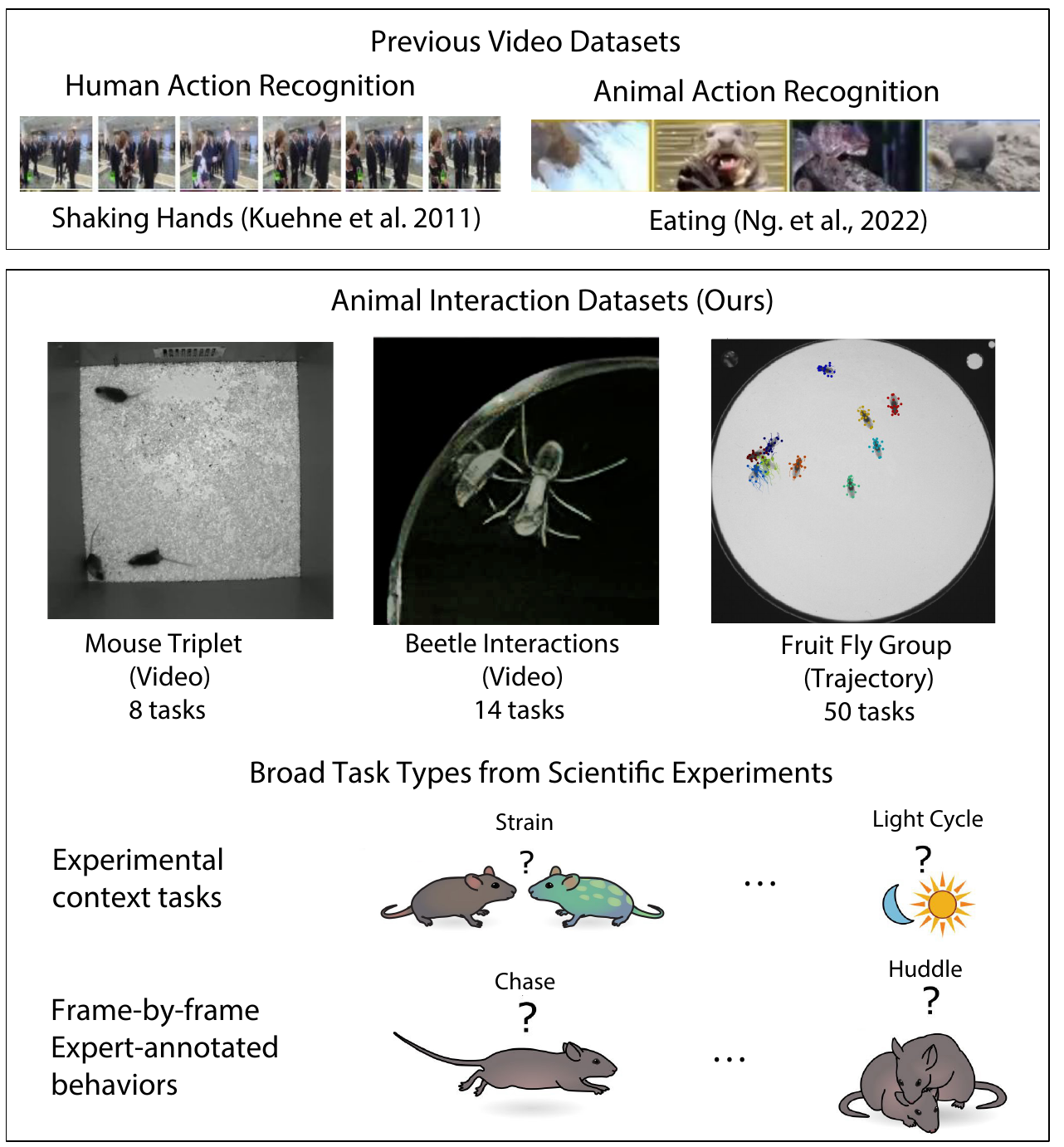}
        \caption{\textbf{MABe22 consists of animal interactions in laboratory experiments}. We propose a dataset to benchmark representation learning methods that focus on multi-agent behavior. 
        Our benchmark includes a large video and trajectory library depicting interactions of mice, beetles, ants, and fruit flies alongside a large suite of downstream tasks to measure representation quality. Tasks differ across model organisms and include the classification of experimental conditions (e.g. species strain, light cycle, optogenetic activations, interaction duration) as well as expert-annotated actions (e.g. chase, huddle, and sniffs for mice).
        \vspace{-0.1in}
    }
    \label{fig:intro}
\end{figure}

The study of interacting agents is important for a range of scientific and engineering applications, from designing safer autonomous vehicles~\cite{chang2019argoverse}, to understanding player behavior in virtual worlds~\cite{hofmann2019minecraft}, to uncovering the biological underpinnings of neurological disorders~\cite{segalin2020mouse,wiltschko2020revealing}.
Across disciplines, there is a need for new techniques to characterize the structure of multi-agent behavior with greater precision, sensitivity, and detail. 
Traditionally, behavior analysis models are trained with full supervision~\cite{burgos2012social,hong2015automated,bohnslav2021deepethogram}, which subjects users to a heavy burden of video annotation.
Efforts to learn behavioral representations without manual annotation~\cite{berman2014mapping,wiltschko2015mapping,hsu2020b,sun2020task} promise to bypass this labor bottleneck, but are difficult to evaluate systematically.
To support the development of learned behavioral representations, and to better evaluate their performance, we need benchmark datasets for behavior.
These benchmarks should cover a broad range of experimental conditions, to avoid overfitting on the statistics of a particular dataset.
Furthermore, when representations are learned without supervision, there is no obvious metric to evaluate the quality of the representation. 
Yet, a metric is needed for quantitative comparisons. 
These two challenges inspired our work.

We have collected and curated a large dataset and benchmark from biology experiments for evaluating learned representations of social behavior (Figure~\ref{fig:intro}).  We chose to focus on videos of laboratory animals for several reasons:
\begin{itemize}[noitemsep,topsep=0pt,leftmargin=*]
    \item Animal behavioral experiments are collected against a uniform uninformative background, such as~\cite{segalin2020mouse,eyjolfsdottir2014detecting,pereira2020quantifying}, and thus behavior classifiers are forced to focus on the dynamic and pictorial cues of the action. In contrast, video of human behavior, e.g., actions in different sports, are usually {\em pictorially informative}, meaning that the action itself can be classified from the appearance of a single or a few frames rather than considering motion over long periods of time.
    \item Animal behavior is often recorded under various experimental manipulations that impact the behavior (Figure~\ref{fig:intro}). Identifying those experimental manipulations provides an objective task that may be used to evaluate the quality of a representation. This complements evaluation based on reproducing human annotations of behavior, which have shorter temporal structure but can be subjective~\cite{anderson2014toward}.
    \item The biologists who provided us with videos of their experiments are engaged in analyzing specific aspects of the animals' behavior. Using a given representation to automate their analysis provides us with an objective performance criterion that is defined outside the field of Computer Vision. Evaluation methods based on {\em downstream tasks}, i.e. tasks where the representation is used to analyze specific aspects of the signal, have been used in other domains, e.g. for evaluating visual representations~\cite{van2021benchmarking} or neural mechanistic models~\cite{Schrimpf2020integrative}.
    \item Our dataset is from real-world neuroscience and evolutionary biology experiments, and progress on this dataset will enable biologists to use the representations generated to study how behavior changes as a function of other experimental variables. 
\end{itemize}

We make three contributions: {\bf 1.} A large and richly annotated video and trajectory dataset, \textbf{M}ulti-\textbf{A}gent \textbf{Be}havior 20\textbf{22} (MABe22), of social behavior in three species: laboratory mice ({\em Mus musculus}) triplets, rove beetles ({\em Sceptobius lativentris}) paired with their symbiotic host species or with other beetles, and vinegar flies ({\em Drosophila melanogaster}).  {\bf 2.} A large and diverse set of downstream evaluation tasks based on the classification of experimental conditions (optogenetic activation, animal strain, time-of-day) and expert-annotated behavior labels.
{\bf 3.} A baseline benchmark of state-of-the-art self-supervised video and trajectory representation learning, as well as community-contributed methods solicited from an \href{https://www.aicrowd.com/challenges/multi-agent-behavior-challenge-2022}{open challenge}.
To the best of our knowledge, our dataset is the first to provide non-annotation-based downstream tasks from scientific experiments for representation evaluation (Table~\ref{tab:related_work}).

Our dataset and related code is available at:\\
\href{https://sites.google.com/view/computational-behavior/our-datasets/mabe2022-dataset}{https://sites.google.com/view/computational-behavior/our-datasets/mabe2022-dataset}.


\begin{table*}
    \begin{center}
        \scalebox{0.9}{
            \begin{tabular}{lccccc}
                \toprule[0.2em]
                \multirow{2}{*}{Dataset}                             &    Number of    &       Annotation     &       Action      &       Downstream    &      \multirow{ 2}{*}{ Size}        \\
                & species & frequency & classes & tasks & \\
                \toprule[0.2em]
                Kinetics400~\cite{kay2017kinetics}      &       1 (human)               &               clip           &               400         &               x           &    306k clips    \\
                \hline
                HMDB~\cite{kuehne2011hmdb}              &       1 (human)               &               clip           &               51         &               x           &    6776 clips     \\
                \hline                
                UCF~\cite{soomro2012ucf101}             &       1 (human)               &               clip           &               101         &               x           &    13320 clips     \\
                \hline
                \hline
                Animal Kingdom~\cite{ng2022animal}                          &           850             &               frame           &               140         &               x           &   4.5M frames    \\
                \hline                
                \multirow{ 2}{*}{CalMS21~\cite{sun2021multi}}     &  \multirow{ 2}{*}{ 1}   &   \multirow{ 2}{*}{frame}  &  \multirow{ 2}{*}{7 }  &   \multirow{ 2}{*}{x }  &   1M frames   \\
                & & & & & +6M unlabelled \\
                \hline
                
                Fly vs. Fly~\cite{eyjolfsdottir2014detecting}   &   1   &   frame    &    10  &  x  &   1.5M frames \\ 
                \hline
                
                CRIM13~\cite{burgos2012social}  & 1 & frame & 13 & x & 8M frames \\ 
                \hline
                \hline
                \multirow{ 2}{*}{Our Dataset}                         &  \multirow{ 2}{*}{4}        &               \multirow{ 2}{*}{frame}        &              \multirow{ 2}{*}{ 16}           &               \textbf{56}          &  15M frames video +        \\    
                & & & & from experiments & 14M frames traj\\   
                \bottomrule[0.1em]
            \end{tabular}
            }
        \caption{
        \textbf{Comparison with commonly used, public video and trajectory datasets}. While existing datasets can be used for behavioral representation learning, the downstream evaluation focuses on a single type of task (detection and classification of human-annotated actions) or a single species. Our benchmark introduces a rich set of downstream analysis tasks that we obtain from scientific experiments on multiple species. 
        \vspace{-0.3in}}~\label{tab:related_work}
    \end{center}
\end{table*}

\section{Related Work}~\label{sec:related}
\vspace{-0.1in}

\textbf{Related Animal Datasets.} The goal of the MABe22 dataset is to benchmark representation learning models for behavior analysis using data from biology experiments. 
There are several existing datasets for studying animal social behavior, including CRIM13~\cite{burgos2012social}, Fly vs. Fly~\cite{eyjolfsdottir2014detecting}, and CalMS21~\cite{sun2021multi}.
These datasets contain video or pose data from interacting animals, as well as human-annotated behavior labels (Table~\ref{tab:related_work}); they all focus on a single species and setting. 
AnimalKingdom~\cite{ng2022animal} is another recent animal behavior dataset that includes social and nonsocial behavior from multiple species, but is focused on human annotation-based action recognition only.
Our dataset is unique in that it defines a range of downstream tasks for each organism; these tasks are motivated by scientific experiments, with the goal of to driving scientific discovery in biology. 

\textbf{Related Human Datasets.} While animal video datasets remain comparatively rate, there are many video datasets designed for work in human action recognition.
Human datasets typically have very different visual characteristics from animal datasets. 
Most notably, many human datasets that are used to benchmark self-supervised video representation learning, such as Kinetics \cite{kay2017kinetics}, UCF101 \cite{soomro2012ucf101} and HMDB51 \cite{kuehne2011hmdb}, contain 'spatially heavy' visual information that informs downstream action classification-- that is, different actions have different backgrounds.
Because of these differences in the visual appearance, agents' actions can be partly distinguished by these visual features alone, without models having to learn any temporal features of the agents' behavior.
In contrast, our animal videos are all acquired against a stationary, neutral background, forcing models to use the temporal structure of the data to distinguish between actions. 

\textbf{Related Problems in Multi-Agent Behavior.}
While our dataset is composed of multi-agent data from biology, there are also multi-agent behavior datasets from other domains, such as from autonomous driving~\cite{chang2019argoverse,sun2020scalability}, sports analytics~\cite{yue2014learning,decroos2018automatic}, and video games~\cite{samvelyan2019starcraft,guss2019minerl}.
These datasets often focus on forecasting, motion planning, and reinforcement learning, whereas our dataset is used for tasks from scientific applications, such as distinguishing animal strains via observed behaviors.

\textbf{Work in Animal Behavior Analysis.} In biology and neuroscience, computational models of behavior have the potential to significantly reduce human data annotation efforts, and to provide more detailed descriptions of the behavior in question~\cite{anderson2014toward,pereira2020quantifying}. 
Automated characterizations of animal behavior have been used to study the relationship between neural activity and behavior~\cite{markowitz2018striatum}, to characterize behavioral differences between species and between different strains within a species~\cite{hernandez2020framework}, and to quantify the effect of functional or pharmacological perturbations~\cite{robie2017mapping,wiltschko2020revealing}.
The input to these models may be video~\cite{bohnslav2021deepethogram} or trajectory data~\cite{sun2020task,segalin2020mouse}.

Supervised behavior models have been trained to identify human-defined behaviors-of-interest~\cite{hong2015automated,segalin2020mouse, marks2022deep,kabra2013jaaba}, often using frame-by-frame behavior annotations from domain experts. 
Another body of work discovers behaviors without human annotations, using unsupervised and self-supervised methods~\cite{berman2014mapping,wiltschko2015mapping,hsu2020b,luxem2020identifying,calhoun2019unsupervised} that learn the latent structure of behavioral data. The learned representation may be continuous~\cite{sun2020task}, or discrete, such as when discovering behavior motifs~\cite{berman2014mapping,wiltschko2015mapping,hsu2020b}. 
There currently does not exist a unified behavioral representation learning dataset that can compare these models across a broad range of behavior analysis settings. Here, we propose MABe 2022 for evaluating the performance of these representation learning methods.

\textbf{Work in Representation Learning.} Representation learning for visual~\cite{gidaris2018unsupervised,chen2020simple,oord2018representation,kolesnikov2019revisiting,han2019video} and trajectory data~\cite{sun2020task,zhan2021unsupervised} has been applied to a variety of tasks, such as for image classification~\cite{chen2020simple}, speech recognition~\cite{oord2018representation}, and behavior classification~\cite{sun2020task}. 
In these works, many different unsupervised / self-supervised methods have been developed, employing various pretext tasks to pre-train a model, such as classifying image rotations~\cite{gidaris2018unsupervised}, predicting future observations~\cite{oord2018representation}, contrastive learning with image augmentations~\cite{chen2020simple}, and decoding programmatic attributes~\cite{sun2020task}. 
The quality of learned representations is often evaluated on downstream tasks. 

\textit{Behavioral Representation Learning.} For behavior analysis, applications of representation learning include discovering behavior motifs~\cite{berman2014mapping, wiltschko2015mapping, hsu2020b, luxem2020identifying}, identifying internal states~\cite{calhoun2019unsupervised}, and improving sample-efficiency of supervised classifiers~\cite{sun2020task}. 
These works use methods such as variational autoencoders~\cite{kingma2013auto}, autoregressive hidden Markov models~\cite{wiltschko2015mapping}, and Uniform Manifold Approximation and Projection (UMAP)~\cite{mcinnes2018umap} to characterize the latent structure of behavior. 
Notably, many groups have proposed methods for unsupervised behavior discovery~\cite{berman2014mapping,klibaite2017unsupervised,wiltschko2015mapping,luxem2020identifying,hsu2020b,marques2018structure}.
These works use different methods to model the temporal structure of behavior, including wavelet transforms~\cite{berman2014mapping}, autoregressive hidden Markov models~\cite{wiltschko2015mapping}, and recurrent NNs~\cite{luxem2020identifying}, as well as different methods for segmenting behavior, such as Gaussian mixture models~\cite{hsu2020b}, k-means clustering~\cite{luxem2020identifying}, and watershed transforms~\cite{berman2014mapping}. 
Our goal is to develop a standardized dataset for evaluating these methods on a common set of behavior analysis tasks.


\section{Dataset Design and Collection}

\begin{figure*}[h]
    \centering
    \includegraphics[width=0.95\linewidth]{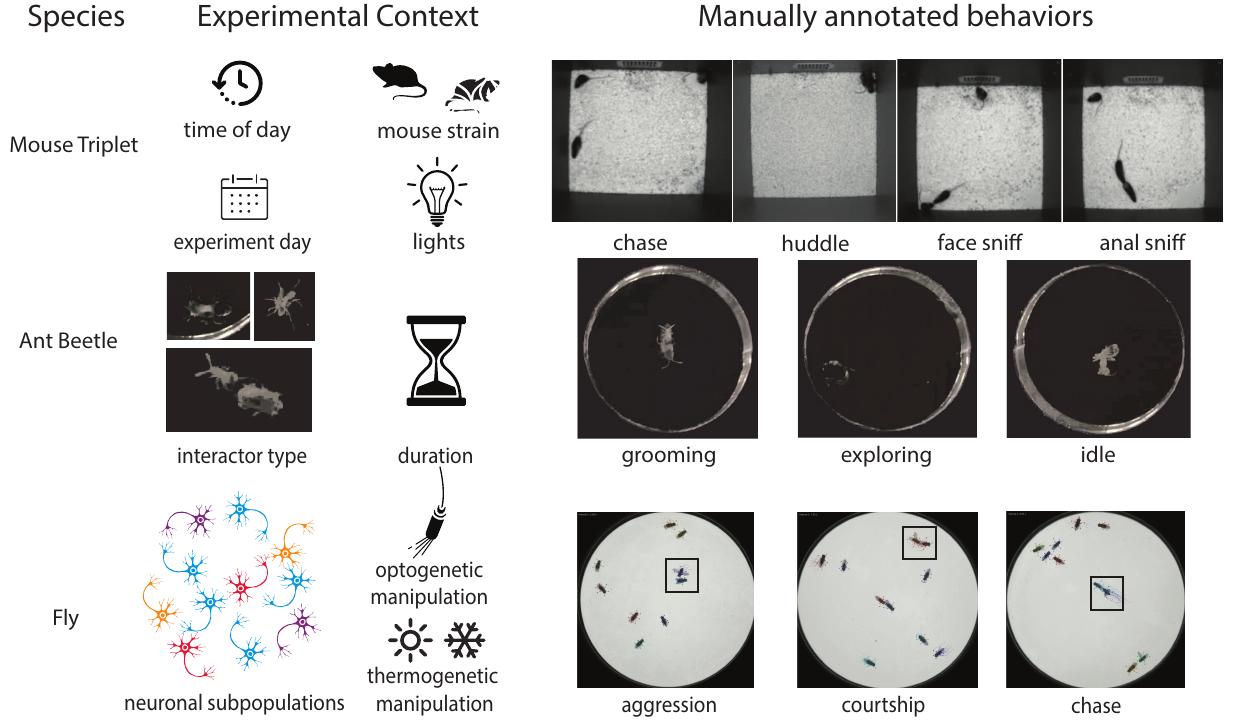}
    \vspace{-0.1in}
    \caption{\textbf{Summary of tasks and actions in our dataset.} Our dataset includes three different species: mice, beetles with an intractor (an ant or other another beetle), and flies. The mouse dataset has both video and trajectory available, the beetle dataset is video-based, and the fly dataset is trajectory based. Classification of experimental conditions is used as a performance metric (examples depicted on the left for each dataset). Additionally, we collected conventionally expert-annotated actions (examples depicted on the right for each dataset), with frame-by-frame labels, e.g., as "chase", "huddle", "face sniff", and "anogenital sniff" for mice. Overall, there are 72 behavior analysis tasks: 8 for mice, 14 for beetles and 50 for flies.
    }
    \label{fig:dataset}
\end{figure*}

We designed and curated MABe22, a multi-agent behavior dataset for the purpose of studying behavioral representation learning.
Our dataset consists of data from multiple model organisms in neuroscience/biology: mice, beetles, and flies.
For each dataset, we constructed a collection of tasks based on real-world scientific applications, including determining the experimental context of the organisms and capturing expert-annotated behaviors.
There are 72 tasks in total: 8 for mice, 14 for beetles, and 50 for flies. 
For the purpose of establishing a benchmark, we define a "good" learned representation of animal behavior that can decode biologically meaningful hidden labels as well as annotations by experts.
Some tasks apply to all frames of the recording (e.g. strain of mice), but not all tasks are apply to all frames (e.g. sniffing, since experts may annotate only a subset of the videos). More details are available in the datasheet for our dataset (Appendix~\ref{appendix:datasheet}).

The mouse dataset (Section~\ref{sec:mice}) consists of 2614 clips of video and trajectory data (1 minute each at 30 Hz) curated from longer videos of a triplet of interacting mice over multiple recording days. 
The video and trajectory datasets are from the same clips, and the mice are tracked using~\cite{sheppard2020gait}.
We additionally release a larger set of 5336 clips of trajectory data for evaluating community-contributed methods (only used in Appendix~\ref{sec:addtrajectory}).
The beetle dataset (Section~\ref{sec:beetle}) consists of 11536 clips of video (30 seconds each at 30 Hz) curated from paired interactions of rove beetles (\textit{Sceptobius lativentris}) with intact or manipulated members of their symbiotic host species, the velvety tree ant (\textit{Liometopum occidentale}), or with other beetle species.
The fly dataset (Section~\ref{sec:fly}) consists of 968 clips of trajectory data (30-second clips at 150 Hz) of groups of 8-11 interacting flies, tracked using~\cite{APT}.

\subsection{Mouse Triplets}~\label{sec:mice}
\textbf{Data Description.} The mouse dataset consists of a set of videos and trajectories from three interacting mice, recorded from an overhead camera in an open field arena measuring 52cm x 52cm, with a grate located at the northern wall of the arena giving access to food and water. Animals were introduced to the arena one by one over the first ten minutes of recording and were recorded continuously for four days at a framerate of 30 Hz and a camera resolution of 800 x 800 pixels. Illumination was provided by an overhead light on a 24-hour reverse light cycle (lights off during the day and on at night); mice are nocturnal and thus are most active during the dark. Behavior was recorded using an IR-pass filter so that light status could not be detected by the eye in the recorded videos. Animals' posture was tracked using a pose estimation model~\cite{sheppard2020gait} based on HRNet~\cite{sun2019deep} with an identity embedding network to track long-term identity.

\textbf{Tasks.} Representations of the mouse dataset are evaluated on 8 tasks that capture information about animals' genetic background, environment, and expert-annotated behaviors. 
These tasks were selected based on their relevance to common scientific applications such as identifying the behavioral effects of differences in animals' genetic backgrounds or experimenter-imposed changes in their environment.
We examined capacity of learned representations to determine animal strain, as well as environmental factors such as whether room lights were on or off (a proxy for day/night cycles, which modulate animal behavior). We also included two tasks to predict the day of the trajectory relative to the start of recording (animal behavior changes across days as they habituate to a new environment~\cite{klibaite2022deep}), and the time of day of the trajectory (animal behavior changes over the course of a day, driven by circadian rhythms).
A learned representation of behavior should also be rich enough to recapitulate human-produced labels of animals' moment-to-moment actions. Therefore our evaluation tasks include the detection of expert-annotated behaviors: huddling, chasing, face sniffing, and anogenital sniffing.
A detailed description of the tasks is listed in Appendix~\ref{appendix:mouse_task}.

\subsection{Beetle Interactions}~\label{sec:beetle}
 \textbf{Dataset description.}  The beetle dataset consists of a rove beetle (\textit{Sceptobius lativentris}) interacting one-on-one with its host ant (\textit{Liometopum occidentale}), manipulated host ant (e.g., with pheromones stripped off) or with other insects (e.g., a nitidulid beetle). The original experiment consisted of two-hour interaction trials, from which we extracted a collection of 30-second clips. These recordings were made in 8-well behavioral interaction chambers (2cm diameter circles) in the dark and illuminated with inferred lights from the side/top. A top-mounted machine vision camera sensitive to IR light monitored the two-hour behavioral trials at 60 Hz. For this dataset, individual circular wells were cropped/parsed from the multi-well video and saved at 800x800 resolution with downsampling to 30 Hz.

\textbf{Tasks.} The beetle dataset includes tasks based on environmental conditions as well as expert-annotated behaviors. Labels for environmental conditions include the interactor type (the species of insect the rove beetle interacts with, and any experimental manipulations applied) as well as how long into the two-hour assay the observed clip occurred. The interactors represent a range of cue types, from the host organism with which the beetle should interact extensively to other insects that the beetle will likely ignore. We also provide expert annotations for six behaviors across the seven different types of one-on-one interactions. 
Generating a meaningful representation that extracts information of interest about the different behaviors adopted by the beetle in response to these disparate cues is crucial for insight into how species interact in nature. Details about the interaction tasks are described in Appendix~\ref{appendix:beetle_interactors}.

\subsection{Fly Groups}~\label{sec:fly}
\textbf{Data Description.} The fly dataset consists of trajectories of groups of 8 to 11 vinegar flies ({\it Drosophila melanogaster}) interacting in a 5cm-diameter dish. The trajectories were derived from 96 videos of length 50k-75k frames, collected at 1024x1024 pixels and 150 frames per second. The flies' bodies and wings were tracked using FlyTracker~\cite{eyjolfsdottir2014detecting}, and landmarks on the body were tracked using the Animal Part Tracker (APT)~\cite{APT} producing a total of 19 keypoints per tracked animal (details in Appendix~\ref{appendix:fly_tracking})

As the brain controls behavior, a good representation of behavior should change with neural activity. Thanks to its tractable genetics, precise neural activity manipulations are straightforward in \textit{Drosophila}. We thus chose to perform experiments using optogenetic (light-activated neural activity via Chrimson) \cite{klapoetke2014independent} and thermogenetic (heat activated, via TrpA) \cite{robie2017mapping} activation of selected sets of neurons. We chose neurons (and the associated GAL4 lines) previously identified as controlling social behaviors, including courtship, avoidance \cite{robie2017mapping}, and female aggression~\cite{schretter2020cell}. For thermogenetic experiments, neural activation is constant and continuous for the entire video. Our optogenetic experiments consisted of activation for short periods of time at weak and strong intensities interspersed with periods of no activation. We combined these neural manipulations with genetic mutations and rearing conditions. Specifically, we selected populations of flies with the norpA mutation, which induces blindness~\cite{bloomquist1988isolation}, and either raised groups of flies together or separated by sex.

\textbf{Tasks.} The representations of the fly dataset are evaluated on a set of 50 tasks. Many of these tasks differentiate which populations of neurons are activated and how they are activated. For example, Task 5 indicates the activation of courtship neurons targeted by the R71G01 GAL4 line in groups of 5 male and 5 female flies. Task 31 compares how neurons were activated -- it compares strong and weak activation of aIPg neurons, which regulate female aggression. Besides neural activation, tasks also differentiate flies based on sex, how the flies were raised, which strain they are from, and genetic mutations. A full list of tasks and the types of flies used are in Appendix~\ref{appendix:fly_tracking}. 

Besides biological differences, we also include tasks based on manual annotations of the flies' behavior for the following social behaviors: any aggressive behavior toward another fly, chasing another fly, any courtship behavior toward another fly, high fencing, wing extension, and wing flick. We annotated behaviors sparsely across all videos with human experts using JAABA~\cite{kabra2013jaaba}, with the goal of including annotations in a wide variety of flies and videos.


\section{Benchmarking \& Methods}

\begin{table*}[h]
  \begin{center}
\scalebox{0.9}{
    \small
    \begin{tabular}{l|cccccc}
        \toprule[0.2em]
        \multirow{2}{*}{Mouse Triplets} &
         Exp.   &   Time of     &   Strain      &       Lights      &   Manual     \\
                &           Day $\downarrow$ &       Day $\downarrow$     &        $\uparrow$           &       $\uparrow$        & Behaviors$\uparrow$ \\        
        \toprule[0.2em]
        $\rho$BYOL (R-50 (Slow Pathway) 8x8~\cite{feichtenhofer2021large} & .0152 & .0913 & .9997 & .9701 & 0.1832 \\
        Maskfeat (MViTv2-S 16x4)~\cite{wei2022masked} & .0393 & .0948 & .9925 & .7309 & 0.1627 \\
        MAE (ViT-B 16x4)~\cite{feichtenhofer2022masked} & \textbf{.0102} & \textbf{.0816} & \textbf{1.0000} & \textbf{.9758} & 0.2309 \\
     \bottomrule[0.1em]
        (pretrained) $\rho$BYOL (-, R-50 (Slow Pathway) 8x8 & .0176 & .0910 & .9994 & .7967 & \textbf{0.2688} \\
        (pretrained) Maskfeat (-, MViTv2-S 16x4)  & .0456 & .0889 & .9998 & .7892 & 0.1896 \\
        (pretrained) MAE (-, ViT-B 16x4)  & .0218 & .0925 & \textbf{1.0000} & .9391 & 0.2301 \\
        \bottomrule[0.1em]
        \multirow{2}{*}{Ant Beetle} &
         Duration  & Interactor  & Manual   & Manual  \\
         & $\downarrow$  & Type $\uparrow$ & Behaviors $\uparrow$ & Behaviors (same) $\uparrow$\\       
        \toprule[0.2em]
        $\rho$BYOL (50 (Slow Pathway) 8x8~\cite{feichtenhofer2021large} & \textbf{.0257} & .9999 & .6178 & .6457 \\
        Maskfeat (MViTv2-S 16x4)~\cite{wei2022masked}  & .0291 & \textbf{1.0000} & .6212 & .6574 \\
        MAE (ViT-B 16x4)~\cite{feichtenhofer2022masked}  & .0283 & \textbf{1.0000} & .6444 & .6874 \\
     \bottomrule[0.1em]
        (pretrained) $\rho$BYOL (R-50 (Slow Pathway) 8x8 & .0300 & .9981 & \textbf{.6967} & \textbf{.7334} \\
        (pretrained) Maskfeat (MViTv2-S 16x4)  & .0297 & .9999 & .6057 & .6463 \\
        (pretrained) MAE (ViT-B 16x4)  & .0300 & .9999 & .6879 & .7077 \\
                \bottomrule[0.1em]        
    \end{tabular}
    }
  \end{center}
  \caption{\textbf{Evaluating self-supervised video representation learning methods}. We evaluate representation learning performance using the linear evaluation protocol on downstream biologically relevant tasks.  (pretrained) indicates pre-training on Kinetics400.
  ↓ indicates MSE and ↑ indicates F1 score. Mouse manual behaviors consist of chase, huddle, face sniff, anal sniff. Beetle manual behaviors consist of grooming, exploring, and idle, either for self (beetle) only or with the interactor. Experimental tasks are described in Table \ref{tab:mouse_all_tasks} and \ref{tab:beetle_behaviors}. The best-performing model is in bold. 
  }
  \label{tab:res_summary_video_methods}
\end{table*}

We study how well behavioral representations generated by state-of-the-art self-supervised video representation learning methods are suited for decoding our hidden downstream biological tasks and human annotations (Section~\ref{sec:benchmarks}). 
We also solicit community-contributed methods for video and trajectory representation learning through an open competition (Section~\ref{sec:community_methods}). 
The representation learned by the models is a mapping from each video frame/trajectory entry to a lower dimensional vector of fixed size. 
Here, we assume the evaluation tasks are hidden during representation learning.
We then use this representation of the data to train a linear model to classify or regress to target values of the hidden downstream task  (Appendix~\ref{sec:evaluation}).

\subsection{Self-supervised Video Representation Learning}~\label{sec:benchmarks}
Self-supervised video representation learning methods rely on designing pretext tasks that make use of prior knowledge about spatial and temporal information in videos to design pretext tasks such as temporal coherence~\cite{goroshin2015unsupervised}, temporal ordering~\cite{misra2016shuffle}, the motion of an object~\cite{agrawal2015learning}, future prediction~\cite{walker2016uncertain}.
Contrastive learning~\cite{chen2020simple, he2020momentum} has been used for learning good visual representations for instance discrimination. 
Another line of work has been introducing methods that solely rely on positive samples~\cite{grill2020bootstrap, caron2020unsupervised}.
In a recent comparison, the video version of Bootstrap Your Own Latent (BYOL)~\cite{grill2020bootstrap} has been shown to perform very well on the classic human benchmarks~\cite{feichtenhofer2021large}, with increased performance for an increased number of positive samples.

\textit{Masked Visual Modeling.} Transformers \cite{vaswani2017attention} set the state-of-the-art across many AI fields, bridging language and vision models. 
Inspired by pretext tasks for language transformer models, such as masking in BERT~\cite{devlin2018bert}, \cite{he2022masked} recently introduced the Masked Auto-Encoder (MAE) for images, an effective pre-training method, by which an image is split into patches, and about 70 percent of the patches are masked. Based on the remaining patches, the task for the transformer is to reconstruct the masked patches.
\cite{feichtenhofer2022masked, tong2022videomae} extended this framework to video, demonstrating transformers can be effectively pre-trained by masking 90 percent of the spatio-temporal volume.
MaskFeat~\cite{wei2022masked} showed that using HOG features~\cite{dalal2005histograms} as reconstruction targets of masked patches is an effective pre-text task.

\subsection{Community-Contributed Methods}~\label{sec:community_methods}
In addition to studying state-of-the-art methods, our benchmarking efforts include community-contributed methods from an \href{https://www.aicrowd.com/challenges/multi-agent-behavior-challenge-2022}{open competition}. 
Our competition was hosted in two stages, where stage 1 consisted of the trajectory datasets from mouse and fly, and stage 2 consisted of video datasets from mouse and beetle.
The test sets were private during the competition phase, and are now released as part of MABe22. 
We obtained around 1500 submissions in total at the end of the competition, and we summarize the top-performing method for the mouse, fly, and beetle datasets from this process for both video and trajectory data, with details for all methods in Appendix Section~\ref{appendix:community}.


\vspace{-0.1in}
\section{Experiments}~\label{sec:experiments}
\vspace{-0.2in}

\begin{table*}[h]
  \begin{center}
\scalebox{0.9}{
    \small
    \begin{tabular}{l|cccccc}
        \toprule[0.2em]
        \multirow{2}{*}{Mice Triplet} &
         Exp.   &   Time of     &   Strain      &       Lights      &   Manual     \\
                &           Day $\downarrow$ &       Day $\downarrow$     &        $\uparrow$           &       $\uparrow$        & Behaviors$\uparrow$ \\        
        \toprule[0.2em]
        
        MAE Frame & .0239 & .0886 & 1.000 & .9525 & .2020 \\
        MAE Cube  & .0102 & \textbf{.0816} & 1.000 & .9758 & \textbf{.2309}\\
        MAE Tube & \textbf{.0072} & .0835 & 1.000 & \textbf{.9846} &  .2249\\
        \bottomrule[0.1em]
        \multirow{2}{*}{Ant Beetle} &
         Duration  & Interactor  & Manual   & Manual  \\
         & $\downarrow$  & Type $\uparrow$ & Behaviors $\uparrow$ & Behaviors (same) $\uparrow$\\       
        \toprule[0.2em]
        MAE Frame & .0301 & .9999 & .6169 & .6497 \\
        MAE Cube & \textbf{.0283} & \textbf{1.0000} & \textbf{.6444} & \textbf{.6874} \\
        MAE Tube  & .0285 & \textbf{1.0000} & .5802 & .6351 \\
                \bottomrule[0.1em]        
    \end{tabular}}
  \end{center}
  \caption{\textbf{Effect of masking strategy on MAE~\cite{feichtenhofer2022masked} performance}. 
  We evaluate different masking strategies (spatiotemporal random/cube, temporal/tube and spatial/frame) on the video datasets of MABe2022. 
  For the mouse dataset cube/tube masking perform best, whereas for the beetle dataset cube/frame masking
perform best.
  ↓ indicates MSE and ↑ indicates F1 score. The best-performing model is in bold.
  }
  \label{tab:res_mae_strategy}
\end{table*}

\begin{table*}[h]
  \begin{center}
\scalebox{0.9}{
    \small
    \begin{tabular}{l|cccccc}
        \toprule[0.2em]
        \multirow{2}{*}{Mice Triplet} &
         Exp.   &   Time of     &   Strain      &       Lights      &   Manual     \\
                &           Day $\downarrow$ &       Day $\downarrow$     &        $\uparrow$           &       $\uparrow$        & Behaviors$\uparrow$ \\        
        \toprule[0.2em]
        2BYOL & .0298 & \textbf{.0882} & .9994 & .9588 & \textbf{.1929} \\
        3BYOL & .0225 & .0906 & .9983 & .9492 & .1733 \\
        4BYOL & \textbf{.0152} & .0913 & .9997 & \textbf{.9701} & .1771 \\
        \bottomrule[0.1em]
        \multirow{2}{*}{Ant Beetle} &
         Duration  & Interactor  & Manual   & Manual  \\
         & $\downarrow$  & Type $\uparrow$ & Behaviors $\uparrow$ & Behaviors (same) $\uparrow$\\       
        \toprule[0.2em]
        2BYOL & \textbf{.0237} & \textbf{1.0000} & .5943 & .6498 \\
        3BYOL & .0246 & \textbf{1.0000} & \textbf{.6249} & \textbf{.6549} \\
        4BYOL & .0257 & .9999 & .6178 & .6457 \\
            \bottomrule[0.1em]
    \end{tabular}}
  \end{center}
  \caption{
  \textbf{Effect of $\rho$ on BYOL~\cite{feichtenhofer2021large} performance}. 
  We evaluated the effect of the number of randomly sampled positives for $\rho$BYOL. We find that for beetle 3 positive samples consistently have the best performance, while for mice, either 2 or 4 positives perform best depending on the task.
  ↓ indicates MSE and ↑ indicates F1 score. The best-performing model is in bold.
  }
  \label{tab:res_byol}
\end{table*}

We perform a large set of experiments to evaluate the performance of representation learning methods on MABe 2022 (Sections~\ref{sec:evaluation_procedure},~\ref{sec:best}). As video representation methods are more common, we focus on state-of-the-art video representation learning methods in this section. We additionally compare both community contributed video and trajectory representation learning methods. 
For each video representation learning method, we perform an ablation study on the key hyperparameter for the respective method and its effect on downstream task performance (Sections~\ref{sec:effect_masking},~\ref{sec:effect_rho}), as well as pre-training on human datasets (Section~\ref{sec:transfer_learning}). Finally, we present results from community-contributed methods on all datasets (Section~\ref{sec:community}), with additional results for the trajectory methods in Appendix~\ref{sec:addtrajectory}.

\subsection{Evaluation Procedure}~\label{sec:evaluation_procedure}
From an input sequence of video/trajectory data of N frames ($N=1800$ for mice and $4500$ for flies), we evaluate models that produce learned representations of size $N \times D$, where $D$ is the dimensionality of the representations. For video representation learning models, we use $D = 128$. For trajectory methods, we use $D=128$ for mice and $D=256$ for flies. We then use these feature vectors or embeddings as inputs for a linear model that is used to classify/regress the hidden task. We use linear least squares with l2 regularized (Ridge) classification/regression as model and F1/mean-squared-error (MSE) as evaluation metrics (See Appendix~\ref{sec:evaluation} for details). 

We evaluate a set of state-of-the-art video representation learning methods on MABe 2022, including Masked Autoencoder (MAE) \cite{feichtenhofer2022masked} with a ViT-B backbone \cite{vaswani2017attention}, MaskFeat \cite{wei2022masked} with a MViTv2-S backbone \cite{li2022mvitv2} and $\rho$BYOL \cite{feichtenhofer2021large} with a SlowFast backbone (Slow pathway 8x8) \cite{frankenhuis2019enriching}.
We trained each method on our mice and beetle data, respectively, as well as used backbones pre-trained on human kinetics 400 \cite{kay2017kinetics}.  For implementation details and hyperparameters see Appendix~\ref{appendix:implementation_details}.

\subsection{Video Representation Results}~\label{sec:best}
We compare the performance of video representation learning methods on the mouse and beetle video datasets (Table \ref{tab:res_summary_video_methods}).
We find that the pre-trained $\rho$BYOL (R-50 (Slow Pathway) 8x8 model performs best for all action recognition tasks (Manuel Behaviors). For all other downstream tasks training, a ViT-B 16x4 Masked Autoencoder (MAE) that is not pre-trained on Kinetics400 generally performs the best. 
This top performing MAE architecture uses spatio-temporal agnostic masking, which likely performs well due to the observation that our datasets have very different spatio-temporal dynamics from each other and even more so from human datasets. We further discuss this in Section~\ref{sec:effect_masking}.
We notice that the model that performs best for human annotated behaviors does not necessarily perform best for our downstream tasks that are based on experimental conditions. 
This indicates that models that pick up features that are most relevant for human perception and behavior definitions may not necessarily be the most informative features for other tasks.

\subsection{Effect of Masking Strategy}~\label{sec:effect_masking}
We explore how different masking strategies (spatiotemporal random/cube, temporal/tube and spatial/frame from MAE~\cite{feichtenhofer2022masked}) affect downstream task performance (Table \ref{tab:res_mae_strategy}), and we use best performing masking ratios used in MAE. 
We find that contrary to \cite{feichtenhofer2022masked, tong2022videomae}, where performances for spatio-temporally agnostic masking (cube) and temporal masking (tube) are very similar to each other, our performance depends on the dataset (mouse or beetle). For the mouse dataset, cube/tube masking have the best overall performance, while for the beetle dataset, cube performs best overall. Overall the differences in performance are also bigger than in  \cite{feichtenhofer2022masked, tong2022videomae}. 
This difference in performance for different masking strategies is likely due to the different spatio-temporal structure of the data, i.e. if the data is more 'temporal heavy' or more 'spatial heavy'.

\begin{table*}
  \begin{center}
\scalebox{0.9}{
    \small
    \begin{tabular}{l|cccccc}
        \toprule[0.2em]
        \multirow{2}{*}{Mice Triplet} &
         Exp.   &   Time of     &   Strain      &       Lights      &   Manual   \\
                &           Day $\downarrow$ &       Day $\downarrow$     &         $\uparrow$          &      $\uparrow$     & Behaviors $\uparrow$   \\             
        \toprule[0.2em]
        BEiT + Hand-crafting & \textbf{.0093} & .0926 & \textbf{1.0000} & \textbf{.9471} & .2603 \\
        Vision Ensemble & .0441 & .0922 & .9832 & .8048 & \textbf{.2750} \\
        Multimodal MoCo/SimCLR & .0394 & \textbf{.0912} & .9902 & .7780 & .2355 \\
        \toprule[0.1em]
        Trajectory-BERT & .0932 & .0996 & .7202 & .6729 & .2379\\
        \bottomrule[0.2em]
        \multirow{2}{*}{Ant Beetle} &
         Duration & Interactor  & Manual  & Manual  \\ 
         & $\downarrow$ & Type $\uparrow$ & Behaviors $\uparrow$ & Behaviors (same) $\uparrow$ \\        
        \toprule[0.2em]
        BEiT + Hand-crafting & .0277 & .9977 & .6761 & .7179 \\
        Vision Ensemble & .0295 & .9636 & .6277 & .6695 \\
        Multimodal MoCo/SimCLR & \textbf{.0262} & \textbf{.9998} & \textbf{.7299}  & \textbf{.7577}  \\
        \bottomrule[0.1em]
        \multirow{2}{*}{Fly Group} &
         Fly & Stimulation, & Stimulation, & Line  & Female  & Manual  \\
         & Type ↑ & Control ↑ & Aggression ↑ &  Category ↑ &  vs. Male ↑ & Behaviors ↑ \\        
        \toprule[0.2em]
        Trajectory-Perceiver & .394 & .418 & .513 & .573 & .982 & .197  \\    
        Trajectory-GPT & .363 & .515 & .500 & .557 & .873 & .246  \\ 
                \bottomrule[0.1em]
    \end{tabular}
    }
  \end{center}
  \caption{\textbf{Benchmarking the community contributed methods}. 
  The best community-contributed methods perform on par or better with self-supervised video representation learning methods. For mice we also have a trajectory-based method to compare to the video-based methods directly. We find that the trajectory-based method generally does not perform as well as the video-based methods on the mouse dataset. For fly task groups, ``Fly type" corresponds to tasks 1 to 11,  ``Stimulation Control" is tasks 12 to 21, ``Stimulation Aggression" is tasks 22 to 36, ``Line Category" is tasks 37 to 43, and ``Manual Behaviors" is tasks 45 to 50 in Appendix Table~\ref{table:flytasks}. 
  ↓ indicates MSE and ↑ indicates F1 score. The best-performing model is in bold.
  }
  \label{tab:res_summary_competition}
\end{table*}

\subsection{Effect of $\rho$ on BYOL}~\label{sec:effect_rho}
We performed $\rho$BYOL \cite{feichtenhofer2021large} with multiple values of $\rho$, i.e., the number of temporal clips sampled as positives (Table~\ref{tab:res_byol}). 
In \cite{feichtenhofer2021large}, a larger number of $\rho$ steadily increases downstream task performance.
This is not true for our datasets, where for mice a value of 2 performs best for 2 tasks and a value of 4 for 2 other tasks.
For the beetle dataset, 3 positive samples achieve the best BYOL performance.
This is likely to the temporally random sampling of positives for BYOL. 
This is likely due to the temporally agnostic sampling method for the clips resulting in positives that are of different actions (as the actions of the animals can change rapidly over temporally close frames). 
Further research is needed on how the temporal sampling strategy for positives needs to be adjusted for temporally heavy datasets.

\subsection{Transfer Learning from Kinetics400}~\label{sec:transfer_learning}
We evaluated how $\rho$BYOL, Maskfeat, and MAE perform when pre-trained on kinetics400~\cite{kay2017kinetics} (Table \ref{tab:res_summary_video_methods}). We find that MAE and Maskfeat training on MABe22 generally performs better than using the pre-trained models. 
Interestingly, for $\rho$BYOL we find the opposite, in that the pre-trained model on Kinetics400 actually performs stronger than counterpart trained on MABe22.
Surprisingly, for action recognition, it performed stronger than any of the other models for both mice and beetle data.
These results suggest that for action recognition, transfer learning from human datasets to animal datasets is possible to a degree.

\subsection{Community-Contributed Methods Results}~\label{sec:community}
We compare community-contributed methods across all datasets in MABe22 (Table~\ref{tab:res_summary_competition}).
The best-performing community methods employ large pre-trained vision models, variations of contrastive learning~\cite{chen2020simple, he2020momentum}, trajectory data as additional inputs and hand-crafted features (See Appendix~\ref{appendix:community_video}). Usually, these features are then concatenated and PCA is performed to produce vectors with the embedding dimension.
For the mouse dataset, we also compared the top trajectory-based method to the video-based methods on the same data subset. While the performance of the trajectory model for behavior classification is similar to the third-best video-based model, the performance on all other downstream tasks is worse. This is likely due to the loss of visual features after transforming the video frames to sparse keypoint locations. 
An interesting direction for future work would be to explore how these modalities can be best combined.
For the fly dataset (which consists of trajectory data only), we find that using a Perceiver model~\cite{jaegle2021perceiver} trained on a masked modeling task works best (See Appendix~\ref{appendix:community_trajectory}).
The second best method is using a GPT~\cite{brown2020language}-like architecture that generates embeddings from the recurrent trajectory data of all agents. 
This method is trained using a prediction pretext task.

In general, we find that performance is comparable between community-contributed methods to state-of-the-art video representation learning methods evaluated in Section~\ref{sec:best}. We note that community methods did perform better at learning manual behaviors. This may be due to the hand-crafted features used in the community-contributed methods, which has been shown to be effective at encoding domain knowledge for behavior analysis~\cite{sun2020task}.


\section{Discussion and Future Directions}

We introduced a novel multi-species multi-task performance benchmark to evaluate representation learning for social behavior from video and trajectory data. The dataset consists of video and trajectory data captured across three organisms. The evaluation methods are based on the performance of a broad palette of tasks that are based on scientific experimental conditions that are independent of the actions annotated by human experts. We demonstrate the use of our benchmark, and provide a baseline, by evaluating state-of-the-art self-supervised video-representation learning. Additionally, we provide results from methods that were part of a recent competition for learning behavioral representations.

We compare method performance on our benchmark with pre-training on existing benchmarks using human video datasets.
We find that methods that perform best on human datasets may not perform the best on our animal datasets. This is likely because human action datasets contain extraneous visual information, whereas our animal datasets minimize these visual cues (consistent backgrounds) and thus behavioral representations need to focus on spatio-temporal information. 
This highlights a crucial shortcoming of current benchmarks, which may be pushing the community to develop methods that do not focus on the spatio-temporal nature of behavior.
We hope to encourage evaluation of representation learning methods on a broader range of settings beyond human videos and annotations, in order to facilitate development of new methods for representation learning and behavior analysis.

\section{Acknowledgements}
This work was generously supported by the Simons Collaboration on the Global Brain grant 543025 (to PP), NIH Award \#R00MH117264 (to AK), NSF Award \#1918839 (to YY), NIH 1R34NS118470-01 (to JP), NSERC Award \#PGSD3-532647-2019 (to JJS), as well as a gift from Charles and Lily Trimble (to PP). We would like to thank Tom Sproule for mouse breeding and dataset collection. The mouse dataset was supported by the National Institute of Health DA041668 (NIDA), DA048634 (NIDA, and Simons Foundation SFARI Director’s Award) (to VK). We also greatly appreciate Google, Amazon, HHMI, and the Simons Foundation for sponsoring the MABe22 Challenge \& Workshop.

\bibliography{example_paper}

\begin{thebibliography}{94}
\providecommand{\natexlab}[1]{#1}
\providecommand{\url}[1]{\texttt{#1}}
\expandafter\ifx\csname urlstyle\endcsname\relax
  \providecommand{\doi}[1]{doi: #1}\else
  \providecommand{\doi}{doi: \begingroup \urlstyle{rm}\Url}\fi

\bibitem[Agrawal et~al.(2015)Agrawal, Carreira, and Malik]{agrawal2015learning}
Agrawal, P., Carreira, J., and Malik, J.
\newblock Learning to see by moving.
\newblock In \emph{Proceedings of the IEEE international conference on computer
  vision}, pp.\  37--45, 2015.

\bibitem[Anderson \& Perona(2014)Anderson and Perona]{anderson2014toward}
Anderson, D.~J. and Perona, P.
\newblock Toward a science of computational ethology.
\newblock \emph{Neuron}, 84\penalty0 (1):\penalty0 18--31, 2014.

\bibitem[Aso et~al.(2014)Aso, Sitaraman, Ichinose, Kaun, Vogt,
  Belliart-Gu{\'e}rin, Pla{\c{c}}ais, Robie, Yamagata, Schnaitmann,
  et~al.]{aso2014mushroom}
Aso, Y., Sitaraman, D., Ichinose, T., Kaun, K.~R., Vogt, K.,
  Belliart-Gu{\'e}rin, G., Pla{\c{c}}ais, P.-Y., Robie, A.~A., Yamagata, N.,
  Schnaitmann, C., et~al.
\newblock Mushroom body output neurons encode valence and guide memory-based
  action selection in drosophila.
\newblock \emph{Elife}, 3:\penalty0 e04580, 2014.

\bibitem[Bao et~al.(2021)Bao, Dong, and Wei]{bao2021beit}
Bao, H., Dong, L., and Wei, F.
\newblock Beit: Bert pre-training of image transformers.
\newblock \emph{arXiv preprint arXiv:2106.08254}, 2021.

\bibitem[Beane et~al.(2022)Beane, Geuther, Sproule, Trapszo, Hession, Kohar,
  and Kumar]{beane2022video}
Beane, G., Geuther, B.~Q., Sproule, T.~J., Trapszo, J., Hession, L., Kohar, V.,
  and Kumar, V.
\newblock Video based phenotyping platform for the laboratory mouse.
\newblock \emph{bioRxiv}, 2022.

\bibitem[Berman et~al.(2014)Berman, Choi, Bialek, and
  Shaevitz]{berman2014mapping}
Berman, G.~J., Choi, D.~M., Bialek, W., and Shaevitz, J.~W.
\newblock Mapping the stereotyped behaviour of freely moving fruit flies.
\newblock \emph{Journal of The Royal Society Interface}, 11\penalty0
  (99):\penalty0 20140672, 2014.

\bibitem[Bloomquist et~al.(1988)Bloomquist, Shortridge, Schneuwly, Perdew,
  Montell, Steller, Rubin, and Pak]{bloomquist1988isolation}
Bloomquist, B.~T., Shortridge, R., Schneuwly, S., Perdew, M., Montell, C.,
  Steller, H., Rubin, G., and Pak, W.
\newblock Isolation of a putative phospholipase c gene of drosophila, norpa,
  and its role in phototransduction.
\newblock \emph{Cell}, 54\penalty0 (5):\penalty0 723--733, 1988.

\bibitem[Bohnslav et~al.(2021)Bohnslav, Wimalasena, Clausing, Dai, Yarmolinsky,
  Cruz, Kashlan, Chiappe, Orefice, Woolf, et~al.]{bohnslav2021deepethogram}
Bohnslav, J.~P., Wimalasena, N.~K., Clausing, K.~J., Dai, Y.~Y., Yarmolinsky,
  D.~A., Cruz, T., Kashlan, A.~D., Chiappe, M.~E., Orefice, L.~L., Woolf,
  C.~J., et~al.
\newblock Deepethogram, a machine learning pipeline for supervised behavior
  classification from raw pixels.
\newblock \emph{Elife}, 10:\penalty0 e63377, 2021.

\bibitem[Brown et~al.(2020)Brown, Mann, Ryder, Subbiah, Kaplan, Dhariwal,
  Neelakantan, Shyam, Sastry, Askell, et~al.]{brown2020language}
Brown, T., Mann, B., Ryder, N., Subbiah, M., Kaplan, J.~D., Dhariwal, P.,
  Neelakantan, A., Shyam, P., Sastry, G., Askell, A., et~al.
\newblock Language models are few-shot learners.
\newblock \emph{Advances in neural information processing systems},
  33:\penalty0 1877--1901, 2020.

\bibitem[Burgos-Artizzu et~al.(2012)Burgos-Artizzu, Doll{\'a}r, Lin, Anderson,
  and Perona]{burgos2012social}
Burgos-Artizzu, X.~P., Doll{\'a}r, P., Lin, D., Anderson, D.~J., and Perona, P.
\newblock Social behavior recognition in continuous video.
\newblock In \emph{2012 IEEE Conference on Computer Vision and Pattern
  Recognition}, pp.\  1322--1329. IEEE, 2012.

\bibitem[Calhoun et~al.(2019)Calhoun, Pillow, and
  Murthy]{calhoun2019unsupervised}
Calhoun, A.~J., Pillow, J.~W., and Murthy, M.
\newblock Unsupervised identification of the internal states that shape natural
  behavior.
\newblock \emph{Nature neuroscience}, 22\penalty0 (12):\penalty0 2040--2049,
  2019.

\bibitem[Caron et~al.(2020)Caron, Misra, Mairal, Goyal, Bojanowski, and
  Joulin]{caron2020unsupervised}
Caron, M., Misra, I., Mairal, J., Goyal, P., Bojanowski, P., and Joulin, A.
\newblock Unsupervised learning of visual features by contrasting cluster
  assignments.
\newblock \emph{Advances in neural information processing systems},
  33:\penalty0 9912--9924, 2020.

\bibitem[Chang et~al.(2019)Chang, Lambert, Sangkloy, Singh, Bak, Hartnett,
  Wang, Carr, Lucey, Ramanan, et~al.]{chang2019argoverse}
Chang, M.-F., Lambert, J., Sangkloy, P., Singh, J., Bak, S., Hartnett, A.,
  Wang, D., Carr, P., Lucey, S., Ramanan, D., et~al.
\newblock Argoverse: 3d tracking and forecasting with rich maps.
\newblock In \emph{Proceedings of the IEEE Conference on Computer Vision and
  Pattern Recognition}, pp.\  8748--8757, 2019.

\bibitem[Chen et~al.(2020{\natexlab{a}})Chen, Radford, Child, Wu, Jun, Luan,
  and Sutskever]{chen2020generative}
Chen, M., Radford, A., Child, R., Wu, J., Jun, H., Luan, D., and Sutskever, I.
\newblock Generative pretraining from pixels.
\newblock In \emph{International conference on machine learning}, pp.\
  1691--1703. PMLR, 2020{\natexlab{a}}.

\bibitem[Chen et~al.(2020{\natexlab{b}})Chen, Kornblith, Norouzi, and
  Hinton]{chen2020simple}
Chen, T., Kornblith, S., Norouzi, M., and Hinton, G.
\newblock A simple framework for contrastive learning of visual
  representations.
\newblock \emph{ICML}, 2020{\natexlab{b}}.

\bibitem[Co-Reyes et~al.(2018)Co-Reyes, Liu, Gupta, Eysenbach, Abbeel, and
  Levine]{co2018self}
Co-Reyes, J.~D., Liu, Y., Gupta, A., Eysenbach, B., Abbeel, P., and Levine, S.
\newblock Self-consistent trajectory autoencoder: Hierarchical reinforcement
  learning with trajectory embeddings.
\newblock \emph{arXiv preprint arXiv:1806.02813}, 2018.

\bibitem[Dalal \& Triggs(2005)Dalal and Triggs]{dalal2005histograms}
Dalal, N. and Triggs, B.
\newblock Histograms of oriented gradients for human detection.
\newblock In \emph{2005 IEEE computer society conference on computer vision and
  pattern recognition (CVPR'05)}, volume~1, pp.\  886--893. Ieee, 2005.

\bibitem[Decroos et~al.(2018)Decroos, Van~Haaren, and
  Davis]{decroos2018automatic}
Decroos, T., Van~Haaren, J., and Davis, J.
\newblock Automatic discovery of tactics in spatio-temporal soccer match data.
\newblock In \emph{Proceedings of the 24th acm sigkdd international conference
  on knowledge discovery \& data mining}, pp.\  223--232, 2018.

\bibitem[Deng et~al.(2009)Deng, Dong, Socher, Li, Li, and
  Fei-Fei]{deng2009imagenet}
Deng, J., Dong, W., Socher, R., Li, L.-J., Li, K., and Fei-Fei, L.
\newblock Imagenet: A large-scale hierarchical image database.
\newblock In \emph{2009 IEEE conference on computer vision and pattern
  recognition}, pp.\  248--255. Ieee, 2009.

\bibitem[Devlin et~al.(2018)Devlin, Chang, Lee, and Toutanova]{devlin2018bert}
Devlin, J., Chang, M.-W., Lee, K., and Toutanova, K.
\newblock Bert: Pre-training of deep bidirectional transformers for language
  understanding.
\newblock \emph{arXiv preprint arXiv:1810.04805}, 2018.

\bibitem[Devlin et~al.(2019)Devlin, Chang, Lee, and Toutanova]{bert}
Devlin, J., Chang, M.-W., Lee, K., and Toutanova, K.
\newblock {BERT}: Pre-training of deep bidirectional transformers for language
  understanding.
\newblock In \emph{Proceedings of the 2019 Conference of the North {A}merican
  Chapter of the Association for Computational Linguistics: Human Language
  Technologies, Volume 1 (Long and Short Papers)}, pp.\  4171--4186,
  Minneapolis, Minnesota, June 2019. Association for Computational Linguistics.
\newblock \doi{10.18653/v1/N19-1423}.
\newblock URL \url{https://aclanthology.org/N19-1423}.

\bibitem[Dutta \& Zisserman(2019)Dutta and Zisserman]{dutta2019via}
Dutta, A. and Zisserman, A.
\newblock The via annotation software for images, audio and video.
\newblock In \emph{Proceedings of the 27th ACM international conference on
  multimedia}, pp.\  2276--2279, 2019.

\bibitem[Eyjolfsdottir et~al.(2014)Eyjolfsdottir, Branson, Burgos-Artizzu,
  Hoopfer, Schor, Anderson, and Perona]{eyjolfsdottir2014detecting}
Eyjolfsdottir, E., Branson, S., Burgos-Artizzu, X.~P., Hoopfer, E.~D., Schor,
  J., Anderson, D.~J., and Perona, P.
\newblock Detecting social actions of fruit flies.
\newblock In \emph{European Conference on Computer Vision}, pp.\  772--787.
  Springer, 2014.

\bibitem[Fan et~al.(2020)Fan, Li, Xiong, Lo, and
  Feichtenhofer]{fan2020pyslowfast}
Fan, H., Li, Y., Xiong, B., Lo, W.-Y., and Feichtenhofer, C.
\newblock Pyslowfast.
\newblock \url{https://github.com/facebookresearch/slowfast}, 2020.

\bibitem[Feichtenhofer et~al.(2021)Feichtenhofer, Fan, Xiong, Girshick, and
  He]{feichtenhofer2021large}
Feichtenhofer, C., Fan, H., Xiong, B., Girshick, R., and He, K.
\newblock A large-scale study on unsupervised spatiotemporal representation
  learning.
\newblock In \emph{Proceedings of the IEEE/CVF Conference on Computer Vision
  and Pattern Recognition}, pp.\  3299--3309, 2021.

\bibitem[Feichtenhofer et~al.(2022)Feichtenhofer, Fan, Li, and
  He]{feichtenhofer2022masked}
Feichtenhofer, C., Fan, H., Li, Y., and He, K.
\newblock Masked autoencoders as spatiotemporal learners.
\newblock \emph{arXiv preprint arXiv:2205.09113}, 2022.

\bibitem[Frankenhuis et~al.(2019)Frankenhuis, Panchanathan, and
  Barto]{frankenhuis2019enriching}
Frankenhuis, W.~E., Panchanathan, K., and Barto, A.~G.
\newblock Enriching behavioral ecology with reinforcement learning methods.
\newblock \emph{Behavioural processes}, 161:\penalty0 94--100, 2019.

\bibitem[Gebru et~al.(2018)Gebru, Morgenstern, Vecchione, Vaughan, Wallach,
  Daum{\'e}~III, and Crawford]{gebru2018datasheets}
Gebru, T., Morgenstern, J., Vecchione, B., Vaughan, J.~W., Wallach, H.,
  Daum{\'e}~III, H., and Crawford, K.
\newblock Datasheets for datasets.
\newblock \emph{arXiv preprint arXiv:1803.09010}, 2018.

\bibitem[Geuther et~al.(2019)Geuther, Deats, Fox, Murray, Braun, White,
  Chesler, Lutz, and Kumar]{geuther2019robust}
Geuther, B.~Q., Deats, S.~P., Fox, K.~J., Murray, S.~A., Braun, R.~E., White,
  J.~K., Chesler, E.~J., Lutz, C.~M., and Kumar, V.
\newblock Robust mouse tracking in complex environments using neural networks.
\newblock \emph{Communications biology}, 2\penalty0 (1):\penalty0 1--11, 2019.

\bibitem[Gidaris et~al.(2018)Gidaris, Singh, and
  Komodakis]{gidaris2018unsupervised}
Gidaris, S., Singh, P., and Komodakis, N.
\newblock Unsupervised representation learning by predicting image rotations.
\newblock \emph{ICLR}, 2018.

\bibitem[Goroshin et~al.(2015)Goroshin, Bruna, Tompson, Eigen, and
  LeCun]{goroshin2015unsupervised}
Goroshin, R., Bruna, J., Tompson, J., Eigen, D., and LeCun, Y.
\newblock Unsupervised learning of spatiotemporally coherent metrics.
\newblock In \emph{Proceedings of the IEEE international conference on computer
  vision}, pp.\  4086--4093, 2015.

\bibitem[Goyal et~al.(2017)Goyal, Doll{\'a}r, Girshick, Noordhuis, Wesolowski,
  Kyrola, Tulloch, Jia, and He]{goyal2017accurate}
Goyal, P., Doll{\'a}r, P., Girshick, R., Noordhuis, P., Wesolowski, L., Kyrola,
  A., Tulloch, A., Jia, Y., and He, K.
\newblock Accurate, large minibatch sgd: Training imagenet in 1 hour.
\newblock \emph{arXiv preprint arXiv:1706.02677}, 2017.

\bibitem[Grill et~al.(2020)Grill, Strub, Altch{\'e}, Tallec, Richemond,
  Buchatskaya, Doersch, Avila~Pires, Guo, Gheshlaghi~Azar,
  et~al.]{grill2020bootstrap}
Grill, J.-B., Strub, F., Altch{\'e}, F., Tallec, C., Richemond, P.,
  Buchatskaya, E., Doersch, C., Avila~Pires, B., Guo, Z., Gheshlaghi~Azar, M.,
  et~al.
\newblock Bootstrap your own latent-a new approach to self-supervised learning.
\newblock \emph{Advances in neural information processing systems},
  33:\penalty0 21271--21284, 2020.

\bibitem[Guss et~al.(2019)Guss, Houghton, Topin, Wang, Codel, Veloso, and
  Salakhutdinov]{guss2019minerl}
Guss, W.~H., Houghton, B., Topin, N., Wang, P., Codel, C., Veloso, M., and
  Salakhutdinov, R.
\newblock Minerl: A large-scale dataset of minecraft demonstrations.
\newblock \emph{arXiv preprint arXiv:1907.13440}, 2019.

\bibitem[Han et~al.(2019)Han, Xie, and Zisserman]{han2019video}
Han, T., Xie, W., and Zisserman, A.
\newblock Video representation learning by dense predictive coding.
\newblock In \emph{Proceedings of the IEEE International Conference on Computer
  Vision Workshops}, pp.\  0--0, 2019.

\bibitem[He et~al.(2016)He, Zhang, Ren, and Sun]{he2016deep}
He, K., Zhang, X., Ren, S., and Sun, J.
\newblock Deep residual learning for image recognition.
\newblock In \emph{Proceedings of the IEEE conference on computer vision and
  pattern recognition}, pp.\  770--778, 2016.

\bibitem[He et~al.(2020)He, Fan, Wu, Xie, and Girshick]{he2020momentum}
He, K., Fan, H., Wu, Y., Xie, S., and Girshick, R.
\newblock Momentum contrast for unsupervised visual representation learning.
\newblock In \emph{Proceedings of the IEEE/CVF conference on computer vision
  and pattern recognition}, pp.\  9729--9738, 2020.

\bibitem[He et~al.(2022)He, Chen, Xie, Li, Doll{\'a}r, and
  Girshick]{he2022masked}
He, K., Chen, X., Xie, S., Li, Y., Doll{\'a}r, P., and Girshick, R.
\newblock Masked autoencoders are scalable vision learners.
\newblock In \emph{Proceedings of the IEEE/CVF Conference on Computer Vision
  and Pattern Recognition}, pp.\  16000--16009, 2022.

\bibitem[Hern{\'a}ndez et~al.(2020)Hern{\'a}ndez, Rivera, Cande, Zhou, Stern,
  and Berman]{hernandez2020framework}
Hern{\'a}ndez, D.~G., Rivera, C., Cande, J., Zhou, B., Stern, D.~L., and
  Berman, G.~J.
\newblock A framework for studying behavioral evolution by reconstructing
  ancestral repertoires.
\newblock \emph{arXiv preprint arXiv:2007.09689}, 2020.

\bibitem[Hofmann(2019)]{hofmann2019minecraft}
Hofmann, K.
\newblock Minecraft as ai playground and laboratory.
\newblock In \emph{Proceedings of the Annual Symposium on Computer-Human
  Interaction in Play}, pp.\  1--1, 2019.

\bibitem[Hong et~al.(2015)Hong, Kennedy, Burgos-Artizzu, Zelikowsky, Navonne,
  Perona, and Anderson]{hong2015automated}
Hong, W., Kennedy, A., Burgos-Artizzu, X.~P., Zelikowsky, M., Navonne, S.~G.,
  Perona, P., and Anderson, D.~J.
\newblock Automated measurement of mouse social behaviors using depth sensing,
  video tracking, and machine learning.
\newblock \emph{Proceedings of the National Academy of Sciences}, 112\penalty0
  (38):\penalty0 E5351--E5360, 2015.

\bibitem[Howard et~al.(2019)Howard, Sandler, Chu, Chen, Chen, Tan, Wang, Zhu,
  Pang, Vasudevan, et~al.]{howard2019searching}
Howard, A., Sandler, M., Chu, G., Chen, L.-C., Chen, B., Tan, M., Wang, W.,
  Zhu, Y., Pang, R., Vasudevan, V., et~al.
\newblock Searching for mobilenetv3.
\newblock In \emph{Proceedings of the IEEE/CVF international conference on
  computer vision}, pp.\  1314--1324, 2019.

\bibitem[Hsu \& Yttri(2020)Hsu and Yttri]{hsu2020b}
Hsu, A.~I. and Yttri, E.~A.
\newblock B-soid: An open source unsupervised algorithm for discovery of
  spontaneous behaviors.
\newblock \emph{bioRxiv}, pp.\  770271, 2020.

\bibitem[Jaegle et~al.(2021)Jaegle, Borgeaud, Alayrac, Doersch, Ionescu, Ding,
  Koppula, Zoran, Brock, Shelhamer, et~al.]{jaegle2021perceiver}
Jaegle, A., Borgeaud, S., Alayrac, J.-B., Doersch, C., Ionescu, C., Ding, D.,
  Koppula, S., Zoran, D., Brock, A., Shelhamer, E., et~al.
\newblock Perceiver io: A general architecture for structured inputs \&
  outputs.
\newblock \emph{arXiv preprint arXiv:2107.14795}, 2021.

\bibitem[Joshi et~al.(2020)Joshi, Chen, Liu, Weld, Zettlemoyer, and
  Levy]{spanbert}
Joshi, M., Chen, D., Liu, Y., Weld, D.~S., Zettlemoyer, L., and Levy, O.
\newblock {S}pan{BERT}: Improving pre-training by representing and predicting
  spans.
\newblock \emph{Transactions of the Association for Computational Linguistics},
  8:\penalty0 64--77, 2020.
\newblock \doi{10.1162/tacl_a_00300}.
\newblock URL \url{https://aclanthology.org/2020.tacl-1.5}.

\bibitem[Kabra et~al.(2013)Kabra, Robie, Rivera-Alba, Branson, and
  Branson]{kabra2013jaaba}
Kabra, M., Robie, A.~A., Rivera-Alba, M., Branson, S., and Branson, K.
\newblock Jaaba: interactive machine learning for automatic annotation of
  animal behavior.
\newblock \emph{Nature methods}, 10\penalty0 (1):\penalty0 64, 2013.

\bibitem[Kabra et~al.(2022)Kabra, Lee, Robie, Egnor, Huston, Rodriguez,
  Edwards, and Branson]{APT}
Kabra, M., Lee, A., Robie, A., Egnor, R., Huston, S., Rodriguez, I.~F.,
  Edwards, A., and Branson, K.
\newblock Apt: Animal part tracker v0.3.4, March 2022.
\newblock URL \url{https://doi.org/10.5281/zenodo.6366082}.

\bibitem[Kay et~al.(2017)Kay, Carreira, Simonyan, Zhang, Hillier,
  Vijayanarasimhan, Viola, Green, Back, Natsev, et~al.]{kay2017kinetics}
Kay, W., Carreira, J., Simonyan, K., Zhang, B., Hillier, C., Vijayanarasimhan,
  S., Viola, F., Green, T., Back, T., Natsev, P., et~al.
\newblock The kinetics human action video dataset.
\newblock \emph{arXiv preprint arXiv:1705.06950}, 2017.

\bibitem[Kingma \& Ba(2014)Kingma and Ba]{kingma2014adam}
Kingma, D.~P. and Ba, J.
\newblock Adam: A method for stochastic optimization.
\newblock \emph{arXiv preprint arXiv:1412.6980}, 2014.

\bibitem[Kingma \& Welling(2014)Kingma and Welling]{kingma2013auto}
Kingma, D.~P. and Welling, M.
\newblock Auto-encoding variational bayes.
\newblock In \emph{International Conference on Learning Representations}, 2014.

\bibitem[Klapoetke et~al.(2014)Klapoetke, Murata, Kim, Pulver, Birdsey-Benson,
  Cho, Morimoto, Chuong, Carpenter, Tian, et~al.]{klapoetke2014independent}
Klapoetke, N.~C., Murata, Y., Kim, S.~S., Pulver, S.~R., Birdsey-Benson, A.,
  Cho, Y.~K., Morimoto, T.~K., Chuong, A.~S., Carpenter, E.~J., Tian, Z.,
  et~al.
\newblock Independent optical excitation of distinct neural populations.
\newblock \emph{Nature methods}, 11\penalty0 (3):\penalty0 338--346, 2014.

\bibitem[Klibaite et~al.(2017)Klibaite, Berman, Cande, Stern, and
  Shaevitz]{klibaite2017unsupervised}
Klibaite, U., Berman, G.~J., Cande, J., Stern, D.~L., and Shaevitz, J.~W.
\newblock An unsupervised method for quantifying the behavior of paired
  animals.
\newblock \emph{Physical biology}, 14\penalty0 (1):\penalty0 015006, 2017.

\bibitem[Klibaite et~al.(2022)Klibaite, Kislin, Verpeut, Bergeler, Sun,
  Shaevitz, and Wang]{klibaite2022deep}
Klibaite, U., Kislin, M., Verpeut, J.~L., Bergeler, S., Sun, X., Shaevitz,
  J.~W., and Wang, S. S.-H.
\newblock Deep phenotyping reveals movement phenotypes in mouse
  neurodevelopmental models.
\newblock \emph{Molecular Autism}, 13\penalty0 (1):\penalty0 1--18, 2022.

\bibitem[Kolesnikov et~al.(2019)Kolesnikov, Zhai, and
  Beyer]{kolesnikov2019revisiting}
Kolesnikov, A., Zhai, X., and Beyer, L.
\newblock Revisiting self-supervised visual representation learning.
\newblock In \emph{Proceedings of the IEEE conference on Computer Vision and
  Pattern Recognition}, pp.\  1920--1929, 2019.

\bibitem[Kuehne et~al.(2011)Kuehne, Jhuang, Garrote, Poggio, and
  Serre]{kuehne2011hmdb}
Kuehne, H., Jhuang, H., Garrote, E., Poggio, T., and Serre, T.
\newblock Hmdb: a large video database for human motion recognition.
\newblock In \emph{2011 International conference on computer vision}, pp.\
  2556--2563. IEEE, 2011.

\bibitem[Li et~al.(2022)Li, Wu, Fan, Mangalam, Xiong, Malik, and
  Feichtenhofer]{li2022mvitv2}
Li, Y., Wu, C.-Y., Fan, H., Mangalam, K., Xiong, B., Malik, J., and
  Feichtenhofer, C.
\newblock Mvitv2: Improved multiscale vision transformers for classification
  and detection.
\newblock In \emph{Proceedings of the IEEE/CVF Conference on Computer Vision
  and Pattern Recognition}, pp.\  4804--4814, 2022.

\bibitem[Loshchilov \& Hutter()Loshchilov and Hutter]{loshchilovstochastic}
Loshchilov, I. and Hutter, F.
\newblock Stochastic gradient descent with warm restarts.
\newblock In \emph{Proceedings of the 5th Int. Conf. Learning Representations},
  pp.\  1--16.

\bibitem[Loshchilov \& Hutter(2017{\natexlab{a}})Loshchilov and Hutter]{adamw}
Loshchilov, I. and Hutter, F.
\newblock Fixing weight decay regularization in adam.
\newblock \emph{CoRR}, abs/1711.05101, 2017{\natexlab{a}}.
\newblock URL \url{http://arxiv.org/abs/1711.05101}.

\bibitem[Loshchilov \& Hutter(2017{\natexlab{b}})Loshchilov and
  Hutter]{loshchilov2017decoupled}
Loshchilov, I. and Hutter, F.
\newblock Decoupled weight decay regularization.
\newblock \emph{arXiv preprint arXiv:1711.05101}, 2017{\natexlab{b}}.

\bibitem[Luxem et~al.(2020)Luxem, Fuhrmann, K{\"u}rsch, Remy, and
  Bauer]{luxem2020identifying}
Luxem, K., Fuhrmann, F., K{\"u}rsch, J., Remy, S., and Bauer, P.
\newblock Identifying behavioral structure from deep variational embeddings of
  animal motion.
\newblock \emph{bioRxiv}, 2020.

\bibitem[Markowitz et~al.(2018)Markowitz, Gillis, Beron, Neufeld, Robertson,
  Bhagat, Peterson, Peterson, Hyun, Linderman, et~al.]{markowitz2018striatum}
Markowitz, J.~E., Gillis, W.~F., Beron, C.~C., Neufeld, S.~Q., Robertson, K.,
  Bhagat, N.~D., Peterson, R.~E., Peterson, E., Hyun, M., Linderman, S.~W.,
  et~al.
\newblock The striatum organizes 3d behavior via moment-to-moment action
  selection.
\newblock \emph{Cell}, 174\penalty0 (1):\penalty0 44--58, 2018.

\bibitem[Marks et~al.(2022)Marks, Jin, Sturman, von Ziegler, Kollmorgen,
  von~der Behrens, Mante, Bohacek, and Yanik]{marks2022deep}
Marks, M., Jin, Q., Sturman, O., von Ziegler, L., Kollmorgen, S., von~der
  Behrens, W., Mante, V., Bohacek, J., and Yanik, M.~F.
\newblock Deep-learning-based identification, tracking, pose estimation and
  behaviour classification of interacting primates and mice in complex
  environments.
\newblock \emph{Nature Machine Intelligence}, 4\penalty0 (4):\penalty0
  331--340, 2022.

\bibitem[Marques et~al.(2018)Marques, Lackner, F{\'e}lix, and
  Orger]{marques2018structure}
Marques, J.~C., Lackner, S., F{\'e}lix, R., and Orger, M.~B.
\newblock Structure of the zebrafish locomotor repertoire revealed with
  unsupervised behavioral clustering.
\newblock \emph{Current Biology}, 28\penalty0 (2):\penalty0 181--195, 2018.

\bibitem[McInnes et~al.(2018)McInnes, Healy, and Melville]{mcinnes2018umap}
McInnes, L., Healy, J., and Melville, J.
\newblock Umap: Uniform manifold approximation and projection for dimension
  reduction.
\newblock \emph{arXiv preprint arXiv:1802.03426}, 2018.

\bibitem[Misra et~al.(2016)Misra, Zitnick, and Hebert]{misra2016shuffle}
Misra, I., Zitnick, C.~L., and Hebert, M.
\newblock Shuffle and learn: unsupervised learning using temporal order
  verification.
\newblock In \emph{European conference on computer vision}, pp.\  527--544.
  Springer, 2016.

\bibitem[Newell et~al.(2017)Newell, Huang, and Deng]{newell2017associative}
Newell, A., Huang, Z., and Deng, J.
\newblock Associative embedding: End-to-end learning for joint detection and
  grouping.
\newblock \emph{Advances in neural information processing systems}, 30, 2017.

\bibitem[Ng et~al.(2022)Ng, Ong, Zheng, Ni, Yeo, and Liu]{ng2022animal}
Ng, X.~L., Ong, K.~E., Zheng, Q., Ni, Y., Yeo, S.~Y., and Liu, J.
\newblock Animal kingdom: A large and diverse dataset for animal behavior
  understanding.
\newblock In \emph{Proceedings of the IEEE/CVF Conference on Computer Vision
  and Pattern Recognition}, pp.\  19023--19034, 2022.

\bibitem[Nilsen et~al.(2004)Nilsen, Chan, Huber, and Kravitz]{nilsen2004gender}
Nilsen, S.~P., Chan, Y.-B., Huber, R., and Kravitz, E.~A.
\newblock Gender-selective patterns of aggressive behavior in drosophila
  melanogaster.
\newblock \emph{Proceedings of the National Academy of Sciences}, 101\penalty0
  (33):\penalty0 12342--12347, 2004.

\bibitem[Oord et~al.(2018)Oord, Li, and Vinyals]{oord2018representation}
Oord, A. v.~d., Li, Y., and Vinyals, O.
\newblock Representation learning with contrastive predictive coding.
\newblock \emph{arXiv preprint arXiv:1807.03748}, 2018.

\bibitem[Pereira et~al.(2020)Pereira, Shaevitz, and
  Murthy]{pereira2020quantifying}
Pereira, T.~D., Shaevitz, J.~W., and Murthy, M.
\newblock Quantifying behavior to understand the brain.
\newblock \emph{Nature neuroscience}, pp.\  1--13, 2020.

\bibitem[Qi et~al.(2017)Qi, Su, Mo, and Guibas]{qi2017pointnet}
Qi, C.~R., Su, H., Mo, K., and Guibas, L.~J.
\newblock Pointnet: Deep learning on point sets for 3d classification and
  segmentation.
\newblock In \emph{Proceedings of the IEEE conference on computer vision and
  pattern recognition}, pp.\  652--660, 2017.

\bibitem[Robie et~al.(2017)Robie, Hirokawa, Edwards, Umayam, Lee, Phillips,
  Card, Korff, Rubin, Simpson, et~al.]{robie2017mapping}
Robie, A.~A., Hirokawa, J., Edwards, A.~W., Umayam, L.~A., Lee, A., Phillips,
  M.~L., Card, G.~M., Korff, W., Rubin, G.~M., Simpson, J.~H., et~al.
\newblock Mapping the neural substrates of behavior.
\newblock \emph{Cell}, 170\penalty0 (2):\penalty0 393--406, 2017.

\bibitem[Samvelyan et~al.(2019)Samvelyan, Rashid, De~Witt, Farquhar, Nardelli,
  Rudner, Hung, Torr, Foerster, and Whiteson]{samvelyan2019starcraft}
Samvelyan, M., Rashid, T., De~Witt, C.~S., Farquhar, G., Nardelli, N., Rudner,
  T.~G., Hung, C.-M., Torr, P.~H., Foerster, J., and Whiteson, S.
\newblock The starcraft multi-agent challenge.
\newblock \emph{arXiv preprint arXiv:1902.04043}, 2019.

\bibitem[Schretter et~al.(2020)Schretter, Aso, Robie, Dreher, Dolan, Chen, Ito,
  Yang, Parekh, Branson, et~al.]{schretter2020cell}
Schretter, C.~E., Aso, Y., Robie, A.~A., Dreher, M., Dolan, M.-J., Chen, N.,
  Ito, M., Yang, T., Parekh, R., Branson, K.~M., et~al.
\newblock Cell types and neuronal circuitry underlying female aggression in
  drosophila.
\newblock \emph{Elife}, 9:\penalty0 e58942, 2020.

\bibitem[Schrimpf et~al.(2020)Schrimpf, Kubilius, Lee, Murty, Ajemian, and
  DiCarlo]{Schrimpf2020integrative}
Schrimpf, M., Kubilius, J., Lee, M.~J., Murty, N. A.~R., Ajemian, R., and
  DiCarlo, J.~J.
\newblock Integrative benchmarking to advance neurally mechanistic models of
  human intelligence.
\newblock \emph{Neuron}, 2020.
\newblock URL \url{https://www.cell.com/neuron/fulltext/S0896-6273(20)30605-X}.

\bibitem[Segalin et~al.(2020)Segalin, Williams, Karigo, Hui, Zelikowsky, Sun,
  Perona, Anderson, and Kennedy]{segalin2020mouse}
Segalin, C., Williams, J., Karigo, T., Hui, M., Zelikowsky, M., Sun, J.~J.,
  Perona, P., Anderson, D.~J., and Kennedy, A.
\newblock The mouse action recognition system (mars): a software pipeline for
  automated analysis of social behaviors in mice.
\newblock \emph{bioRxiv}, 2020.

\bibitem[Sheppard et~al.(2022)Sheppard, Gardin, Sabnis, Peer, Darrell, Deats,
  Geuther, Lutz, and Kumar]{sheppard2020gait}
Sheppard, K., Gardin, J., Sabnis, G., Peer, A., Darrell, M., Deats, S.,
  Geuther, B., Lutz, C.~M., and Kumar, V.
\newblock Stride-level analysis of mouse open field behavior using
  deep-learning-based pose estimation.
\newblock \emph{Cell Reports}, 2022.

\bibitem[Sokolowski(2001)]{sokolowski2001drosophila}
Sokolowski, M.~B.
\newblock Drosophila: genetics meets behaviour.
\newblock \emph{Nature Reviews Genetics}, 2\penalty0 (11):\penalty0 879--890,
  2001.

\bibitem[Soomro et~al.(2012)Soomro, Zamir, and Shah]{soomro2012ucf101}
Soomro, K., Zamir, A.~R., and Shah, M.
\newblock Ucf101: A dataset of 101 human actions classes from videos in the
  wild.
\newblock \emph{arXiv preprint arXiv:1212.0402}, 2012.

\bibitem[Sun et~al.(2021{\natexlab{a}})Sun, Karigo, Chakraborty, Mohanty, Wild,
  Sun, Chen, Anderson, Perona, Yue, et~al.]{sun2021multi}
Sun, J.~J., Karigo, T., Chakraborty, D., Mohanty, S.~P., Wild, B., Sun, Q.,
  Chen, C., Anderson, D.~J., Perona, P., Yue, Y., et~al.
\newblock The multi-agent behavior dataset: Mouse dyadic social interactions.
\newblock \emph{arXiv preprint arXiv:2104.02710}, 2021{\natexlab{a}}.

\bibitem[Sun et~al.(2021{\natexlab{b}})Sun, Kennedy, Zhan, Anderson, Yue, and
  Perona]{sun2020task}
Sun, J.~J., Kennedy, A., Zhan, E., Anderson, D.~J., Yue, Y., and Perona, P.
\newblock Task programming: Learning data efficient behavior representations.
\newblock In \emph{Proceedings of the IEEE/CVF Conference on Computer Vision
  and Pattern Recognition}, pp.\  2876--2885, 2021{\natexlab{b}}.

\bibitem[Sun et~al.(2019)Sun, Xiao, Liu, and Wang]{sun2019deep}
Sun, K., Xiao, B., Liu, D., and Wang, J.
\newblock Deep high-resolution representation learning for human pose
  estimation.
\newblock In \emph{Proceedings of the IEEE/CVF Conference on Computer Vision
  and Pattern Recognition}, pp.\  5693--5703, 2019.

\bibitem[Sun et~al.(2020)Sun, Kretzschmar, Dotiwalla, Chouard, Patnaik, Tsui,
  Guo, Zhou, Chai, Caine, et~al.]{sun2020scalability}
Sun, P., Kretzschmar, H., Dotiwalla, X., Chouard, A., Patnaik, V., Tsui, P.,
  Guo, J., Zhou, Y., Chai, Y., Caine, B., et~al.
\newblock Scalability in perception for autonomous driving: Waymo open dataset.
\newblock In \emph{Proceedings of the IEEE/CVF Conference on Computer Vision
  and Pattern Recognition}, pp.\  2446--2454, 2020.

\bibitem[Tong et~al.(2022)Tong, Song, Wang, and Wang]{tong2022videomae}
Tong, Z., Song, Y., Wang, J., and Wang, L.
\newblock Videomae: Masked autoencoders are data-efficient learners for
  self-supervised video pre-training.
\newblock \emph{arXiv preprint arXiv:2203.12602}, 2022.

\bibitem[Van~Horn et~al.(2021)Van~Horn, Cole, Beery, Wilber, Belongie, and
  Mac~Aodha]{van2021benchmarking}
Van~Horn, G., Cole, E., Beery, S., Wilber, K., Belongie, S., and Mac~Aodha, O.
\newblock Benchmarking representation learning for natural world image
  collections.
\newblock In \emph{Computer Vision and Pattern Recognition}, 2021.

\bibitem[Vaswani et~al.(2017)Vaswani, Shazeer, Parmar, Uszkoreit, Jones, Gomez,
  Kaiser, and Polosukhin]{vaswani2017attention}
Vaswani, A., Shazeer, N., Parmar, N., Uszkoreit, J., Jones, L., Gomez, A.~N.,
  Kaiser, {\L}., and Polosukhin, I.
\newblock Attention is all you need.
\newblock \emph{Advances in neural information processing systems}, 30, 2017.

\bibitem[Walker et~al.(2016)Walker, Doersch, Gupta, and
  Hebert]{walker2016uncertain}
Walker, J., Doersch, C., Gupta, A., and Hebert, M.
\newblock An uncertain future: Forecasting from static images using variational
  autoencoders.
\newblock In \emph{European Conference on Computer Vision}, pp.\  835--851.
  Springer, 2016.

\bibitem[Wei et~al.(2022)Wei, Fan, Xie, Wu, Yuille, and
  Feichtenhofer]{wei2022masked}
Wei, C., Fan, H., Xie, S., Wu, C.-Y., Yuille, A., and Feichtenhofer, C.
\newblock Masked feature prediction for self-supervised visual pre-training.
\newblock In \emph{Proceedings of the IEEE/CVF Conference on Computer Vision
  and Pattern Recognition}, pp.\  14668--14678, 2022.

\bibitem[Wiltschko et~al.(2015)Wiltschko, Johnson, Iurilli, Peterson, Katon,
  Pashkovski, Abraira, Adams, and Datta]{wiltschko2015mapping}
Wiltschko, A.~B., Johnson, M.~J., Iurilli, G., Peterson, R.~E., Katon, J.~M.,
  Pashkovski, S.~L., Abraira, V.~E., Adams, R.~P., and Datta, S.~R.
\newblock Mapping sub-second structure in mouse behavior.
\newblock \emph{Neuron}, 88\penalty0 (6):\penalty0 1121--1135, 2015.

\bibitem[Wiltschko et~al.(2020)Wiltschko, Tsukahara, Zeine, Anyoha, Gillis,
  Markowitz, Peterson, Katon, Johnson, and Datta]{wiltschko2020revealing}
Wiltschko, A.~B., Tsukahara, T., Zeine, A., Anyoha, R., Gillis, W.~F.,
  Markowitz, J.~E., Peterson, R.~E., Katon, J., Johnson, M.~J., and Datta,
  S.~R.
\newblock Revealing the structure of pharmacobehavioral space through motion
  sequencing.
\newblock \emph{Nature Neuroscience}, 23\penalty0 (11):\penalty0 1433--1443,
  2020.

\bibitem[Wu et~al.(2016)Wu, Nern, Williamson, Morimoto, Reiser, Card, and
  Rubin]{wu2016visual}
Wu, M., Nern, A., Williamson, W.~R., Morimoto, M.~M., Reiser, M.~B., Card,
  G.~M., and Rubin, G.~M.
\newblock Visual projection neurons in the drosophila lobula link feature
  detection to distinct behavioral programs.
\newblock \emph{Elife}, 5:\penalty0 e21022, 2016.

\bibitem[Yue et~al.(2014)Yue, Lucey, Carr, Bialkowski, and
  Matthews]{yue2014learning}
Yue, Y., Lucey, P., Carr, P., Bialkowski, A., and Matthews, I.
\newblock Learning fine-grained spatial models for dynamic sports play
  prediction.
\newblock In \emph{2014 IEEE international conference on data mining}, pp.\
  670--679. IEEE, 2014.

\bibitem[Zhan et~al.(2020)Zhan, Tseng, Yue, Swaminathan, and
  Hausknecht]{zhan2019learning}
Zhan, E., Tseng, A., Yue, Y., Swaminathan, A., and Hausknecht, M.
\newblock Learning calibratable policies using programmatic style-consistency.
\newblock \emph{ICML}, 2020.

\bibitem[Zhan et~al.(2021)Zhan, Sun, Kennedy, Yue, and
  Chaudhuri]{zhan2021unsupervised}
Zhan, E., Sun, J.~J., Kennedy, A., Yue, Y., and Chaudhuri, S.
\newblock Unsupervised learning of neurosymbolic encoders.
\newblock \emph{arXiv preprint arXiv:2107.13132}, 2021.

\end{thebibliography}
\bibliographystyle{icml2023}

\newpage
\appendix

\onecolumn

\section*{Appendix for MABe22}

Links to access our code and dataset, including code from challenge winners where available, are on our dataset website at \href{https://sites.google.com/view/computational-behavior/our-datasets/mabe2022-dataset}{https://sites.google.com/view/computational-behavior/our-datasets/mabe2022-dataset}. The sections of our appendix are organized as follows:
\begin{itemize}
    \item[\textbullet] \hyperref[appendix:community]{\textbf{\textcolor{blue}{Section A}}} contains details of community-contributed methods from our open competition.
    \item[\textbullet] \hyperref[appendix:datasheet]{\textbf{\textcolor{blue}{Section B}}} contains dataset documentation and intended uses for MABe2022, following the format of the Datasheet for Datasets\cite{gebru2018datasheets}.
    \item[\textbullet] \hyperref[appendix:dataprep]{\textbf{\textcolor{blue}{Section C}}} contains additional dataset details for mouse, fly, and beetle datasets.
    \item[\textbullet] \hyperref[sec:evaluation]{\textbf{\textcolor{blue}{Section D}}} shows the evaluation metrics for MABe2022, namely the F1 score and Mean Squared Error.
    \item[\textbullet] \hyperref[appendix:implementation_details]{\textbf{\textcolor{blue}{Section E}}} contains additional implementation details of our models.
    \item[\textbullet] \hyperref[sec:addtrajectory]{\textbf{\textcolor{blue}{Section F}}} provides additional evaluation results on the trajectory data of MABe22 (mouse and fly).
\end{itemize}

\textbf{Limitations and next steps.} 
Our dataset is mainly based on three species, and certainly does not saturate the variety of visually distinctive behavior phenomena that one encounters in the world.
Additionally, our dataset includes data from one lab per species/preparation and results may not translate to nominally identical preparations in other labs.
Future work to incorporate larger amounts of species as well as broader range of tasks can enable benchmark model rankings to be more predictive of method behavior on novel species and tasks.
Additionally, we limited our study of self-supervised video representation learning models to meaningful but not complete selection of state-of-the-art methods. This gap will be filled by the community if our benchmark is adopted to evaluate new methods.

\textbf{Broader impact.} While the "quality" of a learned representation will ultimately depend the downstream use, we provide a resource for the assessment of representation utility by scoring learned representations on a large array of hidden tasks, based on common scientific applications. 
We note that methods that perform best on our benchmark are not guaranteed to be the best choice for all possible downstream uses of representation learning. 
Depending on the downstream use, model developers may want to consider different choices of architecture, learning objective, and dataset to optimize for different properties of the representation.
Our goal is to provide a unified set of tasks across a range of behavior analysis settings that can enable quantitative comparison of representation learning methods, in order to facilitate research and method development for representation learning and behavior analysis. 
Additionally, we value any input from the community on MABe2022; you can reach us at \href{mailto:mabe.workshop@gmail.com}{mabe.workshop@gmail.com}.

\section{Community-Contributed Methods Descriptions}~\label{appendix:community}
We document methods from the community from our open challenge, which ran from February to July 2022. The challenge is available at: \href{https://www.aicrowd.com/challenges/multi-agent-behavior-challenge-2022}{https://www.aicrowd.com/challenges/multi-agent-behavior-challenge-2022}. We present the top 3 performing video and trajectory representation learning methods from the challenge winners, for each organism in MABe22. The video challenge uses the mouse and beetle datasets and the trajectory challenge uses the mouse and fly datasets. All methods had access to the train and validation sets during development, and the test set was held out (only released after challenge completion).

The results from the video challenge winners are fully in Section~\ref{sec:experiments} of the main paper. A subset of the trajectory results are also included in the same section. We note that most of the trajectory-only evaluations are in Section~\ref{sec:addtrajectory}.

\subsection{Video-based methods}~\label{appendix:community_video}
\begin{figure}[b]
  \begin{minipage}[c]{0.55\linewidth}
    \centering
    \includegraphics[width=\linewidth]{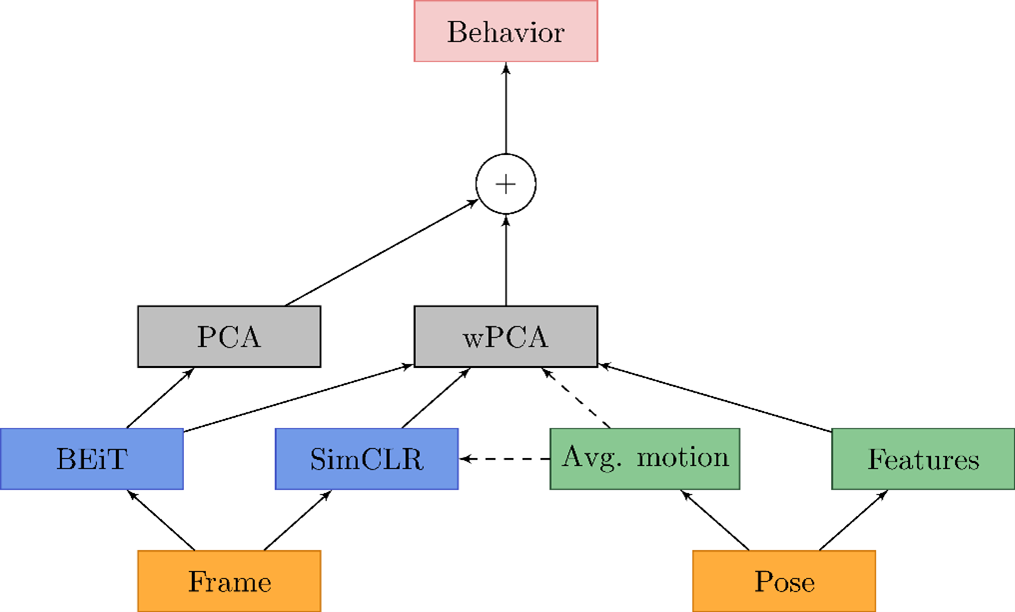}
  \end{minipage}
  \hfill
  \begin{minipage}[c]{0.4\linewidth}
    \caption{\textbf{BEiT + Hand-crafting Model Overview}. We learn a representation using both video and kepoint information, by (1) encoding video data through a pre-trained BEiT model, (2) learning visual representations using SimCLR, and (3) hand-crafted keypoint features.  
    }
    \label{fig:beit_model}
  \end{minipage}
\end{figure}
\subsubsection{BEiT + hand-crafting}
This method uses representations from both the video and trajectory datasets (Figure~\ref{fig:beit_model}). The three components of the model are:
\begin{itemize}
    \item A large vision transformer model (BEiT~\cite{bao2021beit} large, patch-16, 512x512), pre-trained on ImageNet22k~\cite{deng2009imagenet} at 224x224 pixels and fine-tuned on ImageNet1k at 512x512 pixels. We selected this model as it performs very well on ImageNet, and is one of the few such models that has been trained on images of size 512x512.
    \item A SimCLR~\cite{chen2020simple} model, based on the baseline implementation (\url{https://www.aicrowd.com/showcase/unsupervised-model-simclr-mouse-video-data}) but with three important modifications. First, the baseline augmentations were replaced with a version that used the keypoint annotations to constrain the random crops. Second, during training not all frames were sampled with equal probability. Instead, frames where the mice were mobile were sampled with a higher probability than frames where the mice were stationary. This weighting is intended to compensate for the fact that in some of the clips the mice are stationary (presumably sleeping) throughout the video, and as a result all the frames from those clips are highly similar. Third, the encoder was changed from ResNet-18 to ResNet-50, and only one frame was used as input instead of a sequence of frames.
    \item A number of hand-crafted features, based on the keypoint annotations and used in previous works, such as~\cite{segalin2020mouse,sun2020task}. These features consisted of measurements internal to each mouse (e.g. the distance between the nose and the tail), measurements that involve each mouse and the cage (e.g. the distance between the mouse and the nearest corner), and measurements involving more than one mouse (e.g. the distance between two mice and the area of the triangle formed by the three mice).
\end{itemize}

These three parts have complementary strengths, i.e., submissions based on each of them individually received high scores on different tasks. 
These features: 1024 (BEiT) + 2048 (SimCLR) + 214 (hand-crafted) = 3286 were concatenated for each frame, before PCA transforming them. We also found that we could improve results by reweighting both frame-wise and feature-wise before doing the PCA transform. Each frame was weighted by the measure of movement used in SimCLR training. Each of the three feature blocks was weighted by a numerical factor that was empirically determined.
Finally, we had noticed in an earlier experiment that some tasks benefited from including not only the PCA-transformed BEiT embedding for each frame, but also some PCA features averaged over the whole sequence. We therefore replaced the last 8 features of the above PCA embedding with the first 8 features from the PCA-transformed BEiT embedding, averaged over the sequence. For the beetle submission, we simply computed the BEiT features from each frame and reduced them to the allowed 128 features with a standard PCA transformation.

The code is available at \url{https://github.com/IRLAB-Therapeutics/mabe_2022}. For training, we used a batch size of 76, with an Adam optimizer and an initial learning rate of 3e-4, following a cosine annealing schedule. The image resolution used for the SimCLR model is 224x224.

\subsubsection{Vision Ensemble}
\begin{figure}
  \begin{minipage}[c]{0.4\linewidth}
    \centering
    \includegraphics[width=\linewidth]{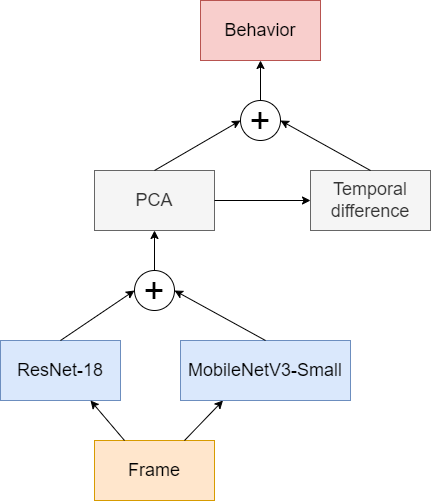}
  \end{minipage}
  \hfill
  \begin{minipage}[c]{0.55\linewidth}
    \caption{\textbf{Vision Ensemble Model Overview}. This model consists of features extracted from two pre-trained vision models, processed using a combination of temporal difference of the features as well as PCA.  
    }
    \label{fig:ensemble_model}
  \end{minipage}
\end{figure}
This model uses visual features only, extracted from pre-trained vision models (Figure~\ref{fig:ensemble_model}).
For both parts of the video challenge, we used an ensemble of pre-trained vision models by concatenating the output feature vectors of ResNet18~\cite{he2016deep} and MobileNetV3-Small~\cite{howard2019searching}, for which the size of the feature vector is respectively 512 and 574. This results in a vector of size 1086 which is subsequently reduced to size 128 by PCA, which forms the final embedding for the beetle dataset. For the mouse dataset, we reduced the size from 1086 down to 64, again by PCA. Subsequently, we concatenated to this the difference of the feature vector from 40 frames in the past and 40 frames in the future, i.e. a window size of 80. The length of the two concatenated vectors is then 128 and this forms the final representation for the mouse dataset.

\subsubsection{Multimodal MoCo/SimCLR}
\begin{figure}
  \begin{minipage}[c]{0.55\linewidth}
    \centering
    \includegraphics[width=\linewidth]{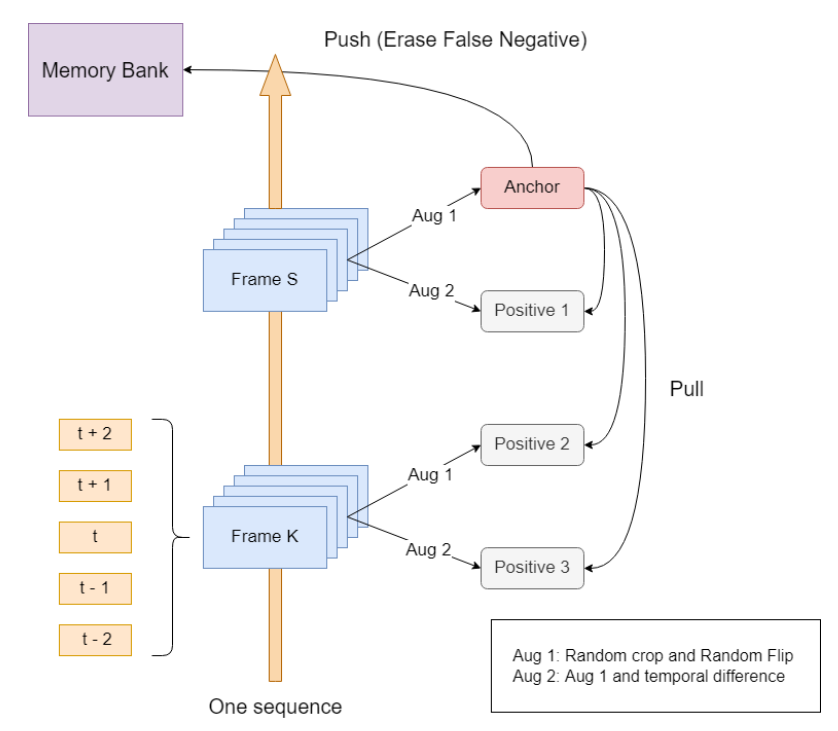}
  \end{minipage}
  \hfill
  \begin{minipage}[c]{0.4\linewidth}
    \caption{\textbf{Multimodal MoCo Model Overview}. We build a MoCo-based self-supervised learning framework
composed of a gradient updated encoder and a momentum updated encoder. This method is used for the beetle dataset, while a SimCLR-based method is used to extract features for the mouse dataset.
    }
    \label{fig:moco_model}
  \end{minipage}
\end{figure}
To leverage data from different modalities, we design different self-supervised methods for each modality individually (Figure~\ref{fig:moco_model}). We leverage three types of features, including visual features from video data, positional features and handcrafted features from keypoint data.
Inspired by MoCo~\cite{he2020momentum}, we build a self-supervised framework containing a memory bank to learn the visual features from video data. We use two types of augmentation strategies. The first augmentation strategy includes RandomResizeCrop, RandomHorizontalFlip, and RandomVerticalFlip. The second augmentation strategy includes the temporal difference in addition. 
As shown in Figure~\ref{fig:moco_model}, we sample two clips from a video and generate four views (two views for each clip). In the inference stage, we use the momentum updated encoder for a smooth result.

To learn the positional information of agents, we propose a generative task on keypoints data. Inspired by the MLM (Masked Language Modeling)~\cite{devlin2018bert} task in NLP, we propose the MPM (Masked Point Modeling) task, which is a frame-level task. The learning objective is to predict the masked keypoint coordinates based on observing unmasked keypoints. Giving a stream of agent-by-agent keypoint sequences, we randomly mask keypoint tokens at a ratio of by replacing keypoint tokens with mask tokens [MASK]. We then aggregate positional information from the rest frames with a vanilla Transformer~\cite{vaswani2017attention} encoder. Then a shallow decoder (i.e. a two-layer MLP) is used to predict the masked keypoint coordinates. We compute the reconstruction loss between the decoder output and original keypoint coordinates. Following our preliminary exploration, we find the representation generalizes better with MSE regularization than L1. Besides visual features and positional features encoded by the deep networks, we also utilize handcrafted features from keypoints data, including distances, angle, and speed. 

\textbf{Beetle Dataset}. We use the MoCo-based method to extract visual features from the ant-beetle video data. We first crop the regions with agents based on the keypoints. We resize images to 224x224 for training and inference. We random sample 2 clips with 7 frames from each video and the temporal stride is 5 frames. We use SGD optimizer and learning rate of 0.0075. The batch size is 128 per GPU and the weight decay is 1e-4. We set K=65536 for the memory bank and T=0.2 for the NT-Xent loss. We train for total 100k steps. The visual backbone is the pretrained Resnet101\_32x8d and the output dimension is 128.

\textbf{Mouse Dataset}. Different from ant-beetles video data, we utilize SimCLR~\cite{chen2020simple} to extract visual features from mouse video data. We directly regard other samples in the batch as negative samples instead of constructing a negative samples queue and an extra momentum updated encoder. We resize images to 224x224 at
training and 256x256 at inference. We random sample 3 clips with 7 frames from each video and the temporal stride is 12 frames. We use Adam optimizer and learning rate of 1e-4. The batch size is 64 per GPU and the weight decay is 1e-6. We train for total 100k steps. The visual backbone is ImageNet-1k pretrained Resnet50~\cite{he2016deep} and the output dimension is 128.

The keypoint coordinates of each agent are converted into a token by flatting and normalization, which results in 24-d input tokens. Then the tokens are fed into the main network. The encoder contains a 24-layer Transformer encoder and a projection head. Each Transformer layer has 768-d states and 12 masked attention heads. The one-layer projection head reduces the feature dimension from 768 to 128. At training, we sample 50 consecutive frames for each step. The learning rate and batch size per GPU are 1e-5 and 32 respectively. We use AdamW optimizer with betas of (0.9, 0.95) and weight decay of 0.1. We train for total 10k steps and warmup at the first 500 steps. We clip gradients with a norm threshold of 1.0. We do not adjust hyper-parameters. For handcrafted features, we calculate the following three types of features (59-d altogether): the mouse-mouse, mouses-wall, nose-tail, and nose-nose distances; the neck-base, nose-neck, and head-body angle of mice; the relative speed of the nose of the mice.
We concatenate the above three types of features and reduce the dimension to 128 by PCA. 

The code is available at \url{https://github.com/JiaHeng-DLUT/MABe2022}.

\subsection{Trajectory-based methods} ~\label{appendix:community_trajectory}


Our benchmark models learn from sequences of trajectory data and maps this data to a behavioral representation, which can then be used for a range of downstream tasks. Let $\mathcal{D}$ be a set of $N$ unlabelled trajectories. Each trajectory $\tau$ is a sequence of states $\tau = \{(s_t)\}_{t=1}^T$ over time, which represents the data for a variable number of agents across a variable number of timestamps. The state $s_i$ at timestamp $i$ corresponds to the location and pose of the agents at that time, often represented by keypoints. Let $\mathbf{z}$ be the behavioral representation. In our framework, models can learn from trajectories across time, but needs to produce a representation at each frame to account for frame-level tasks.

\begin{figure}
  \begin{minipage}[c]{0.6\linewidth}
    \centering
    \includegraphics[width=\linewidth]{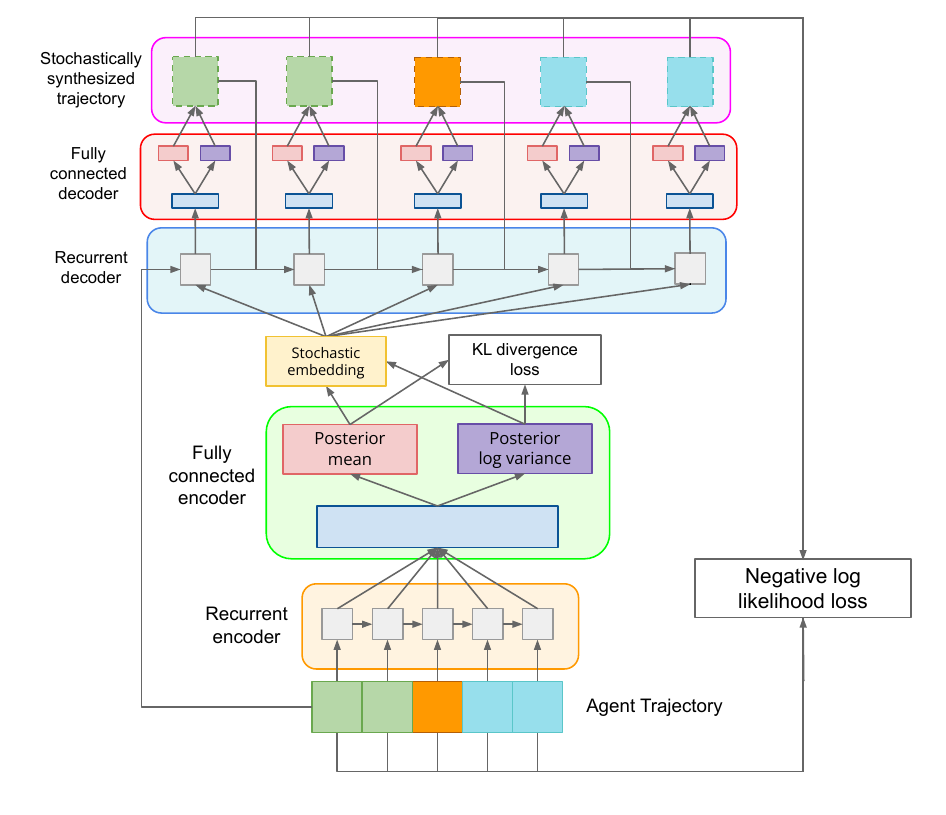}
  \end{minipage}
  \hfill
  \begin{minipage}[c]{0.3\linewidth}
    \caption{\textbf{TVAE Model Overview}. The TVAE learns a representation from trajectory data based on reconstructing the input trajectory. The encoder and decoder are based on recurrent neural networks.  
    }
    \label{fig:tvae_model}
  \end{minipage}
\end{figure}

\subsubsection{TVAE}

The Trajectory Variational Autoencoder (TVAE) is trained in a self-supervised way using trajectory reconstruction (Figure~\ref{fig:tvae_model}). 
To start, the keypoints of multiple agents is stacked to form the state at each timestamp $s_i$ for mouse, whereas the flies are encoded individually to handle the variable number of flies, and missing flies have zero-filled coordinates. The group fly embedding is created by concatenating the individual fly embeddings at each frame.

\textbf{Learning Objective}. The TVAE is a sequential generative model that uses trajectory reconstruction as the signal during training. Given previous states, the goal is to train the model to predict the next state. This architecture has previously been studied to learn trajectory representations in a variety of domains~\cite{co2018self,zhan2019learning,sun2020task}. We embed the input trajectory using an RNN encoder, $q_{\phi}$, and an RNN decoder, $p_{\theta}$, to predict the next state. The TVAE loss is:
%
\begin{equation}
\begin{aligned}
\mathcal{L}^{\text{tvae}} = \mathbb{E}_{q_{\phi}} \bigg[ \sum_{t=1}^T -\log(p_{\theta}(s_{t+1} | s_t, \mathbf{z})) \bigg] + D_{KL}(q_{\phi}(\mathbf{z} | \tau ) || p_{\theta}(\mathbf{z})).
\end{aligned}
\end{equation}
We use the unit Gaussian as a prior distribution $p_{\theta}(\mathbf{z})$ on $\mathbf{z}$ to regularize the learned embeddings.

To predict behavioral representations at each frame, we form a sliding window of size $21$, using $10$ frames before and after the current frame. The encoder and decoder are based on Gated Recurrent Units with 256 hidden layers. The training uses the Adam optimizer~\cite{kingma2014adam} with a batch size of 512 with learning rate $0.0002$.

The code is available at \href{https://github.com/AndrewUlmer/MABe_2022_TVAE}{https://github.com/AndrewUlmer/MABe\_2022\_TVAE}.

\begin{figure}
  \begin{minipage}[c]{0.75\linewidth}
    \centering
    \includegraphics[width=\linewidth]{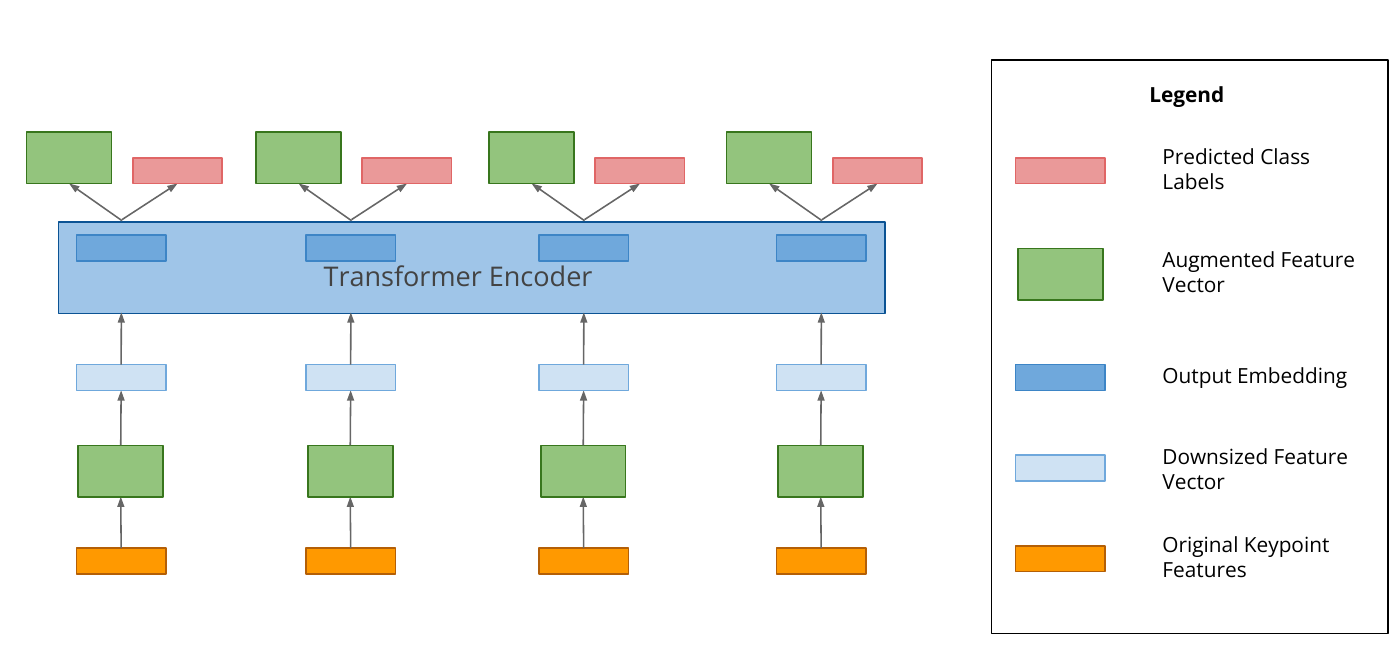}
  \end{minipage}
  \hfill
  \begin{minipage}[c]{0.23\linewidth}
    \caption{\textbf{T-Perceiver Model Overview}. We first compute high-dimensional hand-crafted features (``augmented feature vector") from the input trajectory data, then a Perceiver~\cite{jaegle2021perceiver} model is trained to predict features as well as public labels from the learned representation.
    }
    \label{fig:perceiver_model}
  \end{minipage}
\end{figure}

\subsubsection{T-Perceiver}

The T-Perceiver model (Figure~\ref{fig:perceiver_model}) has two main steps: (1) we first create a richer representation by augmenting keypoint coordinates with additional hand-crafted features and (2) then we learn the temporal relationships and extracted the embedding from a Perceiver model~\cite{jaegle2021perceiver}. The model is trained to reconstruct frame-level features from masked input as well as predict any public tasks.

\textbf{Hand-crafted feature extraction}. The first step transforms the original keypoint features into a high dimensional frame-level representation of the distances, angles and velocities between the keypoints. Feature extraction was performed algorithmically resulting in a 456 dimensional vector for the mouse dataset, and a 2112 dimensional vector for the fly dataset. The fly dataset has larger feature vectors as there were up to 11 individual flies in each frame, and when there were fewer flies, the vector was padded with zeros. Angles are encoded using $(\text{sin}(\theta)$, $\text{cos}(\theta))$. All features are normalized to have a mean of 0 and a standard deviation of 1.

\textbf{Sequence modeling}. The second step is to use an unsupervised sequence to sequence (seq2seq) model to combine these features across frames and map to the desired final embedding dimension. The features are first downsized to the final embedding size using a two layer fully-connected neural network with an intermediate layer size twice the size of the respective final embeddings using a $50\%$ dropout rate and the ELU activation function. This sequence of downsized features are passed through a standard Perceiver model~\cite{jaegle2021perceiver} with the number and dimension of latent vectors equal to embedding size and a sequence length of 512. 
For the fly dataset only, every second frame is dropped for computational reasons due to the high original frame rate.

\textbf{Learning Objective}. During training, a variable number of up to $80\%$ of frames were masked out and there was an additional linear layer to predict the original unmasked high dimensional features as well as labels from the public train split containing a subset of the hidden tasks. The model was trained to simultaneously optimize for two tasks: to minimize the mean square error on the frame-level features and to minimize the cross entropy loss of the label predictions. The first task was given a weight of 10 compared with the second task. The Adam optimizer~\cite{kingma2014adam} was used for training with a learning rate of 0.001.

The code is available at \href{https://colab.research.google.com/drive/13_M6yzF1VQ4STuJsO1at-GWK2_TDGTNV?usp=sharing}{https://colab.research.google.com/drive/13\_M6yzF1VQ4STuJsO1at-GWK2\_TDGTNV?usp=sharing}.

\begin{figure}
  \begin{minipage}[c]{0.65\linewidth}
    \centering
    \includegraphics[width=\linewidth]{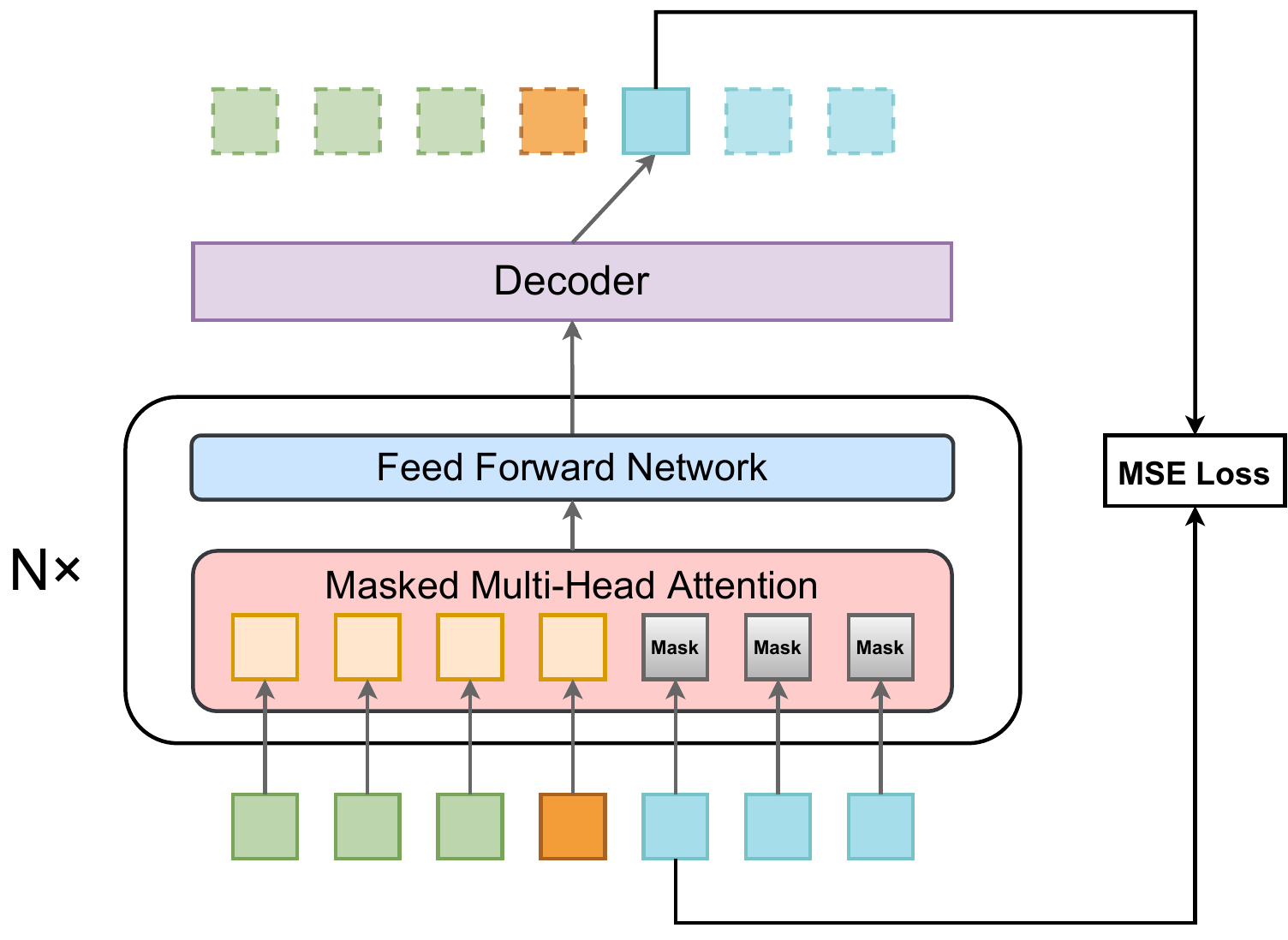}
  \end{minipage}
  \hfill
  \begin{minipage}[c]{0.3\linewidth}
    \caption{\textbf{T-GPT Model Overview}. The T-GPT uses a Transformer architecture~\cite{brown2020language} to predict keypoint coordinates in the next frame, given a representation learned from past frames. This prediction task is done both forwards and backwards in time.
    }
    \label{fig:gpt_model}
  \end{minipage}
\end{figure}

\subsubsection{T-GPT}

The T-GPT model is inspired by the NSP (Next Sentence Prediction)~\cite{bert} task from natural language processing. We instead propose the NFP (Next Frame Prediction) task, which is a frame-level task for predicting the keypoint coordinates in the next frame based on the observation of the past frames (Figure~\ref{fig:gpt_model}).

Giving a stream of frame-by-frame states $\tau = \{(s_t)\}_{t=1}^T$, we first aggregate information from past frames with a \emph{vanilla} Transformer encoder~\cite{vaswani2017attention} $f$:
\begin{equation}
\begin{aligned}
\mathbf{z}_t = f(s_1, s_2, ..., s_t)
\end{aligned}
\end{equation}
Then a shallow decoder \emph{h} (i.e. a two-layer MLP) is used to predict the keypoint coordinates in the next frame:
\begin{equation}
\begin{aligned}
\hat{s}_{t+1} = h(\mathbf{z}_t)
\end{aligned}
\end{equation}
\textbf{Learning Objective}. We compute the reconstruction loss between the decoder output and original keypoint coordinates:
\begin{equation}
\begin{aligned}
\mathcal{L} = MSE(\hat{s}_{t+1}, s_{t+1})
\end{aligned}
\end{equation}
Following our preliminary exploration, we find the representation generalizes better with MSE loss than L1.

We build on the open source implementation of GPT~\cite{brown2020language}. First, the keypoint coordinates in each frame are converted into a token by flatting and normalization, which results in 528-d input tokens.
Then the tokens are fed into the encoder network, with a 24-layer Transformer encoder and a projection head. Each Transformer layer has 768 dimensional states and 12 masked attention heads. The one-layer projection head reduces the feature dimension from 768 to 256 for flies and 128 for mice. A two-layer decoder (Linear-LayerNorm-Tanh-Linear) is used to predict the coordinates in the next frame. In order to only attends to the left context, we use the upper triangular matrix attention mask in each self-attention layer when training. In the inference stage, these masks are removed to better aggregate contextual features from the past and future. 

At training, we use all the available data and sample 50 consecutive frames each iteration. We randomly flip the coordinates horizontally with a probability of 0.5. The learning rate and batch size are 1e-5 and 2 respectively, with the AdamW optimizer~\cite{adamw}. To make better use of the training data, we do the NFP task in a bidirectional way and the corresponding losses are averaged. 

The code is available at \href{https://drive.google.com/drive/folders/1zcZ9lqtf0y4OCtfFdA1S7K3beLcXa-3e?usp=sharing}{https://drive.google.com/drive/folders/1zcZ9lqtf0y4OCtfFdA1S7K3beLcXa-3e?usp=sharing}

\begin{figure}
  \begin{minipage}[c]{0.75\linewidth}
    \centering
    \includegraphics[width=\linewidth]{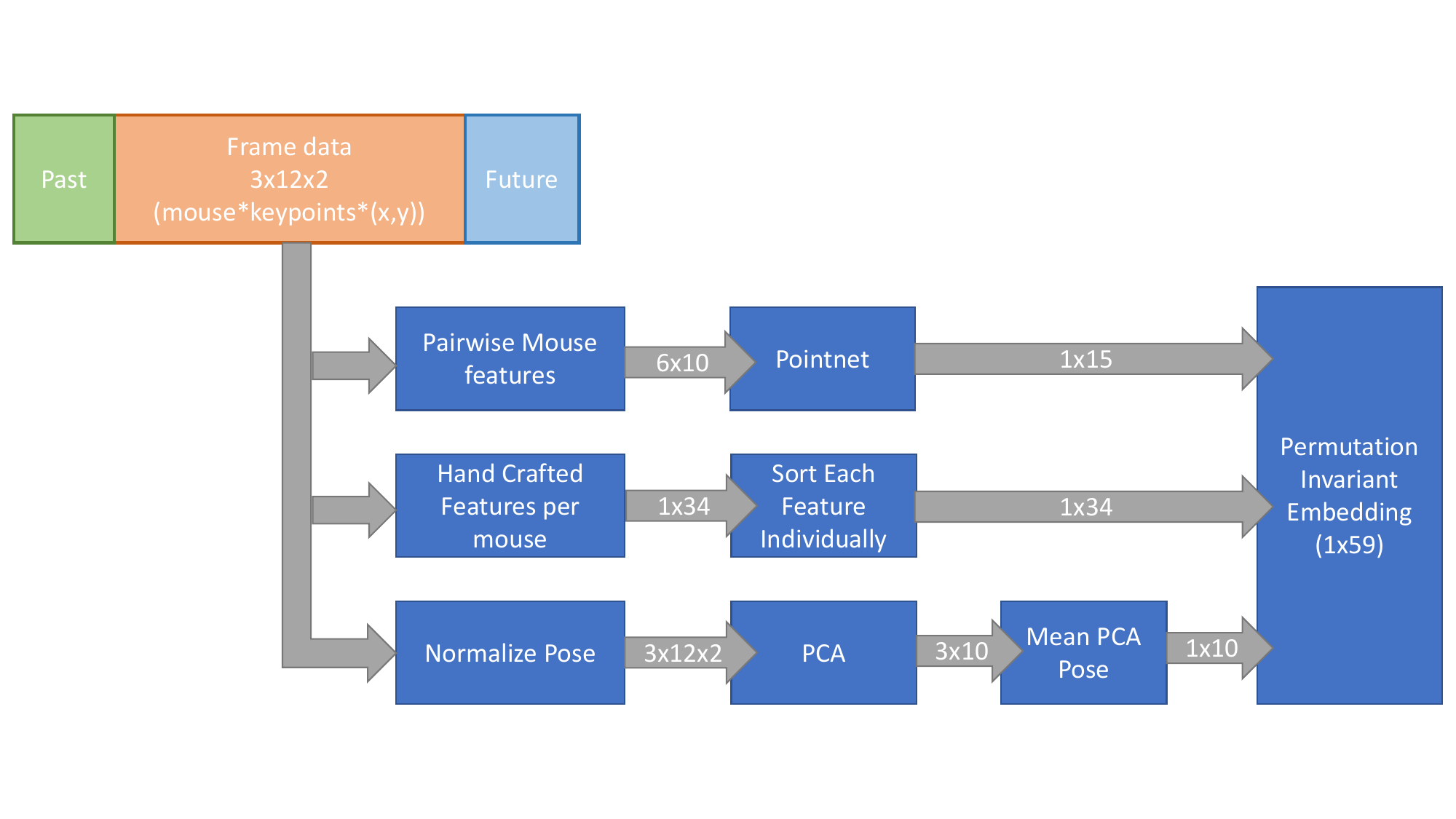}
  \end{minipage}
  \hfill
  \begin{minipage}[c]{0.23\linewidth}
    \caption{\textbf{T-PointNet Model Overview}. We combine hand-crafted features, PCA of pose keypoints, and PointNet~\cite{qi2017pointnet} embeddings as a permutation-invariant representation of the agents at each frame.
    }
    \label{fig:pointnet_model}
  \end{minipage}
\end{figure}

\subsubsection{T-PointNet}

We use PointNet~\cite{qi2017pointnet} alongside hand-crafted features and PCA to extract permutation-invariant features from the keypoint data (Figure~\ref{fig:pointnet_model}). As the embedding will be used to train a network for the hidden tasks, its important that embedding vector remains same even the order of the mice is switched. We note that this model is only applied to the mouse data, and not to the fly data, where some of the tasks are fly-dependent.

The hand-crafted features used are similar to the ones from~\cite{sun2020task}, and 10 PCA components are computed for each mice and averaged to generate the group embeddings. Based on the goal of generating permutation-invariant embeddings, we select a PointNet based architecture~\cite{qi2017pointnet}, which has been popular for learning patterns in unordered point cloud datasets. It fundamentally relies on commutative operations like sum, average, max to create permutation invariant features. 

\textbf{Learning Objective}. Each ``point'' fed into PointNet represents one pair of agents, and the coordinates are hand crafted features between each pair such as distance, angle, and speed (each 10 dimensions). PointNet is trained using a cosine similarity loss, where nearby frames in a sequence are treated as positives whereas a random frame chosen from a random sequence is chosen as negatives.
The advantage of this network is that the
embedding remain same regardless of the input order of the
agents. The final combined embedding size is 69 dimensions.

We use the vanilla PointNet network as described by authors in~\cite{qi2017pointnet} with a reduced set of parameters and filters. The original network is designed for point clouds in order of 1000 and in contrast, in this application there are only 6 animal pairs corresponding to 6 points, thus
we reduce the network capacity to prevent overfitting. This model is trained with an Adam optimizer~\cite{kingma2014adam} with learning rate $0.005$ and batch size 512.

The code is available at \href{https://github.com/Param-Uttarwar/mabe_2022
}{https://github.com/Param-Uttarwar/mabe\_2022
}.

\begin{figure}
  \begin{minipage}[c]{0.75\linewidth}
    \centering
    \includegraphics[width=\linewidth]{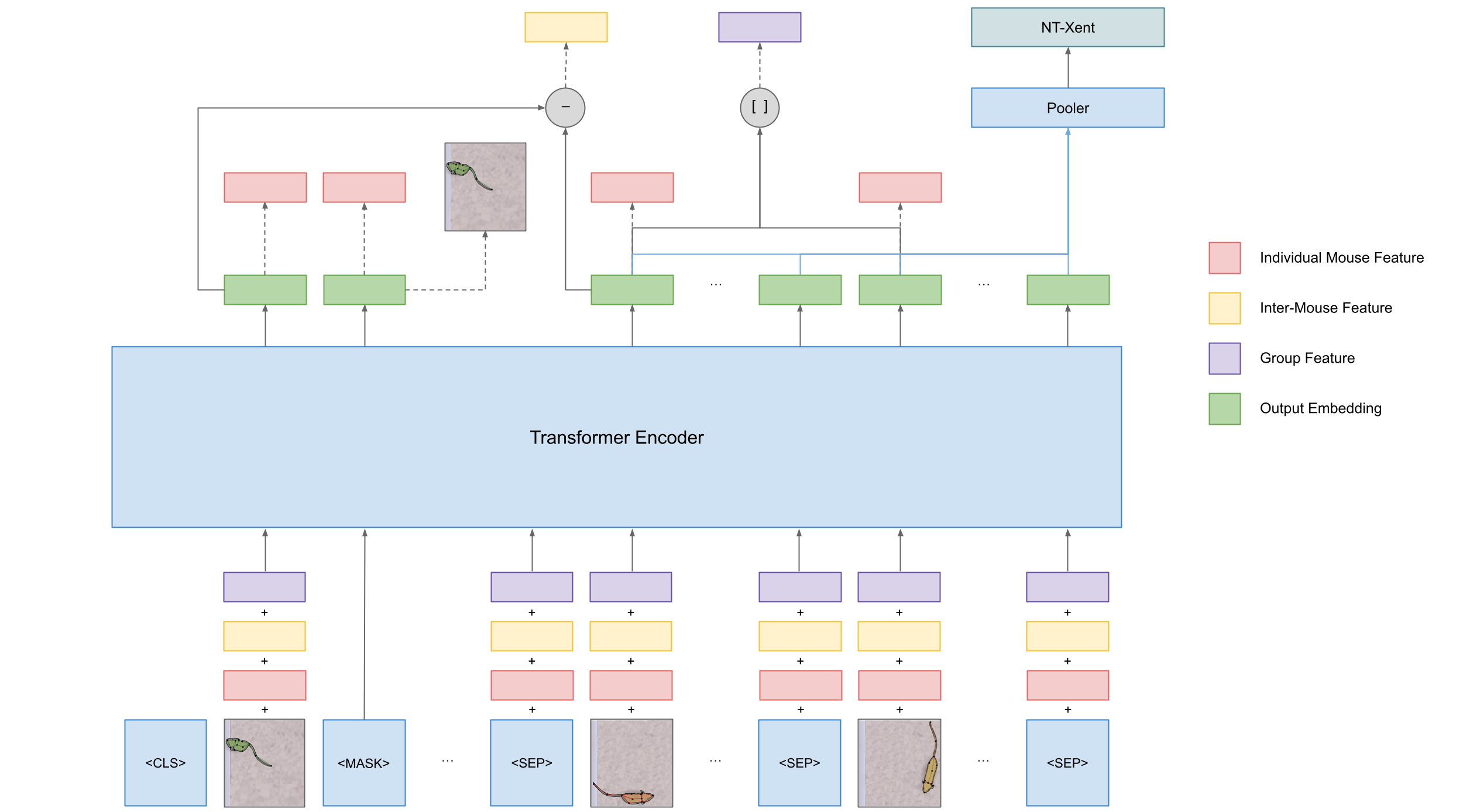}
  \end{minipage}
  \hfill
  \begin{minipage}[c]{0.23\linewidth}
    \caption{\textbf{T-BERT Model Overview}. The trajectory of each agent is concatenated and encoded using BERT~\cite{bert}, trained on masked modeling, predicting hand-crafted features, contrastive loss, and predicting publicly available train tasks. 
    }
    \label{fig:BERT_model}
  \end{minipage}
\end{figure}

\subsubsection{T-BERT}

We extend BERT~\cite{bert} to learn separate embeddings for each agent in the enclosure which are then concatenated for the group embedding (Figure~\ref{fig:BERT_model}). We train the model using three main tasks: 1) Masked modelling, 2) hand-crafted feature predictions similar to that of \cite{sun2020task}, and 3) contrastive learning. This model is only applied to the mouse dataset.

We sample a window of 80 frames, encoding the keypoints with a linear projection layer. The sequence of keypoints for each agent is separated by a special learned embedding, similar to a [SEP] token~\cite{bert}. We use three different kinds of features: 1) Individual-agent features, which are agent specific. 2) Inter-agent, which are features between each pair of agents. Note that these pairings can be directional. 3) Group features which apply to the entire group. Each feature type is encoded and their embeddings are added. In the case of inter-agent features, we encode and add each pair.

\textbf{Masked Modeling}. We mask 70\% of the input keypoints and features. Because of the high sampling frequency of the dataset, masked modelling may be trivial through interpolation of nearby frames. We therefore mask spans of the input, following the same masking scheme as SpanBERT~\cite{spanbert}. We set the minimum and maximum span length to 3 and 20 respectively and sample lengths according to $l \sim Geo(p=0.2)$. The input subsequence is encoded with a stack of 12 transformer self-attention blocks with hidden size $912$, followed by a projection onto a $42$ dimensional space for the output embeddings. We apply dropout to these and predict the normalized masked keypoints.

\textbf{Feature Predictions} We predict individual-agent features directly from each output embedding. Inter-agent features are predicted by taking the output embeddings for the agents in the pair and subtracting them, then regressing the features from this pair embedding. We obtain the final representation for the group by concatenating the embeddings for each agent. We use the group embedding to predict group features and for the final submission. Group embeddings are pooled across frames using a weighted average to get a single embedding for the entire input sequence. This pooled embedding is then used for the contrastive task.

\textbf{Contrastive Task} We perform a contrastive learning task by taking two randomly subsequences from the same 1 minute clip as the positive pair. Negative pairs are created by pairing with other sequences within the batch. We encode the pooled sequence representation using a 2 layer MLP onto a 42 dimensional space. We take the NT-Xent loss~\cite{chen2020simple} with $\tau=0.1$.

\textbf{Learning Objectives}.  The task losses are weighted:
\begin{equation}
    L = L_m + 0.8L_x + 0.8L_y + 0.4L_z + 0.05L_c + 0.1L_{cl}
\end{equation}

Where $m$ is masked modelling, $x$ is the individual agent feature prediction task, $y$ is the inter-agent feature prediction task, $z$ is the group feature prediction task, $c$ is the chases task (public task on mouse dataset) and $cl$ is the contrastive task. 53 individual agent features were computed for each agent, with 13 inter-agent features for pairs, and 1 group feature for all three mice. Features concerning distances, velocities and accelerations are normalised by mouse length. Angles are encoded $(\text{sin}(\theta)$, $\text{cos}(\theta))$. We apply rotation, reflection and adding gaussian noise to the keypoints~\cite{sun2020task}, each are applied with probability $p=0.5$. To create frame-level embeddings for a 1 minute sequence, we encode overlapping $80$ frame windows of the input using a stride of $40$ frames. 

An exhaustive hyperparameter search was not possible due to computational constraints, so most parameters were not tuned. We tested input lengths of $60$, $80$ and $100$ frames and found that $80$ was optimal.
We split the dataset into training and validation splits, with 95\% and 5\% respectively. 
We train the model for $160$ epochs with a batch size of $16$. We used AdamW~\cite{adamw} with a learning rate of $0.00003$ and a linear schedule. The model with the lowest validation loss is chosen.

The code is available at \href{https://github.com/edhayes1/MABe}{https://github.com/edhayes1/MABe}

\section{Datasheets}~\label{appendix:datasheet}

\subsection{Mouse Datasheet}~\label{appendix:mouse_datasheet}


\dssectionheader{Motivation}

\dsquestionex{For what purpose was the dataset created?}{Was there a specific task in mind? Was there a specific gap that needed to be filled? Please provide a description.}

\dsanswer{Automated animal pose estimation has become an increasingly popular tool in the neuroscience community, fueled by the publication of several easy-to-train animal pose estimation systems. Building on these pose estimation tools, pose-based approaches to supervised or unsupervised analysis of animal behavior are currently an area of active research. New computational approaches for automated behavior analysis are probing the detailed temporal structure of animal behavior, its relationship to the brain, and how both brain and behavior are altered in conditions such as Parkinson’s, PTSD, Alzheimer’s, and autism spectrum disorders. Due to a lack of publicly available animal behavior datasets, most new behavior analysis tools are evaluated on their own in-house data. There are no established community standards by which behavior analysis tools are evaluated, and it is unclear how well available software can be expected to perform in new conditions, particularly in cases where training data is limited. Labs looking to incorporate these tools in their experimental pipelines therefore often struggle to evaluate available analysis options, and can waste significant effort training and testing multiple systems without knowing what results to expect.

The Multi-Agent Behavior 2022 (MABe22) dataset is a new set of animal tracking, pose, video, and behavior datasets, intended to serve as a benchmark dataset for evaluation of unsupervised/self-supervised behavior representation learning and discovery methods. This datasheet is specific to the Mouse Triplets dataset, which consists of snippets of video and trajectory data from triplets of interacting mice. Accompanying the data is a collection of 8 "hidden labels": for each video frame of the dataset, we provide annotations of animal strain, time of day, light cycle, and a set of behaviors. These hidden labels can be used to evaluate the quality of learned representations of animal behavior, by asking how well the information they represent can be decoded from a given representation.
}

\dsquestion{Who created this dataset (e.g., which team, research group) and on behalf of which entity (e.g., company, institution, organization)?}

\dsanswer{The MABe22 Mouse Triplets dataset was collected and analyzed in the laboratory of Vivek Kumar at Jackson Labs (JAX), and was assembled by Ann Kennedy at Northwestern University. Mice were bred and videos of interacting mice were collected by Tom Sproule at JAX. The video dataset was tracked by Brian Geuther and Keith Sheppard at JAX, with pose estimation performed using a modified version of HRnet described in~\cite{sheppard2020gait}. Tracking and video data were screened for tracking quality and segmented into one-minute "sequences" by Ann Kennedy. Sequences were manually annotated for four social behaviors of interest by Markus Marks.
}

\dsquestionex{Who funded the creation of the dataset?}{If there is an associated grant, please provide the name of the grantor and the grant name and number.}

\dsanswer{Acquisition of behavioral data was supported by NIH grants DA041668 (NIDA), DA048034 (NIDA), and Simons Foundation SFARI Director's Award (to VK). Curation of data task design was funded by NIMH award \#R00MH117264 (to AK) and NSERC Award \#PGSD3-532647-2019 (to JJS).
}

\dsquestion{Any other comments?}

\dsanswer{None.}

\bigskip
\dssectionheader{Composition}

\dsquestionex{What do the instances that comprise the dataset represent (e.g., documents, photos, people, countries)?}{ Are there multiple types of instances (e.g., movies, users, and ratings; people and interactions between them; nodes and edges)? Please provide a description.}

\dsanswer{The core element of this dataset, called a \textit{sequence}, consists of raw video, tracked postures, sequence-level experimental conditions, and hand-scored actions of three mice interacting in a 52 cm x 52 cm arena, filmed from above at 30 Hz. All three mice are adult males from the same strain, either C57Bl/6J or BTBR. Postures of animals are estimated in terms of a set of twelve anatomically defined "keypoints" that capture the detailed two-dimensional pose of the animal. Because the three mice are not easily distinguished, temporal filtering methods are used to track the identity of animals across frames. Because both of these processing steps are automated, some errors in pose estimation or swaps of mouse identity do occur in the dataset.

Accompanying each sequence are frame-by-frame annotations for 8 "hidden tasks" capturing experimental conditions, animal background, and animal behavior. The 8 hidden tasks for this dataset include four "sequence-level" tasks where annotation values are the same for all frames in a one-minute sequence, and nine "frame-level" tasks where annotation values vary from frame to frame. Descriptions of each task are provided in Table~\ref{tab:mouse_tasks}; all behaviors are defined between any given pair of animals.

The core element of a \textit{sequence} is called a \textit{frame}; this refers to the posture of the three animals on a particular frame of video, as well as annotations for the 8 hidden tasks.

\begin{table}[t]
\centering
\begin{tabular}{ |c|c|c|p{8cm}| } 
 \hline
 Task Name & Type & Values & Description \\ 
 \hline
 \hline
 Experiment day & Sequence & 1-4 & Mice were filmed interacting for four days after introduction to a new arena; task is to determine which day a sequence comes from. \\ 
 \hline
 Time of day & Sequence & 0-1440 & Mice show circadian changes in their level of activity; task is to infer time of day from behavior. \\
 \hline
 Strain & Sequence & 0 or 1 & Mice are from either C57Bl/6J or BTBR genetic background. Strain field is 1 for BTBR and 0 for C57Bl/6J. \\
 \hline
 Lights & Sequence & 0 or 1 & Mice are more active when the lights are off, which occurs between 6am and 6pm; task is to infer light condition from behavior.\\
 \hline
 Chase & Frame & 0 or 1 & A pair of mice moving quickly with one mouse following close behind the other.\\
 \hline 
 Huddle & Frame & 0 or 1 & Bodies of the mice are in close contact and the animals are stationary for at least several seconds; can occur between either pairs or triplets of animals.\\
 \hline
 Face sniffing & Frame & 0 or 1 & A close-investigation behavior in which the nose of one mouse is in close contact with the nose or face of another mouse.\\
 \hline
 Anogenital sniffing & Frame & 0 or 1 & A close-investigation behavior in which one mouse is investigating the anogenital area of another, typically with its nose near the base of the tail or pushed underneath the hindquarters of the other animal.\\
 \hline
 \hline
\end{tabular}
\caption{Format of hidden tasks for mouse dataset.}\label{tab:mouse_all_tasks}
\end{table}
}

\dsquestion{How many instances are there in total (of each type, if appropriate)?}

\dsanswer{
This dataset is composed of 2614 one-minute-long sequences filmed at 30 Hz. 

}

\dsquestionex{Does the dataset contain all possible instances or is it a sample (not necessarily random) of instances from a larger set?}{ If the dataset is a sample, then what is the larger set? Is the sample representative of the larger set (e.g., geographic coverage)? If so, please describe how this representativeness was validated/verified. If it is not representative of the larger set, please describe why not (e.g., to cover a more diverse range of instances, because instances were withheld or unavailable).}

\dsanswer{The dataset is derived from a larger experiment, in which three mice were allowed to freely interact in an open arena for a period of four days. To generate the trajectories used for this dataset, we randomly sampled up to five one-minute intervals from each recorded hour of approximately 12 such four-day experiments. In initial sampling, we observed that during the lights-on phase of the light/dark cycle the mice spent the majority of the time huddled together sleeping. As this does not generate particularly interesting behavioral data, we randomly discarded 80\% of sampled one-minute intervals in which no substantial movement of the animals occurred, and replaced these with substitute samples drawn from the same one-hour time period. If after five attempts we could not randomly draw a replacement sample containing movement, we omitted the trajectory from the dataset. As a result, the dataset contains a higher proportion of trajectories with movement than is present in the source videos, and a slightly lower proportion of trajectories sampled from the light portion of the light/dark cycle.
}

\dsquestionex{What data does each instance consist of? “Raw” data (e.g., unprocessed text or images) or features?}{In either case, please provide a description.}

\dsanswer{Each sequence has three elements. 1) \textit{Keypoints} are the locations of twelve body parts on each mouse: the nose tip, left and right ears, base of neck, body centroid, base, middle, and tip of tail, and the four paws. Keypoints are estimated using a modified version of HRnet documented in ~\cite{sheppard2020gait}. 2) \textit{Annotations} are sequence-level or frame-level labels of experimental conditions or animal's actions. Definitions of these annotations are provided in Table~\ref{tab:mouse_tasks}. The behavior labels were generated using a series of short scripts based on features of detected animal poses; it is therefore possible that some mis-identification of behaviors occurs.

Note that this dataset does not include the original raw videos from which pose estimates were produced. This is because the objective of releasing this dataset was to determine the accuracy with which animal behavior could be detected using tracked keypoints alone.
}

\dsquestionex{Is there a label or target associated with each instance?}{If so, please provide a description.}

\dsanswer{Yes: each annotation (as described above) is provided for every frame in the dataset.
}

\dsquestionex{Is any information missing from individual instances?}{If so, please provide a description, explaining why this information is missing (e.g., because it was unavailable). This does not include intentionally removed information, but might include, e.g., redacted text.}

\dsanswer{There is no missing data.
}

\dsquestionex{Are relationships between individual instances made explicit (e.g., users’ movie ratings, social network links)?}{If so, please describe how these relationships are made explicit.}

\dsanswer{Each instance (\textit{sequence}) is to be treated as an independent observation with no relationship to other instances in the dataset. Although the identities of the interacting animals are the same in some sequences, this information is not tracked in the dataset.
}

\dsquestionex{Are there recommended data splits (e.g., training, development/validation, testing)?}{If so, please provide a description of these splits, explaining the rationale behind them.}

\dsanswer{The dataset includes a recommended train/test split which was used for the Multi-Agent Behavior Challenge. Data was randomly split into training, test, and private-test sets (where the private test set was withheld from challenge evaluation until the end of the competition period, to avoid overfitting.)
}

\dsquestionex{Are there any errors, sources of noise, or redundancies in the dataset?}{If so, please provide a description.}

\dsanswer{Pose keypoints in this dataset are produced using automated pose estimation software. The dataset was screened to remove sequences with poor pose estimation, detected as large jumps in the detected location of an animal, however some errors in pose estimation, missing keypoints, and noise in keypoint placement still occur. These are most common on frames when the two animals are in close contact or moving very quickly.

Frame-by-frame annotations of behavior were generated using a series of scripts that were manually tuned by a human expert. Pose estimation errors can contribute to missed bouts or false positives for behaviors in these annotations.
}

\dsquestionex{Is the dataset self-contained, or does it link to or otherwise rely on external resources (e.g., websites, tweets, other datasets)?}{If it links to or relies on external resources, a) are there guarantees that they will exist, and remain constant, over time; b) are there official archival versions of the complete dataset (i.e., including the external resources as they existed at the time the dataset was created); c) are there any restrictions (e.g., licenses, fees) associated with any of the external resources that might apply to a future user? Please provide descriptions of all external resources and any restrictions associated with them, as well as links or other access points, as appropriate.}

\dsanswer{The dataset is self-contained.
}

\dsquestionex{Does the dataset contain data that might be considered confidential (e.g., data that is protected by legal privilege or by doctor-patient confidentiality, data that includes the content of individuals non-public communications)?}{If so, please provide a description.}

\dsanswer{No.
}

\dsquestionex{Does the dataset contain data that, if viewed directly, might be offensive, insulting, threatening, or might otherwise cause anxiety?}{If so, please describe why.}

\dsanswer{No.
}

\dsquestionex{Does the dataset relate to people?}{If not, you may skip the remaining questions in this section.}

\dsanswer{No.
}

\dsquestionex{Does the dataset identify any subpopulations (e.g., by age, gender)?}{If so, please describe how these subpopulations are identified and provide a description of their respective distributions within the dataset.}

\dsanswer{n/a
}

\dsquestionex{Is it possible to identify individuals (i.e., one or more natural persons), either directly or indirectly (i.e., in combination with other data) from the dataset?}{If so, please describe how.}

\dsanswer{n/a
}

\dsquestionex{Does the dataset contain data that might be considered sensitive in any way (e.g., data that reveals racial or ethnic origins, sexual orientations, religious beliefs, political opinions or union memberships, or locations; financial or health data; biometric or genetic data; forms of government identification, such as social security numbers; criminal history)?}{If so, please provide a description.}

\dsanswer{n/a
}

\dsquestion{Any other comments?}

\dsanswer{
None.}

\bigskip
\dssectionheader{Collection Process}

\dsquestionex{How was the data associated with each instance acquired?}{Was the data directly observable (e.g., raw text, movie ratings), reported by subjects (e.g., survey responses), or indirectly inferred/derived from other data (e.g., part-of-speech tags, model-based guesses for age or language)? If data was reported by subjects or indirectly inferred/derived from other data, was the data validated/verified? If so, please describe how.}

\dsanswer{\textit{Sequences} in the dataset are derived from video of triplets of socially interacting mice in an open arena. Video data was processed to extract pose estimates and track identity of the animals, and to generate automated annotations of several behaviors of interest, included in the hidden labels in this dataset.
}

\dsquestionex{What mechanisms or procedures were used to collect the data (e.g., hardware apparatus or sensor, manual human curation, software program, software API)?}{How were these mechanisms or procedures validated?}

\dsanswer{Behavioral data was collected in the JAX Animal Behavior System ~\cite{beane2022video}. Videos were recorded using Basler acA1300-75gm camera with Tamron 4-12mm lens and 800nm longpass filter, at a framerate of 30Hz and camera resolution of 800 x 800 pixels. The camera was mounted 105+/-5 cm above the floor of an open field measuring 52cm x 52cm; a grate located at the northern wall of the arena provides animals access to food and water. Animals were introduced to the arena one by one over the first ten minutes of recording, and were recorded continuously for four days.

Pose estimation was performed using a modified version of HRnet documented in ~\cite{sheppard2020gait}. Manual annotation of animal behavior was performed by a trained human expert using the VIA video annotator~\cite{dutta2019via}.
}

\dsquestion{If the dataset is a sample from a larger set, what was the sampling strategy (e.g., deterministic, probabilistic with specific sampling probabilities)?}

\dsanswer{
Repeated from a previous section: to generate the trajectories used for this dataset, we randomly sampled up to five one-minute intervals from each recorded hour of approximately 12 such four-day experiments. In initial sampling, we observed that during the lights-on phase of the light/dark cycle the mice spent the majority of the time huddled together sleeping. As this does not generate particularly interesting behavioral data, we randomly discarded 80\% of sampled one-minute intervals in which no substantial movement of the animals occurred, and replaced these with substitute samples drawn from the same one-hour time period. If after five attempts we could not randomly draw a replacement sample containing movement, we omitted the trajectory from the dataset. As a result, the dataset contains a higher proportion of trajectories with movement than is present in the source videos, and a slightly lower proportion of trajectories sampled from the light portion of the light/dark cycle.
}

\dsquestion{Who was involved in the data collection process (e.g., students, crowdworkers, contractors) and how were they compensated (e.g., how much were crowdworkers paid)?}

\dsanswer{Behavioral data collection was performed by graduate student, postdoc, and technician members of the Kumar lab at Jackson Laboratories, as a part of another ongoing research project studying animal gait and behavior. (No videos or annotations were explicitly generated for this dataset release.) Lab members are full-time employees of Jackson Labs, and their compensation was not dependent on their participation in this project. Manual annotation of animal behavior was performed by Markus Marks, who is a full-time employee of Caltech and whose compensation was also not dependent on participation in this project.
}

\dsquestionex{Over what timeframe was the data collected? Does this timeframe match the creation timeframe of the data associated with the instances (e.g., recent crawl of old news articles)?}{If not, please describe the timeframe in which the data associated with the instances was created.}

\dsanswer{Source experiments associated with this dataset were performed in 2019, with pose estimation performed in 2019-2020 and manual annotation performed in Sept-Nov 2022. This dataset was assembled from December 2022 - March 2023.
}

\dsquestionex{Were any ethical review processes conducted (e.g., by an institutional review board)?}{If so, please provide a description of these review processes, including the outcomes, as well as a link or other access point to any supporting documentation.}

\dsanswer{All experiments included here were performed in accordance with NIH guidelines and approved by the Institutional Animal Care and Use Committee (IACUC) and Institutional Biosafety Committee at Jackson Labs. Review of experimental design by the IACUC follows the steps outlined in the NIH-published \href{https://grants.nih.gov/grants/olaw/Guide-for-the-Care-and-Use-of-Laboratory-Animals.pdf}{Guide for the Care and Use of Laboratory Animals}. All individuals performing behavioral experiments underwent animal safety training prior to data collection. Animals were maintained under close veterinary supervision.
}

\dsquestionex{Does the dataset relate to people?}{If not, you may skip the remaining questions in this section.}

\dsanswer{No.
}

\dsquestion{Did you collect the data from the individuals in question directly, or obtain it via third parties or other sources (e.g., websites)?}

\dsanswer{n/a
}

\dsquestionex{Were the individuals in question notified about the data collection?}{If so, please describe (or show with screenshots or other information) how notice was provided, and provide a link or other access point to, or otherwise reproduce, the exact language of the notification itself.}

\dsanswer{n/a
}

\dsquestionex{Did the individuals in question consent to the collection and use of their data?}{If so, please describe (or show with screenshots or other information) how consent was requested and provided, and provide a link or other access point to, or otherwise reproduce, the exact language to which the individuals consented.}

\dsanswer{n/a
}

\dsquestionex{If consent was obtained, were the consenting individuals provided with a mechanism to revoke their consent in the future or for certain uses?}{If so, please provide a description, as well as a link or other access point to the mechanism (if appropriate).}

\dsanswer{n/a
}

\dsquestionex{Has an analysis of the potential impact of the dataset and its use on data subjects (e.g., a data protection impact analysis) been conducted?}{If so, please provide a description of this analysis, including the outcomes, as well as a link or other access point to any supporting documentation.}

\dsanswer{n/a
}

\dsquestion{Any other comments?}

\dsanswer{None.
}

\bigskip
\dssectionheader{Preprocessing/cleaning/labeling}

\dsquestionex{Was any preprocessing/cleaning/labeling of the data done (e.g., discretization or bucketing, tokenization, part-of-speech tagging, SIFT feature extraction, removal of instances, processing of missing values)?}{If so, please provide a description. If not, you may skip the remainder of the questions in this section.}

\dsanswer{No preprocessing was performed on the \textit{sequence} data released in this dataset.
}

\dsquestionex{Was the “raw” data saved in addition to the preprocessed/cleaned/labeled data (e.g., to support unanticipated future uses)?}{If so, please provide a link or other access point to the “raw” data.}

\dsanswer{n/a
}

\dsquestionex{Is the software used to preprocess/clean/label the instances available?}{If so, please provide a link or other access point.}

\dsanswer{n/a
}

\dsquestion{Any other comments?}

\dsanswer{None.
}

\bigskip
\dssectionheader{Uses}

\dsquestionex{Has the dataset been used for any tasks already?}{If so, please provide a description.}

\dsanswer{Yes: this dataset was released to accompany the 2022 Multi-Agent Behavior (MABe) Challenge, posted \href{https://www.aicrowd.com/challenges/multi-agent-behavior-challenge-2022}{here}. This competition was aimed at generating learned representations of animals' actions using unsupervised or self-supervised techniques.
}

\dsquestionex{Is there a repository that links to any or all papers or systems that use the dataset?}{If so, please provide a link or other access point.}

\dsanswer{Papers that use or cite this dataset may be submitted by their authors for display on the MABe22 website at \href{https://sites.google.com/view/computational-behavior/our-datasets/mabe2022-dataset}{https://sites.google.com/view/computational-behavior/our-datasets/mabe2022-dataset}
}

\dsquestion{What (other) tasks could the dataset be used for?}

\dsanswer{While this dataset was designed for development of methods for representation learning, the annotations can also be used for supervised learning tasks.
}

\dsquestionex{Is there anything about the composition of the dataset or the way it was collected and preprocessed/cleaned/labeled that might impact future uses?}{For example, is there anything that a future user might need to know to avoid uses that could result in unfair treatment of individuals or groups (e.g., stereotyping, quality of service issues) or other undesirable harms (e.g., financial harms, legal risks) If so, please provide a description. Is there anything a future user could do to mitigate these undesirable harms?}

\dsanswer{Occasional errors and identity swaps during pose estimation may impact future use of the dataset for some purposes.
}

\dsquestionex{Are there tasks for which the dataset should not be used?}{If so, please provide a description.}

\dsanswer{None.
}

\dsquestion{Any other comments?}

\dsanswer{None.
}

\bigskip
\dssectionheader{Distribution}

\dsquestionex{Will the dataset be distributed to third parties outside of the entity (e.g., company, institution, organization) on behalf of which the dataset was created?}{If so, please provide a description.}

\dsanswer{Yes - the full dataset will be made publicly available for download by all interested parties by July 1st, 2023.
}

\dsquestionex{How will the dataset will be distributed (e.g., tarball on website, API, GitHub)}{Does the dataset have a digital object identifier (DOI)?}

\dsanswer{The dataset is available on the Caltech public data repository at \href{https://data.caltech.edu/records/rdsa8-rde65}{https://data.caltech.edu/records/rdsa8-rde65}.
A previous version, containing only the trajectory data, is available at
\href{https://data.caltech.edu/records/20186}{https://data.caltech.edu/records/20186}. The data will be retained indefinitely and available for download by all third parties. The data.caltech.edu posting has accompanying DOI \href{https://doi.org/10.22002/rdsa8-rde65}{https://doi.org/10.22002/rdsa8-rde65}.

The dataset as used for the MABe Challenge (lacking hidden task labels) is available for download on the AIcrowd page, located at (\href{https://www.aicrowd.com/challenges/multi-agent-behavior-challenge-2022/problems/mabe-2022-mouse-triplets-video-data}{https://www.aicrowd.com/challenges/multi-agent-behavior-challenge-2022/problems/mabe-2022-mouse-triplets-video-data}). 

}

\dsquestion{When will the dataset be distributed?}

\dsanswer{Yes - the full dataset will be made publicly available for download by all interested parties by July 1st, 2023.
}

\dsquestionex{Will the dataset be distributed under a copyright or other intellectual property (IP) license, and/or under applicable terms of use (ToU)?}{If so, please describe this license and/or ToU, and provide a link or other access point to, or otherwise reproduce, any relevant licensing terms or ToU, as well as any fees associated with these restrictions.}

\dsanswer{The MABe22 dataset is distributed under the CreativeCommons Attribution-NonCommercial-ShareAlike license (CC-BY-NC-SA). The terms of this license may be found at \href{https://creativecommons.org/licenses/by-nc-sa/2.0/legalcode}{https://creativecommons.org/licenses/by-nc-sa/2.0/legalcode}.
}

\dsquestionex{Have any third parties imposed IP-based or other restrictions on the data associated with the instances?}{If so, please describe these restrictions, and provide a link or other access point to, or otherwise reproduce, any relevant licensing terms, as well as any fees associated with these restrictions.}

\dsanswer{There are no third party restrictions on the data.
}

\dsquestionex{Do any export controls or other regulatory restrictions apply to the dataset or to individual instances?}{If so, please describe these restrictions, and provide a link or other access point to, or otherwise reproduce, any supporting documentation.}

\dsanswer{No export controls or regulatory restrictions apply.
}

\dsquestion{Any other comments?}

\dsanswer{None.
}

\bigskip
\dssectionheader{Maintenance}

\dsquestion{Who will be supporting/hosting/maintaining the dataset?}

\dsanswer{The dataset is hosted on the Caltech Research Data Repository at \href{https://data.caltech.edu/}{data.caltech.edu}. Dataset hosting is maintained by the library of the California Institute of Technology. Long-term support for users of the dataset is provided by Jennifer J. Sun and by the laboratory of Ann Kennedy.
}

\dsquestion{How can the owner/curator/manager of the dataset be contacted (e.g., email address)?}

\dsanswer{The managers of the dataset (JJS and AK) can be contacted at \href{mailto:mabe.workshop@gmail.com}{mabe.workshop@gmail.com}, or AK can be contacted at \href{mailto:ann.kennedy@northwestern.edu}{ann.kennedy@northwestern.edu} and JJS can be contacted at \href{mailto:jjsun@caltech.edu}{jjsun@caltech.edu}.
}

\dsquestionex{Is there an erratum?}{If so, please provide a link or other access point.}

\dsanswer{No.
}

\dsquestionex{Will the dataset be updated (e.g., to correct labeling errors, add new instances, delete instances)?}{If so, please describe how often, by whom, and how updates will be communicated to users (e.g., mailing list, GitHub)?}

\dsanswer{Users of the dataset have the option to subscribe to a mailing list to receive updates regarding corrections or extensions of the MABe22 dataset. Mailing list sign-up can be found on the MABe22 webpage at \href{https://sites.google.com/view/computational-behavior/our-datasets/mabe2022-dataset}{https://sites.google.com/view/computational-behavior/our-datasets/mabe2022-dataset}.

Updates to correct errors in the dataset will be made promptly, and announced via update messages posted to the MABe22 website and data.caltech.edu page.

Updates that extend the scope of the dataset, such as additional hidden tasks, or improved pose estimation, will be released as new named instantiations on at most a yearly basis. Previous versions of the dataset will remain online, but obsolescence notes will be sent out to the MABe22 mailing list. In updates, dataset version will be indicated by the year in the dataset name (here 22). Dataset updates may accompany new instantiations of the MABe Challenge.
}

\dsquestionex{If the dataset relates to people, are there applicable limits on the retention of the data associated with the instances (e.g., were individuals in question told that their data would be retained for a fixed period of time and then deleted)?}{If so, please describe these limits and explain how they will be enforced.}

\dsanswer{N/a (no human data.)
}

\dsquestionex{Will older versions of the dataset continue to be supported/hosted/maintained?}{If so, please describe how. If not, please describe how its obsolescence will be communicated to users.}

\dsanswer{Yes, the dataset will be permanently available on the Caltech Research Data Repository (data.caltech.edu), which is managed by the Caltech Library.
}

\dsquestionex{If others want to extend/augment/build on/contribute to the dataset, is there a mechanism for them to do so?}{If so, please provide a description. Will these contributions be validated/verified? If so, please describe how. If not, why not? Is there a process for communicating/distributing these contributions to other users? If so, please provide a description.}

\dsanswer{Extensions to the dataset will take place through at-most-yearly updates. We welcome community contributions of behavioral data, novel tracking methods, and novel hidden tasks; these may be submitted by contacting the authors or emailing \href{mailto:mabe.workshop@gmail.com}{mabe.workshop@gmail.com}. All community contributions will be reviewed by the managers of the dataset for quality of tracking and annotation data. Community contributions will not be accepted without a data maintenance plan (similar to this document), to ensure support for future users of the dataset.
}

\dsquestion{Any other comments?}

\dsanswer{If you enjoyed this dataset and would like to contribute other multi-agent behavioral data for future versions of the dataset or MABe Challenge, contact us at \href{mailto:mabe.workshop@gmail.com}{mabe.workshop@gmail.com}!
}

\subsection{Fly Datasheet}


\dssectionheader{Motivation}

\dsquestionex{For what purpose was the dataset created?}{Was there a specific task in mind? Was there a specific gap that needed to be filled? Please provide a description.}

\dsanswer{
The prospect of discovering structure previously unknown to humans from large datasets has tremendous potential, particularly for science. However, progress has been inhibited by a lack of common datasets and quantitative evaluation criteria for assessing and comparing different algorithms. In the field of video-based behavior analysis, there has been a lot of progress in tools for tracking the pose of people and animals. To make use of these methods in biology, we now need computational methods to probe the temporal structure in these still large time-series datasets, and learn representations amenable to comparison and further study.

The MABe22 dataset is a new animal behavior dataset, intended to a) serve as a benchmark dataset for comparison of unsupervised or self-supervised behavior analysis tools, and establish community standards for evaluation of unsupervised techniques, b) highlight critical challenges in computational behavior analysis, particularly pertaining to unsupervised representation learning, and c) foster interaction between behavioral biologists and the greater machine learning community. This datasheet is specific to the Fly Group dataset, which consists of tracking data for a group of 8 to 11 fruit flies with 50 ``hidden labels" for evaluating the quality of the learned representation.

Also see MABe22 mouse triplet data sheet (Section~\ref{appendix:mouse_datasheet}) for more details.
}

\dsquestion{Who created this dataset (e.g., which team, research group) and on behalf of which entity (e.g., company, institution, organization)?}

\dsanswer{The MABe22 fly dataset was created as a collaborative effort between Kristin Branson, Alice Robie, and Catherine Schretter at HHMI Janelia Research Campus within the labs of Kristin Branson and Gerry Rubin. Fly lines were generated by Gerry Rubin with the help of the Janelia Fly Core, PTR, and Fly Light project teams. Fly crosses and offspring were set up and collected by Alice Robie and Catherine Schretter, the behavior rig was developed by Alice Robie and Kristin Branson, and video were recorded by Alice Robie and Catherine Schretter, with help from Janelia Shared Resources. Analysis was done by Kristin Branson, Alice Robie, and Catherine Schretter, with help from Adam Taylor. The dataset tasks were designed by Kristin Branson, Alice Robie, and Catherine Schretter. 
}

\dsquestionex{Who funded the creation of the dataset?}{If there is an associated grant, please provide the name of the grantor and the grant name and number.}

\dsanswer{Acquisition of behavioral data was funded by the Howard Hughes Medical Institute. 
}

\dsquestion{Any other comments?}

\dsanswer{None.}

\bigskip
\dssectionheader{Composition}

\dsquestionex{What do the instances that comprise the dataset represent (e.g., documents, photos, people, countries)?}{ Are there multiple types of instances (e.g., movies, users, and ratings; people and interactions between them; nodes and edges)? Please provide a description.}

\dsanswer{The core element of this dataset, called a \textit{sequence}, captures the tracked postures of $\approx 10$ flies over 30s (4,500 frames) on a 5-cm-diameter domed plate filmed from above at 150Hz. 

The core element of a \textit{sequence} is called a \textit{frame}; this refers to the posture of all animals on a particular frame of video, as binary categorization for each of the 50 tasks. 

Tasks were based on the genotype, rearing, mutation, and environmental conditions of the flies. Flies from the following genotypes were assayed: dTrpA1 x pBDPGAL4U (Control)~\cite{robie2017mapping}, dTrpA1 x R71G01 (R71G01)~\cite{robie2017mapping}, dTrpA1 x R65F12 (R65F12)~\cite{robie2017mapping}, 20xCsChrimson x SS36551 (aIPg)\cite{schretter2020cell}, NorpA,20xCsChrimson x NorpA;SS36564 (Blind aIPg), 20x CsChrimson x SS56987 (pC1d)\cite{schretter2020cell}, 20x CsChrimson x BPp65AD-x-BPZpGal4DBD (Control 2)\cite{schretter2020cell}, NorpA,20xCsChrimson x NorpA;BPp65AD-x-BPZpGal4DBD (Blind control). Neural populations in CsChrimson flies were activated by periods of red light illumination from an LED panel below the flies. Neural populations in dTrpA1 flies were activated by performing the experiments at the permissive temperature for TrpA. In addition, we manually annotated 6 social behaviors sparsely across the dataset. 
}

\dsquestion{How many instances are there in total (of each type, if appropriate)?}

\dsanswer{
Instances for each dataset are shown in Table ~\ref{tab:fly_nframes}, divided into user train, evaluator train, test 1, and test 2 setss. Number of instances is reported as \textit{frames}. As frames within a sequence are temporally contiguous and sampled at 150Hz, they are not statistically independent observations. 

\begin{table}[]
\centering
\begin{scriptsize}
\begin{tabular}{|*{9}{c|}}\hline
&\multicolumn{4}{|c|}{Category 1}&\multicolumn{4}{|c|}{Category 0}\\\hline
Task & User train & Eval train & Test 1 & Test 2 & User train & Eval train & Test 1 & Test 2 \\\hline
Female vs male & 13,808,901 & 7,696,105 & 5,155,335 & 6,452,311 & 4,470,088 & 1,744,888 & 1,338,165 & 1,562,188\\\hline
Control 1 & 1,863,000 & 729,000 & 364,491 & 405,000 & 11,657,257 & 5,642,878 & 4,173,622 & 5,085,126\\\hline
Control 1 sex separated & 405,000 & 364,491 & 405,000 & 364,500 & 13,115,257 & 6,007,387 & 4,133,113 & 5,125,626\\\hline
Control 2 & 726,798 & 516,548 & 287,012 & 283,151 & 12,793,459 & 5,855,330 & 4,251,101 & 5,206,975\\\hline
71G01 & 2,668,497 & 769,500 & 364,509 & 405,011 & 10,851,760 & 5,602,378 & 4,173,604 & 5,085,115\\\hline
Male R71G01 female control & 405,008 & 9 & 405,000 & 405,000 & 13,115,249 & 6,371,869 & 4,133,113 & 5,085,126\\\hline
R65F12 & 1,853,994 & 1,003,505 & 810,000 & 764,998 & 11,666,263 & 5,368,373 & 3,728,113 & 4,725,128\\\hline
R91B01 & 1,944,000 & 729,000 & 364,500 & 810,000 & 11,576,257 & 5,642,878 & 4,173,613 & 4,680,126\\\hline
Blind control & 1,011,001 & 543,761 & 236,961 & 468,234 & 12,509,256 & 5,828,117 & 4,301,152 & 5,021,892\\\hline
aIPg & 1,166,857 & 418,714 & 262,500 & 520,350 & 12,353,400 & 5,953,164 & 4,275,613 & 4,969,776\\\hline
pC1d & 520,770 & 516,990 & 518,450 & 518,890 & 12,999,487 & 5,854,888 & 4,019,663 & 4,971,236\\\hline
Blind aIPg & 955,332 & 780,360 & 519,690 & 544,992 & 12,564,925 & 5,591,518 & 4,018,423 & 4,945,134\\\hline
Blind control on vs off & 1,011,001 & 543,761 & 236,961 & 468,234 & 1,094,999 & 590,239 & 249,039 & 503,766\\\hline
Blind control strong vs off & 505,572 & 272,425 & 118,314 & 232,335 & 1,094,999 & 590,239 & 249,039 & 503,766\\\hline
Blind control weak vs off & 505,429 & 271,336 & 118,647 & 235,899 & 1,094,999 & 590,239 & 249,039 & 503,766\\\hline
Blind control strong vs weak & 505,572 & 272,425 & 118,314 & 232,335 & 505,429 & 271,336 & 118,647 & 235,899\\\hline
Blind control last vs first & 169,813 & 88,736 & 38,970 & 78,507 & 168,405 & 91,702 & 39,771 & 78,048\\\hline
Control 2 on vs off & 726,798 & 516,548 & 287,012 & 283,151 & 785,202 & 563,452 & 306,988 & 310,849\\\hline
Control 2 strong vs off & 361,015 & 258,143 & 143,836 & 141,922 & 785,202 & 563,452 & 306,988 & 310,849\\\hline
Control 2 weak vs off & 365,783 & 258,405 & 143,176 & 141,229 & 785,202 & 563,452 & 306,988 & 310,849\\\hline
Control 2 strong vs weak & 361,015 & 258,143 & 143,836 & 141,922 & 365,783 & 258,405 & 143,176 & 141,229\\\hline
Control 2 last vs first & 121,560 & 85,523 & 48,609 & 48,081 & 120,672 & 86,526 & 46,849 & 46,761\\\hline
Blind aIPg on vs off & 955,332 & 780,360 & 519,690 & 544,992 & 1,042,668 & 839,630 & 560,310 & 589,008\\\hline
Blind aIPg strong vs off & 477,531 & 389,800 & 260,930 & 271,073 & 1,042,668 & 839,630 & 560,310 & 589,008\\\hline
Blind aIPg weak vs off & 477,801 & 390,560 & 258,760 & 273,919 & 1,042,668 & 839,630 & 560,310 & 589,008\\\hline
Blind aIPg strong vs weak & 477,531 & 389,800 & 260,930 & 271,073 & 477,801 & 390,560 & 258,760 & 273,919\\\hline
Blind aIPg last vs first & 159,555 & 130,120 & 85,810 & 90,343 & 158,332 & 129,920 & 87,240 & 89,876\\\hline
aIPg on vs off & 1,166,857 & 418,714 & 262,500 & 520,350 & 1,276,633 & 512,784 & 277,500 & 559,650\\\hline
aIPg strong vs off & 598,592 & 210,374 & 131,340 & 259,890 & 1,276,633 & 512,784 & 277,500 & 559,650\\\hline
aIPg weak vs off & 568,265 & 208,340 & 131,160 & 260,460 & 1,276,633 & 512,784 & 277,500 & 559,650\\\hline
aIPg strong vs weak & 598,592 & 210,374 & 131,340 & 259,890 & 568,265 & 208,340 & 131,160 & 260,460\\\hline
aIPg last vs first & 199,900 & 77,067 & 44,210 & 86,860 & 198,901 & 76,662 & 44,120 & 86,150\\\hline
pC1 on vs off & 520,770 & 516,990 & 518,450 & 518,890 & 559,230 & 563,010 & 561,550 & 561,100\\\hline
pC1d strong vs off & 258,760 & 258,370 & 260,100 & 257,520 & 559,230 & 563,010 & 561,550 & 561,100\\\hline
pC1d weak vs off & 262,010 & 258,620 & 258,350 & 261,370 & 559,230 & 563,010 & 561,550 & 561,100\\\hline
pC1d strong vs weak & 258,760 & 258,370 & 260,100 & 257,520 & 262,010 & 258,620 & 258,350 & 261,370\\\hline
pC1d last vs first & 86,520 & 86,760 & 86,110 & 86,780 & 85,320 & 85,660 & 87,030 & 86,490\\\hline
Any courtship & 4,927,499 & 1,773,014 & 1,579,509 & 1,575,009 & 8,592,758 & 4,598,864 & 2,958,604 & 3,915,117\\\hline
Any control & 4,005,799 & 2,153,800 & 1,293,464 & 1,520,885 & 9,514,458 & 4,218,078 & 3,244,649 & 3,969,241\\\hline
Any blind & 1,966,333 & 1,324,121 & 756,651 & 1,013,226 & 11,553,924 & 5,047,757 & 3,781,462 & 4,476,900\\\hline
Any aIPg & 2,122,189 & 1,199,074 & 782,190 & 1,065,342 & 11,398,068 & 5,172,804 & 3,755,923 & 4,424,784\\\hline
Any aggression & 2,642,959 & 1,716,064 & 1,300,640 & 1,584,232 & 10,877,298 & 4,655,814 & 3,237,473 & 3,905,894\\\hline
Any R71G01 & 3,073,505 & 769,509 & 769,509 & 810,011 & 10,446,752 & 5,602,369 & 3,768,604 & 4,680,115\\\hline
Any sex-separated & 810,008 & 364,500 & 810,000 & 769,500 & 12,710,249 & 6,007,378 & 3,728,113 & 4,720,626\\\hline
Aggression manual annotation & 610 & 972 & 1,279 & 890 & 480 & 1,092 & 1,014 & 1,487\\\hline
Chase manual annotation & 1,496 & 15,351 & 5,611 & 20,810 & 2,218 & 51,938 & 31,232 & 34,382\\\hline
Courtship manual annotation & 591 & 743 & 273 & 108 & 3,388 & 2,979 & 2,465 & 1,728\\\hline
High fence manual annotation & 188 & 157 & 106 & 158 & 751 & 584 & 570 & 629\\\hline
Wing ext.~manual annotation & 0 & 1,594 & 1,524 & 3,130 & 0 & 13,396 & 11,728 & 15,104\\\hline
Wing flick manual annotation & 230 & 149 & 95 & 176 & 1,740 & 1,404 & 840 & 1,469\\\hline
\end{tabular}
\end{scriptsize}
\caption{Number of frames in each split set for each task and category.}
\label{tab:fly_nframes}
\end{table}

}

\begin{table}[]
\centering
\begin{scriptsize}
\begin{tabular}{|*{9}{c|}}\hline
&\multicolumn{4}{|c|}{Category 1}&\multicolumn{4}{|c|}{Category 0}\\\hline
Task & User train & Eval train & Test 1 & Test 2 & User train & Eval train & Test 1 & Test 2 \\\hline
Female vs male & 426 & 217 & 147 & 179 & 221 & 91 & 67 & 76\\\hline
Control 1 & 50 & 20 & 9 & 10 & 373 & 193 & 136 & 166\\\hline
Control 1 sex separated & 10 & 9 & 10 & 10 & 408 & 202 & 135 & 165\\\hline
Control 2 & 33 & 22 & 11 & 11 & 385 & 189 & 133 & 164\\\hline
71G01 & 66 & 20 & 11 & 11 & 359 & 193 & 135 & 166\\\hline
Male R71G01 female control & 11 & 1 & 10 & 10 & 407 & 211 & 134 & 166\\\hline
R65F12 & 45 & 25 & 20 & 18 & 376 & 187 & 126 & 158\\\hline
R91B01 & 49 & 19 & 10 & 20 & 374 & 192 & 134 & 156\\\hline
Blind control & 44 & 22 & 11 & 22 & 374 & 189 & 133 & 153\\\hline
aIPg & 54 & 21 & 11 & 22 & 364 & 190 & 133 & 153\\\hline
pC1d & 22 & 22 & 22 & 22 & 396 & 189 & 122 & 153\\\hline
Blind aIPg & 44 & 33 & 22 & 22 & 374 & 178 & 122 & 153\\\hline
Blind control on vs off & 44 & 22 & 11 & 22 & 52 & 26 & 13 & 26\\\hline
Blind control strong vs off & 24 & 12 & 6 & 12 & 52 & 26 & 13 & 26\\\hline
Blind control weak vs off & 20 & 10 & 5 & 10 & 52 & 26 & 13 & 26\\\hline
Blind control strong vs weak & 24 & 12 & 6 & 12 & 20 & 10 & 5 & 10\\\hline
Blind control last vs first & 8 & 4 & 2 & 4 & 8 & 4 & 2 & 4\\\hline
Control 2 on vs off & 33 & 22 & 11 & 11 & 38 & 26 & 13 & 13\\\hline
Control 2 strong vs off & 18 & 12 & 6 & 6 & 38 & 26 & 13 & 13\\\hline
Control 2 weak vs off & 15 & 10 & 5 & 5 & 38 & 26 & 13 & 13\\\hline
Control 2 strong vs weak & 18 & 12 & 6 & 6 & 15 & 10 & 5 & 5\\\hline
Control 2 last vs first & 6 & 4 & 2 & 2 & 6 & 4 & 2 & 2\\\hline
Blind aIPg on vs off & 44 & 33 & 22 & 22 & 52 & 38 & 26 & 26\\\hline
Blind aIPg strong vs off & 24 & 18 & 12 & 12 & 52 & 38 & 26 & 26\\\hline
Blind aIPg weak vs off & 20 & 15 & 10 & 10 & 52 & 38 & 26 & 26\\\hline
Blind aIPg strong vs weak & 24 & 18 & 12 & 12 & 20 & 15 & 10 & 10\\\hline
Blind aIPg last vs first & 8 & 6 & 4 & 4 & 8 & 6 & 4 & 4\\\hline
aIPg on vs off & 54 & 21 & 11 & 22 & 62 & 25 & 13 & 26\\\hline
aIPg strong vs off & 29 & 11 & 6 & 12 & 62 & 25 & 13 & 26\\\hline
aIPg weak vs off & 25 & 10 & 5 & 10 & 62 & 25 & 13 & 26\\\hline
aIPg strong vs weak & 29 & 11 & 6 & 12 & 25 & 10 & 5 & 10\\\hline
aIPg last vs first & 10 & 4 & 2 & 4 & 10 & 4 & 2 & 4\\\hline
pC1 on vs off & 22 & 22 & 22 & 22 & 26 & 26 & 26 & 24\\\hline
pC1d strong vs off & 12 & 12 & 12 & 12 & 26 & 26 & 26 & 24\\\hline
pC1d weak vs off & 10 & 10 & 10 & 10 & 26 & 26 & 26 & 24\\\hline
pC1d strong vs weak & 12 & 12 & 12 & 12 & 10 & 10 & 10 & 10\\\hline
pC1d last vs first & 4 & 4 & 4 & 4 & 4 & 4 & 4 & 4\\\hline
Any courtship & 120 & 45 & 40 & 37 & 304 & 168 & 106 & 138\\\hline
Any control & 137 & 73 & 41 & 53 & 286 & 140 & 105 & 123\\\hline
Any blind & 88 & 55 & 33 & 44 & 330 & 156 & 111 & 131\\\hline
Any aIPg & 98 & 54 & 33 & 44 & 320 & 157 & 111 & 131\\\hline
Any aggression & 120 & 76 & 55 & 66 & 298 & 135 & 89 & 109\\\hline
Any R71G01 & 77 & 20 & 21 & 20 & 348 & 192 & 125 & 156\\\hline
Any sex-separated & 21 & 10 & 20 & 20 & 397 & 202 & 125 & 156\\\hline
Aggression manual annotation & 11 & 16 & 15 & 17 & 10 & 20 & 17 & 30\\\hline
Chase manual annotation & 2 & 23 & 11 & 23 & 4 & 76 & 63 & 65\\\hline
Courtship manual annotation & 5 & 6 & 6 & 2 & 38 & 32 & 34 & 22\\\hline
High fence manual annotation & 12 & 17 & 13 & 17 & 23 & 27 & 16 & 18\\\hline
Wing ext.~manual annotation & 0 & 4 & 8 & 8 & 0 & 52 & 49 & 54\\\hline
Wing flick manual annotation & 28 & 19 & 16 & 28 & 52 & 40 & 24 & 50\\\hline
\end{tabular}
\end{scriptsize}
\caption{Number of sequences in each split set for each task and category.}
\label{tab:fly_nseqs}
\end{table}

\dsquestionex{Does the dataset contain all possible instances or is it a sample (not necessarily random) of instances from a larger set?}{ If the dataset is a sample, then what is the larger set? Is the sample representative of the larger set (e.g., geographic coverage)? If so, please describe how this representativeness was validated/verified. If it is not representative of the larger set, please describe why not (e.g., to cover a more diverse range of instances, because instances were withheld or unavailable).}

\dsanswer{We used all videos from chosen genotypes and conditions containing at least 9 flies. Frames for manual annotation of behavior were chosen using JAABA's interactive system~\cite{kabra2013jaaba} to help find instances of rare behaviors. When cutting a video into sequences, we chose segments to avoid obvious identity tracking errors (trajectory births or deaths). We left gaps of a randomly chosen length between .5 and 2s (75 and 300 frames) between sequences. 
}

\dsquestionex{What data does each instance consist of? “Raw” data (e.g., unprocessed text or images) or features?}{In either case, please provide a description.}

\dsanswer{Each sequence has three elements. 1) \textit{Tracking features} consist of, for each of the $\approx 10$ flies, the locations of 19 body parts (left wing tip, right wing tip, antennae midpoint, right eye, left eye, left front of thorax, right front of thorax, base of thorax, tip of abdomen, right middle femur base, right middle femur-tibia joint, left middle femur-base, left middle femur-tibia joint, right front leg tip, right middle leg tip, right rear leg tip, left front leg tip, left middle leg tip, left rear leg tip), information about an ellipse fit to the fly body (Fit ellipse center, orientation, major and minor axis length), and information about the segmented animal (body and foreground area, image contrast). Tracking features are estimated using the Animal Part Tracker (APT) and the FlyTracker. Videos have between 9 and 11 flies. All data are stored as matrices with space for 11 flies, with nan values if there are $< 11$ flies. 2) \textit{Task categories} are frame- and fly-wise binary categorizations for each of the 50 tasks we defined, and will have values 1, 0, or nan, with nan indicating no data (the task is irrelevant or ill-defined for this frame and fly, or this frame and fly was not manually annotated). For some tasks, all flies in the same frame will have the same value. For some tasks, all frames will have the same value for the entire sequence, or for long periods of contiguous time. 
}

\dsquestionex{Is there a label or target associated with each instance?}{If so, please provide a description.}

\dsanswer{The \textit{annotation} field for a given sequence consists of frame- and fly-wise categorizations for each of the 50 tasks. For fly-frames for which the task is irrelevant or ill-defined, or no manual annotation was made, this label will be missing (indicated by nan). In the MABe22 challenge, these task annotations were kept secret, and used for evaluation purposes, not for training. 
}

\dsquestionex{Is any information missing from individual instances?}{If so, please provide a description, explaining why this information is missing (e.g., because it was unavailable). This does not include intentionally removed information, but might include, e.g., redacted text.}

\dsanswer{As described above, all data are stored as matrices with space for 11 flies, with nan values if there are $< 11$ flies. \textit{Annotations} will be nan if the task is irrelevant or ill-defined for this frame and fly, or this frame and fly was not manually annotated.
}

\dsquestionex{Are relationships between individual instances made explicit (e.g., users’ movie ratings, social network links)?}{If so, please describe how these relationships are made explicit.}

\dsanswer{Each instance (\textit{sequence}) is to be treated as an independent observation. Some sequence come from the same groups of flies in the same video. Each sequence is at least 0.5s (75 frames) from another sequence. Frames within a sequence are temporally contiguous, and highly correlated. 
}

\dsquestionex{Are there recommended data splits (e.g., training, development/validation, testing)?}{If so, please provide a description of these splits, explaining the rationale behind them.}

\dsanswer{The dataset includes a recommended split into User train (for unsupervised representation learning), Evaluator train (for training evaluator classifier), Test 1 (for validating the classifier), and Test 2 (for final evaluation score) sets. Each set containing distinct videos and flies. The splits were designed to provide a roughly consistent, small amount of training data for each task. 
}

\dsquestionex{Are there any errors, sources of noise, or redundancies in the dataset?}{If so, please provide a description.}

\dsanswer{Tracking in this dataset are produced using automated tracking software (\href{https://github.com/kristinbranson/FlyTracker}{FlyTracker} and \href{https://github.com/kristinbranson/APT}{APT}). In addition, manual annotations of animal behavior are inherently subjective, and individual annotators show some variability in the precise frame-by-frame labeling of behavior sequences.
}

\dsquestionex{Is the dataset self-contained, or does it link to or otherwise rely on external resources (e.g., websites, tweets, other datasets)?}{If it links to or relies on external resources, a) are there guarantees that they will exist, and remain constant, over time; b) are there official archival versions of the complete dataset (i.e., including the external resources as they existed at the time the dataset was created); c) are there any restrictions (e.g., licenses, fees) associated with any of the external resources that might apply to a future user? Please provide descriptions of all external resources and any restrictions associated with them, as well as links or other access points, as appropriate.}

\dsanswer{The dataset is self-contained.
}

\dsquestionex{Does the dataset contain data that might be considered confidential (e.g., data that is protected by legal privilege or by doctor-patient confidentiality, data that includes the content of individuals non-public communications)?}{If so, please provide a description.}

\dsanswer{No.
}

\dsquestionex{Does the dataset contain data that, if viewed directly, might be offensive, insulting, threatening, or might otherwise cause anxiety?}{If so, please describe why.}

\dsanswer{No such material; dataset contains only trajectories (no video or images) and text labels pertaining to fly social behaviors.
}

\dsquestionex{Does the dataset relate to people?}{If not, you may skip the remaining questions in this section.}

\dsanswer{No.
}

\dsquestionex{Does the dataset identify any subpopulations (e.g., by age, gender)?}{If so, please describe how these subpopulations are identified and provide a description of their respective distributions within the dataset.}

\dsanswer{n/a
}

\dsquestionex{Is it possible to identify individuals (i.e., one or more natural persons), either directly or indirectly (i.e., in combination with other data) from the dataset?}{If so, please describe how.}

\dsanswer{n/a
}

\dsquestionex{Does the dataset contain data that might be considered sensitive in any way (e.g., data that reveals racial or ethnic origins, sexual orientations, religious beliefs, political opinions or union memberships, or locations; financial or health data; biometric or genetic data; forms of government identification, such as social security numbers; criminal history)?}{If so, please provide a description.}

\dsanswer{n/a
}

\dsquestion{Any other comments?}

\dsanswer{
None.
}

\bigskip
\dssectionheader{Collection Process}

\dsquestionex{How was the data associated with each instance acquired?}{Was the data directly observable (e.g., raw text, movie ratings), reported by subjects (e.g., survey responses), or indirectly inferred/derived from other data (e.g., part-of-speech tags, model-based guesses for age or language)? If data was reported by subjects or indirectly inferred/derived from other data, was the data validated/verified? If so, please describe how.}

\dsanswer{See above for details on collection process. All data pertains to groups of interacting flies in carefully controlled environments.
}

\dsquestionex{What mechanisms or procedures were used to collect the data (e.g., hardware apparatus or sensor, manual human curation, software program, software API)?}{How were these mechanisms or procedures validated?}

\dsanswer{Details of fly genotypes and rearing are above. Flies were recorded in our custom developed behavior rig, which consists of a custom LED panel for back-illumination for recording in NIR and timed optogenetic activation in red, a custom 5-cm-diameter domed circular dish designed to reduce interactions with the arena wall and ceiling, a visual surround to isolate the flies, and a camera with an NIR-pass filter (FLIR Flea3) recording at 1024x1024 at 150Hz. We used data capture software based on the FlyBowlDataCapture system~\cite{robie2017mapping} and the Basic Image Acquisition System (BIAS, IORodeo). As described above, manual annotations were made using JAABA.
}

\dsquestion{If the dataset is a sample from a larger set, what was the sampling strategy (e.g., deterministic, probabilistic with specific sampling probabilities)?}

\dsanswer{As described above, we included videos with at least 9 flies in them. When cutting a video into sequences, we chose segments to avoid obvious identity tracking errors (trajectory births or deaths). We left gaps of a randomly chosen length between .5 and 2s (75 and 300 frames) between sequences. 
}

\dsquestion{Who was involved in the data collection process (e.g., students, crowdworkers, contractors) and how were they compensated (e.g., how much were crowdworkers paid)?}

\dsanswer{Full-time employees of Janelia's Shared Resources teams (Fly Core, Fly Light, Media, and Project Technical Resources) were involved in producing and maintaining flies.
}

\dsquestionex{Over what timeframe was the data collected? Does this timeframe match the creation timeframe of the data associated with the instances (e.g., recent crawl of old news articles)?}{If not, please describe the timeframe in which the data associated with the instances was created.}

\dsanswer{Videos associated with this dataset were collected between December 2020 and September 2021. Tracking and annotation was performed in October 2021 - February 2022. 
}

\dsquestionex{Were any ethical review processes conducted (e.g., by an institutional review board)?}{If so, please provide a description of these review processes, including the outcomes, as well as a link or other access point to any supporting documentation.}

\dsanswer{No.
}

\dsquestionex{Does the dataset relate to people?}{If not, you may skip the remaining questions in this section.}

\dsanswer{No.
}

\dsquestion{Did you collect the data from the individuals in question directly, or obtain it via third parties or other sources (e.g., websites)?}

\dsanswer{n/a
}

\dsquestionex{Were the individuals in question notified about the data collection?}{If so, please describe (or show with screenshots or other information) how notice was provided, and provide a link or other access point to, or otherwise reproduce, the exact language of the notification itself.}

\dsanswer{n/a
}

\dsquestionex{Did the individuals in question consent to the collection and use of their data?}{If so, please describe (or show with screenshots or other information) how consent was requested and provided, and provide a link or other access point to, or otherwise reproduce, the exact language to which the individuals consented.}

\dsanswer{n/a
}

\dsquestionex{If consent was obtained, were the consenting individuals provided with a mechanism to revoke their consent in the future or for certain uses?}{If so, please provide a description, as well as a link or other access point to the mechanism (if appropriate).}

\dsanswer{n/a
}

\dsquestionex{Has an analysis of the potential impact of the dataset and its use on data subjects (e.g., a data protection impact analysis) been conducted?}{If so, please provide a description of this analysis, including the outcomes, as well as a link or other access point to any supporting documentation.}

\dsanswer{n/a
}

\dsquestion{Any other comments?}

\dsanswer{None.
}

\bigskip
\dssectionheader{Preprocessing/cleaning/labeling}

\dsquestionex{Was any preprocessing/cleaning/labeling of the data done (e.g., discretization or bucketing, tokenization, part-of-speech tagging, SIFT feature extraction, removal of instances, processing of missing values)?}{If so, please provide a description. If not, you may skip the remainder of the questions in this section.}

\dsanswer{No preprocessing was performed on the \textit{sequence} data released in this dataset.
}

\dsquestionex{Was the “raw” data saved in addition to the preprocessed/cleaned/labeled data (e.g., to support unanticipated future uses)?}{If so, please provide a link or other access point to the “raw” data.}

\dsanswer{n/a
}

\dsquestionex{Is the software used to preprocess/clean/label the instances available?}{If so, please provide a link or other access point.}

\dsanswer{n/a
}

\dsquestion{Any other comments?}

\dsanswer{None.
}

\bigskip
\dssectionheader{Uses}

\dsquestionex{Has the dataset been used for any tasks already?}{If so, please provide a description.}

\dsanswer{Yes: this dataset was released to accompany the three tasks of the 2022 Multi-Agent Behavior (MABe) Challenge, posted \href{https://www.aicrowd.com/challenges/multi-agent-behavior-challenge-2022}{here}.
 In this challenge, competitors are provided video of multiple interacting animals and tasked with learning a general-purpose, low-dimensional representation of the video. They upload their learned representations to the evaluation site, which then trained simple linear classifiers on the set of secret tasks described above, and returns accuracy measures. 
 }

\dsquestionex{Is there a repository that links to any or all papers or systems that use the dataset?}{If so, please provide a link or other access point.}

\dsanswer{Papers that use or cite this dataset may be submitted by their authors for display on the MABe22 website at \href{https://sites.google.com/view/computational-behavior/our-datasets/mabe2022-dataset}{https://sites.google.com/view/computational-behavior/our-datasets/mabe2022-dataset}
}

\dsquestion{What (other) tasks could the dataset be used for?}

\dsanswer{Besides unsupervised representation learning, this dataset could also be used for supervised representation learning, using the hidden labels as supervision. 
}

\dsquestionex{Is there anything about the composition of the dataset or the way it was collected and preprocessed/cleaned/labeled that might impact future uses?}{For example, is there anything that a future user might need to know to avoid uses that could result in unfair treatment of individuals or groups (e.g., stereotyping, quality of service issues) or other undesirable harms (e.g., financial harms, legal risks) If so, please provide a description. Is there anything a future user could do to mitigate these undesirable harms?}

\dsanswer{No.
}

\dsquestionex{Are there tasks for which the dataset should not be used?}{If so, please provide a description.}

\dsanswer{None.
}

\dsquestion{Any other comments?}

\dsanswer{None.
}

\bigskip
\dssectionheader{Distribution}

\dsquestionex{Will the dataset be distributed to third parties outside of the entity (e.g., company, institution, organization) on behalf of which the dataset was created?}{If so, please provide a description.}

\dsanswer{Yes - the full dataset will be made publicly available for download by all interested parties by July 1st, 2023.
}

\dsquestionex{How will the dataset will be distributed (e.g., tarball on website, API, GitHub)}{Does the dataset have a digital object identifier (DOI)?}

\dsanswer{The dataset is available on the Caltech public data repository at \href{https://data.caltech.edu/records/rdsa8-rde65}{https://data.caltech.edu/records/rdsa8-rde65}.
A previous version, containing only the trajectory data, is available at
\href{https://data.caltech.edu/records/20186}{https://data.caltech.edu/records/20186}. The data will be retained indefinitely and available for download by all third parties. The data.caltech.edu posting has accompanying DOI \href{https://doi.org/10.22002/rdsa8-rde65}{https://doi.org/10.22002/rdsa8-rde65}.

The dataset as used for the MABe Challenge (lacking hidden task labels) is available for download on the AIcrowd page, located at (\href{https://www.aicrowd.com/challenges/multi-agent-behavior-challenge-2022/problems/mabe-2022-fruit-fly-groups}{https://www.aicrowd.com/challenges/multi-agent-behavior-challenge-2022/problems/mabe-2022-fruit-fly-groups}).  

}

\dsquestion{When will the dataset be distributed?}

\dsanswer{Yes - the full dataset will be made publicly available for download by all interested parties by July 1st, 2023.
}

\dsquestionex{Will the dataset be distributed under a copyright or other intellectual property (IP) license, and/or under applicable terms of use (ToU)?}{If so, please describe this license and/or ToU, and provide a link or other access point to, or otherwise reproduce, any relevant licensing terms or ToU, as well as any fees associated with these restrictions.}

\dsanswer{The MABe22 dataset is distributed under the CreativeCommons Attribution-NonCommercial-ShareAlike license (CC-BY-NC-SA). The terms of this license may be found at \href{https://creativecommons.org/licenses/by-nc-sa/2.0/legalcode}{https://creativecommons.org/licenses/by-nc-sa/2.0/legalcode}.
}

\dsquestionex{Have any third parties imposed IP-based or other restrictions on the data associated with the instances?}{If so, please describe these restrictions, and provide a link or other access point to, or otherwise reproduce, any relevant licensing terms, as well as any fees associated with these restrictions.}

\dsanswer{There are no third party restrictions on the data.
}

\dsquestionex{Do any export controls or other regulatory restrictions apply to the dataset or to individual instances?}{If so, please describe these restrictions, and provide a link or other access point to, or otherwise reproduce, any supporting documentation.}

\dsanswer{No export controls or regulatory restrictions apply.
}

\dsquestion{Any other comments?}

\dsanswer{None.
}

\bigskip
\dssectionheader{Maintenance}

\dsquestion{Who will be supporting/hosting/maintaining the dataset?}

\dsanswer{The dataset is hosted on the Caltech Research Data Repository at \href{https://data.caltech.edu/}{data.caltech.edu}. Dataset hosting is maintained by the library of the California Institute of Technology. Long-term support for users of the dataset is provided by Jennifer J. Sun and by the laboratory of Ann Kennedy.
}

\dsquestion{How can the owner/curator/manager of the dataset be contacted (e.g., email address)?}

\dsanswer{The managers of the dataset (JJS and AK) can be contacted at \href{mailto:mabe.workshop@gmail.com}{mabe.workshop@gmail.com}, or AK can be contacted at \href{mailto:ann.kennedy@northwestern.edu}{ann.kennedy@northwestern.edu} and JJS can be contacted at \href{mailto:jjsun@caltech.edu}{jjsun@caltech.edu}.
}

\dsquestionex{Is there an erratum?}{If so, please provide a link or other access point.}

\dsanswer{No.
}

\dsquestionex{Will the dataset be updated (e.g., to correct labeling errors, add new instances, delete instances)?}{If so, please describe how often, by whom, and how updates will be communicated to users (e.g., mailing list, GitHub)?}

\dsanswer{Users of the dataset have the option to subscribe to a mailing list to receive updates regarding corrections or extensions of the MABe22 dataset. Mailing list sign-up can be found on the MABe22 webpage at \href{https://sites.google.com/view/computational-behavior/our-datasets/mabe2022-dataset}{https://sites.google.com/view/computational-behavior/our-datasets/mabe2022-dataset}.

Updates to correct errors in the dataset will be made promptly, and announced via update messages posted to the MABe22 website and data.caltech.edu page.

Updates that extend the scope of the dataset, such as additional hidden tasks, or improved pose estimation, will be released as new named instantiations on at most a yearly basis. Previous versions of the dataset will remain online, but obsolescence notes will be sent out to the MABe22 mailing list. In updates, dataset version will be indicated by the year in the dataset name (here 22). Dataset updates may accompany new instantiations of the MABe Challenge.
}

\dsquestionex{If the dataset relates to people, are there applicable limits on the retention of the data associated with the instances (e.g., were individuals in question told that their data would be retained for a fixed period of time and then deleted)?}{If so, please describe these limits and explain how they will be enforced.}

\dsanswer{N/a (no human data.)
}

\dsquestionex{Will older versions of the dataset continue to be supported/hosted/maintained?}{If so, please describe how. If not, please describe how its obsolescence will be communicated to users.}

\dsanswer{Yes, the dataset will be permanently available on the Caltech Research Data Repository (data.caltech.edu), which is managed by the Caltech Library.
}

\dsquestionex{If others want to extend/augment/build on/contribute to the dataset, is there a mechanism for them to do so?}{If so, please provide a description. Will these contributions be validated/verified? If so, please describe how. If not, why not? Is there a process for communicating/distributing these contributions to other users? If so, please provide a description.}

\dsanswer{Extensions to the dataset will take place through at-most-yearly updates. We welcome community contributions of behavioral data, novel tracking methods, and novel hidden tasks; these may be submitted by contacting the authors or emailing \href{mailto:mabe.workshop@gmail.com}{mabe.workshop@gmail.com}. All community contributions will be reviewed by the managers of the dataset for quality of tracking and annotation data. Community contributions will not be accepted without a data maintenance plan (similar to this document), to ensure support for future users of the dataset.
}

\dsquestion{Any other comments?}

\dsanswer{If you enjoyed this dataset and would like to contribute other multi-agent behavioral data for future versions of the dataset or MABe Challenge, contact us at \href{mailto:mabe.workshop@gmail.com}{mabe.workshop@gmail.com}!
}


\subsection{Beetle Datasheet}


\dssectionheader{Motivation}

\dsquestionex{For what purpose was the dataset created?}{Was there a specific task in mind? Was there a specific gap that needed to be filled? Please provide a description.}

\dsanswer{Interactions between different animal species constitute a core component of how ecological communities function. How these interactions work mechanistically promises to provide rich insight for the neuroscience community, as well as critical information on how networks of organisms operate in nature. Studying these interactions consist of understanding how sensory systems control response to the many different species an animal will encounter, what simple modules string together to build complex behaviors, how stereotyped are the behavioral outputs in response to particular stimuli, etc. Most quantitative behavioral data to this point is composed of either solo organisms, or members of the same species interacting. Our dataset provides behavioral video data of pairs of different species interacting. 

The Multi-Agent Behavior 2022 (MABe22) dataset is a new set of animal tracking, pose, video, and behavior datasets, intended to serve as a benchmark dataset for evaluation of unsupervised/self-supervised behavior representation learning and discovery methods. This datasheet is specific to the Ant-Beetle Interaction dataset, which consists of video recordings of rove beetles (\textit{Sceptobius lativentris}) interacting with velvety tree ants (\textit{Liometopum occidentale}, a species that rove beetles interact with symbiotically) and with other beetle species. This data offers a test case for algorithmic approaches to identify and assess the behavior space that these interaction partners traverse.
}

\dsquestion{Who created this dataset (e.g., which team, research group) and on behalf of which entity (e.g., company, institution, organization)?}

\dsanswer{The behavioral video data was collected and annotated by Julian Wagner in the lab of Joseph Parker at Caltech. Julian Wagner collected insects in the Los Angeles National Forest, filmed their interactions, and annotated their behavior in Behavioral Observation Research Interactive Software (BORIS) (\href{https://boris.readthedocs.io/en/latest/}{documentation link}) by Julian Wagner. Data was parsed into 30 second sections, downscaled, and pre-processed by Jennifer Sun.
}

\dsquestionex{Who funded the creation of the dataset?}{If there is an associated grant, please provide the name of the grantor and the grant name and number.}

\dsanswer{Acquisition of behavioral data was supported by Army Research Office MURI award W911NF1910269 (JP) and a US National Science Foundation CAREER award (2047472) (JP).
}

\dsquestion{Any other comments?}

\dsanswer{None.}

\bigskip
\dssectionheader{Composition}

\dsquestionex{What do the instances that comprise the dataset represent (e.g., documents, photos, people, countries)?}{ Are there multiple types of instances (e.g., movies, users, and ratings; people and interactions between them; nodes and edges)? Please provide a description.}

\dsanswer{The core element of this dataset, called a \textit{sequence}, is one 30-second video of a rove beetle (textit{Sceptobius lativentris}) interacting with another insect or object. Each video is accompanied by 14 frame- or sequence-level labels describing the species/type of interactor, the time elapsed since the start of the interaction session, as well as frame-wise manual annotations for six behaviors of interest. Video and annotations were originally acquired at 60 Hz, and are downsampled to 30 Hz in the released dataset.
}

\dsquestion{How many instances are there in total (of each type, if appropriate)?}

\dsanswer{
The dataset is composed of 11,536 30 second sequences.
}

\dsquestionex{Does the dataset contain all possible instances or is it a sample (not necessarily random) of instances from a larger set?}{ If the dataset is a sample, then what is the larger set? Is the sample representative of the larger set (e.g., geographic coverage)? If so, please describe how this representativeness was validated/verified. If it is not representative of the larger set, please describe why not (e.g., to cover a more diverse range of instances, because instances were withheld or unavailable).}

\dsanswer{The source dataset consists of 2-hour-long videos of rove beetle-interactor pairings, with each video capturing eight such pairings simultaneously (housed within the wells of an eight-well plate.) The raw video recordings were screened to identify wells that appeared to have occurrences of multiple types of behavior of interest; manual annotation of animal behavior was performed on this subset of wells. It is therefore possible that this dataset is biased for videos with higher rates of animal movement than in the full raw video dataset; this was done to provide a larger number of representative examples of animal behavior.

The 30 second clips comprising each instance in this dataset are extracted from the subset of wells for which annotation was performed. The extracted sequences included in this dataset are uniformly sampled from the source dataset as follows: first, the video is cropped to contain only the subject well for which behavior was annotated. Next, starting at the beginning of each video, we discard a randomly chosen segment of between 0.5 and 2 seconds (75 and 300 frames), then save the following 30-second clip as one dataset instance; this process is then repeated from the point where the preceding 30-second clip ended, onward through the end of the video. The clips therefore comprise a representative sample of the annotated ant-beetle interaction experiment.
}

\dsquestionex{What data does each instance consist of? “Raw” data (e.g., unprocessed text or images) or features?}{In either case, please provide a description.}

\dsanswer{Each instance consists of 30 seconds of raw video data (800x800 resolution and sampled at 30 Hz, i.e. 900 frames of images), accompanied by eight "sequence-level" labels of the interactor type and time since start of the interaction, and six "frame-level" labels which are manual annotations for the occurrence of various behaviors of interest (i.e. six binary vectors of length 900 indicating the presence or absence of each behavior on each frame of the video.)}

\dsquestionex{Is there a label or target associated with each instance?}{If so, please provide a description.}

\dsanswer{Yes, each instance is associated with a sequence-level label describing the species/object that the beetle is interacting with and a sequence-level label indicating the time elapsed since the start of the interaction session (between 0 and 4 hours). Each instance is also associated with six binary frame-wise labels indicating the presence or absence of a set of behaviors of interest.

}

\dsquestionex{Is any information missing from individual instances?}{If so, please provide a description, explaining why this information is missing (e.g., because it was unavailable). This does not include intentionally removed information, but might include, e.g., redacted text.}

\dsanswer{There is no missing data.
}

\dsquestionex{Are relationships between individual instances made explicit (e.g., users’ movie ratings, social network links)?}{If so, please describe how these relationships are made explicit.}

\dsanswer{Each instance (\textit{sequence}) is to be treated as an independent observation with no relationship to other instances in the dataset. Although the identities of the interacting animals are the same in some sequences, this information is not tracked in the dataset.
}

\dsquestionex{Are there recommended data splits (e.g., training, development/validation, testing)?}{If so, please provide a description of these splits, explaining the rationale behind them.}

\dsanswer{The dataset includes a recommended train/test split which was used for the Multi-Agent Behavior Challenge. Data was randomly split into training, test, and private-test sets (where the private test set was withheld from challenge evaluation until the end of the competition period, to avoid overfitting.)
}

\dsquestionex{Are there any errors, sources of noise, or redundancies in the dataset?}{If so, please provide a description.}

\dsanswer{The frame-wise annotations of behavior are manually generated by a trained human expert based on visual inspection the behavioral video, and are done by only one annotator. The initiation point of a particular behavior can be difficult to assess accurately and will be biased by the style of a given annotator. This makes the start and stop point of some behavioral categories (e.g. where a long grooming bout begins) more likely to be noisy and subjective to call than, say, the behavioral category in the middle of a protracted bout of a given behavior.

}

\dsquestionex{Is the dataset self-contained, or does it link to or otherwise rely on external resources (e.g., websites, tweets, other datasets)?}{If it links to or relies on external resources, a) are there guarantees that they will exist, and remain constant, over time; b) are there official archival versions of the complete dataset (i.e., including the external resources as they existed at the time the dataset was created); c) are there any restrictions (e.g., licenses, fees) associated with any of the external resources that might apply to a future user? Please provide descriptions of all external resources and any restrictions associated with them, as well as links or other access points, as appropriate.}

\dsanswer{The dataset is self-contained.
}

\dsquestionex{Does the dataset contain data that might be considered confidential (e.g., data that is protected by legal privilege or by doctor-patient confidentiality, data that includes the content of individuals non-public communications)?}{If so, please provide a description.}

\dsanswer{No.
}

\dsquestionex{Does the dataset contain data that, if viewed directly, might be offensive, insulting, threatening, or might otherwise cause anxiety?}{If so, please describe why.}

\dsanswer{No.
}

\dsquestionex{Does the dataset relate to people?}{If not, you may skip the remaining questions in this section.}

\dsanswer{No.
}

\dsquestionex{Does the dataset identify any subpopulations (e.g., by age, gender)?}{If so, please describe how these subpopulations are identified and provide a description of their respective distributions within the dataset.}

\dsanswer{n/a
}

\dsquestionex{Is it possible to identify individuals (i.e., one or more natural persons), either directly or indirectly (i.e., in combination with other data) from the dataset?}{If so, please describe how.}

\dsanswer{n/a
}

\dsquestionex{Does the dataset contain data that might be considered sensitive in any way (e.g., data that reveals racial or ethnic origins, sexual orientations, religious beliefs, political opinions or union memberships, or locations; financial or health data; biometric or genetic data; forms of government identification, such as social security numbers; criminal history)?}{If so, please provide a description.}

\dsanswer{n/a
}

\dsquestion{Any other comments?}

\dsanswer{
None.}

\bigskip
\dssectionheader{Collection Process}

\dsquestionex{How was the data associated with each instance acquired?}{Was the data directly observable (e.g., raw text, movie ratings), reported by subjects (e.g., survey responses), or indirectly inferred/derived from other data (e.g., part-of-speech tags, model-based guesses for age or language)? If data was reported by subjects or indirectly inferred/derived from other data, was the data validated/verified? If so, please describe how.}

\dsanswer{The raw behavioral videos were collected in a custom recording setup described in the following section. Videos were cropped and matted to isolate individual interaction wells, annotated by hand for behaviors and then split into the sequences. Sequence-level labels of interactor type and time since experiment start are ground-truth information known to the experimenter. Frame-wise labels of subject behavior are manually scored by a trained human expert; no secondary validation of these annotations was performed.
}

\dsquestionex{What mechanisms or procedures were used to collect the data (e.g., hardware apparatus or sensor, manual human curation, software program, software API)?}{How were these mechanisms or procedures validated?}

\dsanswer{Behavioral trials were performed in custom arenas made from 1/8th inch infrared transmitting acrylic (Plexiglass IR acrylic 3143, https://www.eplastics.com/plexiglass/acrylic-sheets/ir-transmitting) which transmits far red and infrared while blocking visible light. Arenas consist of a base layer of finely wet-sanded acrylic (to provide texture for beetles to walk on) a layer with eight two-centimeter round wells, a roof of anti-static acrylic (https://www.mcmaster.com/8774K17/) and a final top of inferred transmitting acrylic. Behavioral interactions were run at 15 C in a dark incubator with door closed. Arenas were top lit with IR850nm led flood lights. Recordings of interactions were made using a Flir machine vision camera (BFS-U3-51S5M-C: 5.0 MP) at 60 frames per second with a Pentax 12mm 1:1.2 TV lens (by Ricoh, FL-HC1212B-VG), for 2 hours.

To split multiplexed arena videos into individual wells, we manually set crop parameters for each well in each video, and cropped and matted the edges using openCV. We annotated videos of individual interaction wells with BORIS (Behavioral Observation Research Interactive Software (BORIS) user guide — BORIS latest documentation).
}

\dsquestion{If the dataset is a sample from a larger set, what was the sampling strategy (e.g., deterministic, probabilistic with specific sampling probabilities)?}

\dsanswer{
This answer is repeated from an earlier section: the source dataset consists of 2-hour-long videos of rove beetle-interactor pairings, with each video capturing eight such pairings simultaneously (housed within the wells of an eight-well plate.) The raw video recordings were screened to identify wells that appeared to have occurrences of multiple types of behavior of interest; manual annotation of animal behavior was performed on this subset of wells. It is therefore possible that this dataset is biased for videos with higher rates of animal movement than in the full raw video dataset; this was done to provide a larger number of representative examples of animal behavior.

The 30 second clips comprising each instance in this dataset are extracted from the subset of wells for which annotation was performed. The extracted sequences included in this dataset are uniformly sampled from the source dataset as follows: first, the video is cropped to contain only the subject well for which behavior was annotated. Next, starting at the beginning of each video, we discard a randomly chosen segment of between 0.5 and 2 seconds (75 and 300 frames), then save the following 30-second clip as one dataset instance; this process is then repeated from the point where the preceding 30-second clip ended, onward through the end of the video. The clips therefore comprise a representative sample of the annotated ant-beetle interaction experiment.
}

\dsquestion{Who was involved in the data collection process (e.g., students, crowdworkers, contractors) and how were they compensated (e.g., how much were crowdworkers paid)?}

\dsanswer{All data was collected and annotated by Julian Wagner, a graduate student in the lab of Joseph Parker, as part of their thesis work studying social symbiotic beetles. As a full-time employee of the Parker lab, Wagner's compensation was not dependent on participation in this project.
}

\dsquestionex{Over what timeframe was the data collected? Does this timeframe match the creation timeframe of the data associated with the instances (e.g., recent crawl of old news articles)?}{If not, please describe the timeframe in which the data associated with the instances was created.}

\dsanswer{Video data was collected and annotated over the course of several months in 2021.
}

\dsquestionex{Were any ethical review processes conducted (e.g., by an institutional review board)?}{If so, please provide a description of these review processes, including the outcomes, as well as a link or other access point to any supporting documentation.}

\dsanswer{No; because all species studied are invertebrates, these experiments are not subject to monitoring by an institutional review board.
}

\dsquestionex{Does the dataset relate to people?}{If not, you may skip the remaining questions in this section.}

\dsanswer{No.
}

\dsquestion{Did you collect the data from the individuals in question directly, or obtain it via third parties or other sources (e.g., websites)?}

\dsanswer{n/a
}

\dsquestionex{Were the individuals in question notified about the data collection?}{If so, please describe (or show with screenshots or other information) how notice was provided, and provide a link or other access point to, or otherwise reproduce, the exact language of the notification itself.}

\dsanswer{n/a
}

\dsquestionex{Did the individuals in question consent to the collection and use of their data?}{If so, please describe (or show with screenshots or other information) how consent was requested and provided, and provide a link or other access point to, or otherwise reproduce, the exact language to which the individuals consented.}

\dsanswer{n/a
}

\dsquestionex{If consent was obtained, were the consenting individuals provided with a mechanism to revoke their consent in the future or for certain uses?}{If so, please provide a description, as well as a link or other access point to the mechanism (if appropriate).}

\dsanswer{n/a
}

\dsquestionex{Has an analysis of the potential impact of the dataset and its use on data subjects (e.g., a data protection impact analysis) been conducted?}{If so, please provide a description of this analysis, including the outcomes, as well as a link or other access point to any supporting documentation.}

\dsanswer{n/a
}

\dsquestion{Any other comments?}

\dsanswer{None.
}

\bigskip
\dssectionheader{Preprocessing/cleaning/labeling}

\dsquestionex{Was any preprocessing/cleaning/labeling of the data done (e.g., discretization or bucketing, tokenization, part-of-speech tagging, SIFT feature extraction, removal of instances, processing of missing values)?}{If so, please provide a description. If not, you may skip the remainder of the questions in this section.}

\dsanswer{The raw behavioral videos are 2448x2048 pixel resolution and are sampled at 60 Hz, viewed from above an arena with 8 individual circular wells. We split these videos by cropping each well out, and blacking out the edges of the frame outside the focal circle for that well. Videos were then downsampled to 800x800 pixel resolution, and temporally downsampled to 30 Hz.
}

\dsquestionex{Was the “raw” data saved in addition to the preprocessed/cleaned/labeled data (e.g., to support unanticipated future uses)?}{If so, please provide a link or other access point to the “raw” data.}

\dsanswer{The raw two-hour movies with all wells visible are not available.
}

\dsquestionex{Is the software used to preprocess/clean/label the instances available?}{If so, please provide a link or other access point.}

\dsanswer{Labeling instances was done in BORIS (Behavioral Observation Research Interactive Software (BORIS) user guide — BORIS latest documentation).
}

\dsquestion{Any other comments?}

\dsanswer{None.
}

\bigskip
\dssectionheader{Uses}

\dsquestionex{Has the dataset been used for any tasks already?}{If so, please provide a description.}

\dsanswer{Yes: this dataset was released to accompany the 2022 Multi-Agent Behavior (MABe) Challenge, posted \href{https://www.aicrowd.com/challenges/multi-agent-behavior-challenge-2022}{here}. This competition was aimed at generating learned representations of animals' actions using unsupervised or self-supervised techniques.
}

\dsquestionex{Is there a repository that links to any or all papers or systems that use the dataset?}{If so, please provide a link or other access point.}

\dsanswer{Papers that use or cite this dataset may be submitted by their authors for display on the MABe22 website at \href{https://sites.google.com/view/computational-behavior/our-datasets/mabe2022-dataset}{https://sites.google.com/view/computational-behavior/our-datasets/mabe2022-dataset}
}

\dsquestion{What (other) tasks could the dataset be used for?}

\dsanswer{While this dataset was designed for development of methods for representation learning, the annotations can also be used for supervised learning tasks.
}

\dsquestionex{Is there anything about the composition of the dataset or the way it was collected and preprocessed/cleaned/labeled that might impact future uses?}{For example, is there anything that a future user might need to know to avoid uses that could result in unfair treatment of individuals or groups (e.g., stereotyping, quality of service issues) or other undesirable harms (e.g., financial harms, legal risks) If so, please provide a description. Is there anything a future user could do to mitigate these undesirable harms?}

\dsanswer{No.
}

\dsquestionex{Are there tasks for which the dataset should not be used?}{If so, please provide a description.}

\dsanswer{None.
}

\dsquestion{Any other comments?}

\dsanswer{None.
}

\bigskip
\dssectionheader{Distribution}

\dsquestionex{Will the dataset be distributed to third parties outside of the entity (e.g., company, institution, organization) on behalf of which the dataset was created?}{If so, please provide a description.}

\dsanswer{Yes - the full dataset will be made publicly available for download by all interested parties by July 1st, 2023.
}

\dsquestionex{How will the dataset will be distributed (e.g., tarball on website, API, GitHub)}{Does the dataset have a digital object identifier (DOI)?}

\dsanswer{The dataset is available on the Caltech public data repository at \href{https://data.caltech.edu/records/rdsa8-rde65}{https://data.caltech.edu/records/rdsa8-rde65}.
A previous version, containing only the trajectory data, is available at
\href{https://data.caltech.edu/records/20186}{https://data.caltech.edu/records/20186}. The data will be retained indefinitely and available for download by all third parties. The data.caltech.edu posting has accompanying DOI \href{https://doi.org/10.22002/rdsa8-rde65}{https://doi.org/10.22002/rdsa8-rde65}.

The dataset as used for the MABe Challenge (lacking hidden task labels) is available for download on the AIcrowd page, located at (\href{https://www.aicrowd.com/challenges/multi-agent-behavior-challenge-2022/problems/mabe-2022-mouse-triplets}{https://www.aicrowd.com/challenges/multi-agent-behavior-challenge-2022/problems/mabe-2022-mouse-triplets}). 

}

\dsquestion{When will the dataset be distributed?}

\dsanswer{The full dataset will be made publicly available for download by all interested third parties by July 1st, 2023.
}

\dsquestionex{Will the dataset be distributed under a copyright or other intellectual property (IP) license, and/or under applicable terms of use (ToU)?}{If so, please describe this license and/or ToU, and provide a link or other access point to, or otherwise reproduce, any relevant licensing terms or ToU, as well as any fees associated with these restrictions.}

\dsanswer{The MABe22 dataset is distributed under the CreativeCommons Attribution-NonCommercial-ShareAlike license (CC-BY-NC-SA). The terms of this license may be found at \href{https://creativecommons.org/licenses/by-nc-sa/2.0/legalcode}{https://creativecommons.org/licenses/by-nc-sa/2.0/legalcode}.
}

\dsquestionex{Have any third parties imposed IP-based or other restrictions on the data associated with the instances?}{If so, please describe these restrictions, and provide a link or other access point to, or otherwise reproduce, any relevant licensing terms, as well as any fees associated with these restrictions.}

\dsanswer{There are no third party restrictions on the data.
}

\dsquestionex{Do any export controls or other regulatory restrictions apply to the dataset or to individual instances?}{If so, please describe these restrictions, and provide a link or other access point to, or otherwise reproduce, any supporting documentation.}

\dsanswer{No export controls or regulatory restrictions apply.
}

\dsquestion{Any other comments?}

\dsanswer{None.
}

\bigskip
\dssectionheader{Maintenance}

\dsquestion{Who will be supporting/hosting/maintaining the dataset?}

\dsanswer{The dataset is hosted on the Caltech Research Data Repository at \href{https://data.caltech.edu/}{data.caltech.edu}. Dataset hosting is maintained by the library of the California Institute of Technology. Long-term support for users of the dataset is provided by Jennifer J. Sun and by the laboratory of Ann Kennedy.
}

\dsquestion{How can the owner/curator/manager of the dataset be contacted (e.g., email address)?}

\dsanswer{The managers of the dataset (JJS and AK) can be contacted at \href{mailto:mabe.workshop@gmail.com}{mabe.workshop@gmail.com}, or AK can be contacted at \href{mailto:ann.kennedy@northwestern.edu}{ann.kennedy@northwestern.edu} and JJS can be contacted at \href{mailto:jjsun@caltech.edu}{jjsun@caltech.edu}.
}

\dsquestionex{Is there an erratum?}{If so, please provide a link or other access point.}

\dsanswer{No.
}

\dsquestionex{Will the dataset be updated (e.g., to correct labeling errors, add new instances, delete instances)?}{If so, please describe how often, by whom, and how updates will be communicated to users (e.g., mailing list, GitHub)?}

\dsanswer{Users of the dataset have the option to subscribe to a mailing list to receive updates regarding corrections or extensions of the MABe22 dataset. Mailing list sign-up can be found on the MABe22 webpage at \href{https://sites.google.com/view/computational-behavior/our-datasets/mabe2022-dataset}{https://sites.google.com/view/computational-behavior/our-datasets/mabe2022-dataset}.

Updates to correct errors in the dataset will be made promptly, and announced via update messages posted to the MABe22 website and data.caltech.edu page.

Updates that extend the scope of the dataset, such as additional hidden tasks, or improved pose estimation, will be released as new named instantiations on at most a yearly basis. Previous versions of the dataset will remain online, but obsolescence notes will be sent out to the MABe22 mailing list. In updates, dataset version will be indicated by the year in the dataset name (here 22). Dataset updates may accompany new instantiations of the MABe Challenge.
}

\dsquestionex{If the dataset relates to people, are there applicable limits on the retention of the data associated with the instances (e.g., were individuals in question told that their data would be retained for a fixed period of time and then deleted)?}{If so, please describe these limits and explain how they will be enforced.}

\dsanswer{N/a (no human data.)
}

\dsquestionex{Will older versions of the dataset continue to be supported/hosted/maintained?}{If so, please describe how. If not, please describe how its obsolescence will be communicated to users.}

\dsanswer{Yes, the dataset will be permanently available on the Caltech Research Data Repository (data.caltech.edu), which is managed by the Caltech Library.
}

\dsquestionex{If others want to extend/augment/build on/contribute to the dataset, is there a mechanism for them to do so?}{If so, please provide a description. Will these contributions be validated/verified? If so, please describe how. If not, why not? Is there a process for communicating/distributing these contributions to other users? If so, please provide a description.}

\dsanswer{Extensions to the dataset will take place through at-most-yearly updates. We welcome community contributions of behavioral data, novel tracking methods, and novel hidden tasks; these may be submitted by contacting the authors or emailing \href{mailto:mabe.workshop@gmail.com}{mabe.workshop@gmail.com}. All community contributions will be reviewed by the managers of the dataset for quality of tracking and annotation data. Community contributions will not be accepted without a data maintenance plan (similar to this document), to ensure support for future users of the dataset.
}

\dsquestion{Any other comments?}

\dsanswer{If you enjoyed this dataset and would like to contribute other multi-agent behavioral data for future versions of the dataset or MABe Challenge, contact us at \href{mailto:mabe.workshop@gmail.com}{mabe.workshop@gmail.com}!
}

\section{Dataset Description Details}~\label{appendix:dataprep}

\subsection{Fly Groups}~\label{appendix:fly_tracking}

\subsubsection{Experimental Setup}

Optogenetic experiments used group-housed, mated female flies (4–5 days post eclosion) that were sorted into 10 flies per vial. Flies were reared in the dark in a 12:12 light-dark cycle incubator (25$^\circ$, 50\% relative humidity) on standard food supplemented with retinal (Sigma-Aldrich, St. Louis, MO) (0.2 and mM all trans-retinal prior to eclosion and 0.4 mM all trans-retinal post eclosion). Control lines, lines labeling cell types involved in the female aggression circuit, and the CsChrimson effector line were described previously \cite{schretter2020cell,aso2014mushroom}. Blind control and blind aIPg lines were generated through crossing established lines with a mutation in norpA~\cite{bloomquist1988isolation} and lines described previously~\cite{schretter2020cell}. All experiments were performed during the morning activity peak (ZT0-ZT3).

For thermogenetic experiments, flies were reared in a 12:12 light:dark incubator (22$^\circ$C 50\% relative humidity) on a standard molasses food. They were cold anesthetized and sorted into groups of 5 males and 5 females, unless noted as ``male71G01 + female control'' and ``control sex-separated''. These flies were housed separately in groups of 5 males or 5 females prior to the experiments. All flies were food deprived on agar media for 24 hours directly before recording. Experiments were conducted at the permissive temperature for TrpA, 30$^\circ$, and 50\% relative humidity during the evening activity peak (ZT8-ZT12). Control lines, the TrpA effector line, and lines labeling cell types involved in courtship or avoidance were previously described  \cite{robie2017mapping,wu2016visual}. 

The circular assay chamber was 50 mm in diameter and 3.5 mm tall, with a domed translucent ceiling coated with silicon (Sigma Cote, Sigma Aldridge) to prevent upside-down walking and a translucent acrylic floor. The chambers were illuminated from below with infrared light from custom LED panels and recorded from above with a USB3 camera at 150 fps (Flea3, FLIR) with an 800-nm long-pass filter. Visible white light was present at all times so that the flies could see. 

For optogenetic experiments, neurons expressing CsChrimson were activated with 617-nm red light from custom LED panels. Experiments were run with one of two activation protocols. Protocol 1 consisted of 2 repeats of a 30s (red) lights-off period then a 30s ``strong'' lights-on period (7 mW/cm$^2$, pulsed at 30 Hz with on period 10/33 ms), followed by a 30s lights-off period, then 2 repeats of a 30s lights-off period then a 30s ``weak'' lights-on period (3 mW/cm$^2$ constant illumination), then a 30s lights-off period. In total, these videos were 300s (45000 frames) long. Protocol 2 consisted of 3 repeats of a 30s lights-off period then a 30s ``weak'' lights-on period (1 mW/cm$^2$, constant) followed by 3 repeats of a 30s lights-off period then a 30s ``strong'' lights-on period (3 mW/cm$^2$). In total, these videos were 390s long (58500 frames). For thermogenetic experiments, videos were recorded for 300 seconds (45000 frames). 


\begin{table}
\begin{center}
\begin{tabular}{|p{.15\textwidth}|p{.1\textwidth}|p{.6\textwidth}|}\hline
Fly type & N. videos & Description \\ \hline\hline
Control 1 & 9 & Groups of 5 female and 5 male flies from control line pBDPGAL4u x TrpA that were raised together.\\\hline
Control 1 sex-separated & 4 & Groups of 5 female and 5 male flies from control line pBDPGAL4u x TrpA that were raised separately, with groups encountering each other for the first time in the videos.
\\\hline
Control 2 & 6 & Groups of 10 female flies from control line JHS\_K\_85321 x CsChrimson \\\hline
R71G01 & 13 & Groups of 5 female and 5 male flies from courtship line R71G01 x TrpA \\\hline
Male R71G01 female control & 5 & Groups of 5 female flies from the control line pBDPGAL4U x TrpA and 5 male flies from courtship line R71G01 x TrpA \\\hline
R65F12 & 12 & Groups of 5 female and 5 male flies from courtship line R65F12 x TrpA \\\hline
R91B01 & 10 & Groups of 5 female and 5 male flies from visual avoidance line R91B01 x TrpA \\\hline
Blind control & 9 & Groups of 10 blind female flies from control line JHS\_K\_85321 x ChR with the norpA mutation \\\hline
aIPg & 9 & Groups of 10 female flies from aggression line SS36564 x ChR, which targets aIPg neurons \\\hline
pC1d & 8 & Groups of 10 female flies from aggression line SS56987 x ChR, which targets pC1d neurons. \\\hline
Blind aIPg & 11 & Groups of 10 blind female flies with the norpA mutation from aggression line SS36564, which targets aIPg neurons \\\hline
Any courtship & 30 & Any of R71G01, Male R71G01 + female control, or R65F12. \\\hline
Any control & 28 & Any of Control 1, Control 1 sex-separated, Control 2, or Blind control. \\\hline
Any blind & 20 & Any of Blind control, Blind aIPg.\\\hline
Any aIPg & 20 & Any of aIPg or Blind aIPg. \\\hline
Any aggression & 28 & Any of aIPg, pC1d, blind aIPg. \\\hline
Any R71G01 & 18 & Any of R71G01 or Male R71G01 + female control \\\hline
Any sex separated & 9 & Any of Control 1 sex-separated or Male R71G01 + female control. \\\hline
\end{tabular}
\end{center}
\caption{Descriptions of types of flies used in each task. \label{table:flytypes}}
\end{table}

\begin{table}
\begin{center}
\begin{tabular}{|p{.2\textwidth}|p{.7\textwidth}|}\hline
Task type & Description \\ \hline\hline
Fly type & 1 indicates activation periods (whole video for TrpA, any lights-on periods for ChR) of the selected fly type. 0 indicates activation periods for other lines. nan indicates lights-off periods. \\\hline
On vs off & 1 indicates activation lights-on periods for the selected fly type, 0 lights-off periods for that fly type. nan indicates other fly types. \\\hline
Strong vs off & 1 indicates strong activation lights-on periods for the selected fly type, 0 lights-off periods for that fly type. nan indicates other fly types.\\\hline
Weak vs off & 1 indicates weak activation lights-on periods for the selected fly type, 0 lights-off periods for that fly type. nan indicates other fly types.\\\hline
Strong vs weak & 1 indicates strong activation lights-on periods for the selected fly type, 0 weak activation lights-on periods for that fly type. nan indicates lights-off periods for that fly type, or any other fly type.\\\hline
Last vs first & 1 indicates the last strong activation lights-on period for the selected fly ty[e, 0 the first strong activation lights-on period for that fly type. nan indicates other lights-on periods or lights off periods for that fly type, or any other fly type.\\\hline
Manual annotation & 1 indicates frames from any fly type manually labeled as the selected behavior, 0 frames manually labeled as not the selected behavior, nan frames that were not labeled.  \\\hline
Female vs male & 1 indicate female flies, 0 indicates male flies. \\\hline
\end{tabular}
\end{center}
\caption{Descriptions of types of comparisons made in each task. \label{table:tasktypes}}
\end{table}

\begin{table}
\begin{center}
\begin{tabular}{|c|p{.15\textwidth}|c|}\hline
Task & Flies/Behavior & Task type\\ \hline\hline
1 & Control 1 & Fly type\\\hline
2 & Control 1 sex-separated & Fly type \\\hline
3 & Control 2 & Fly type \\\hline
4 & R71G01 & Fly type \\\hline
5 & male R71G01 female control & Fly type \\\hline
6 & R65F12 & Fly type \\\hline
7 & R91B01 & Fly type \\\hline
8 & Blind Control & Fly type \\\hline
9 & aIPG & Fly type \\\hline
10 & pC1d & Fly type \\\hline
11 & Blind aIPG & Fly type \\\hline
12 & Blind control & On vs off\\\hline
13 & Blind control & Strong vs off\\\hline
14 & Blind control & Weak vs off \\\hline
15 & Blind control & Strong vs weak \\\hline
16 & Blind control & Last vs first \\\hline
17 & Control 2 & On vs off\\\hline
18 & Control 2 & Strong vs off\\\hline
19 & Control 2 & Weak vs off \\\hline
20 & Control 2 & Strong vs weak \\\hline
21 & Control 2 & Last vs first \\\hline
22 & Blind aIPg & On vs off\\\hline
23 & Blind aIPg & Strong vs off\\\hline
24 & Blind aIPg & Weak vs off \\\hline
25 & Blind aIPg & Strong vs weak \\\hline
\end{tabular}
\hspace{.1cm}
\begin{tabular}{|c|p{.15\textwidth}|c|}\hline
Task & Flies/Behavior & Task type\\ \hline\hline
26 & Blind aIPg & Last vs first \\\hline
27 & aIPg & On vs off\\\hline
28 & aIPg & Strong vs off\\\hline
29 & aIPg & Weak vs off \\\hline
30 & aIPg & Strong vs weak \\\hline
31 & aIPg & Last vs first \\\hline
32 & pC1d & On vs off\\\hline
33 & pC1d & Strong vs off\\\hline
34 & pC1d & Weak vs off \\\hline
35 & pC1d & Strong vs weak \\\hline
36 & pC1d & Last vs first \\\hline
37 & Any courtship & Fly type \\\hline
38 & Any control & Fly type \\\hline
39 & Any blind & Fly type \\\hline
40 & Any aIPg & Fly type \\\hline
41 & Any aggression & Fly type \\\hline
42 & Any R71G01 & Fly type \\\hline
43 & Any sex-separated & Fly type \\\hline
44 & All & Female vs male\\\hline
45 & Aggression & Manual annotation \\\hline
46 & Chase & Manual annotation \\\hline
47 & Courtship & Manual annotation \\\hline
48 & High fence & Manual annotation \\\hline
49 & Wing ext. & Manual annotation \\\hline
50 & Wing flick & Manual annotation \\\hline
\end{tabular}
\end{center}
\caption{Descriptions of fly tasks. \label{table:flytasks}}
\end{table}

\subsubsection{Fly Tracking}

\begin{figure}[h]
    \centering
    \includegraphics[width=\linewidth]{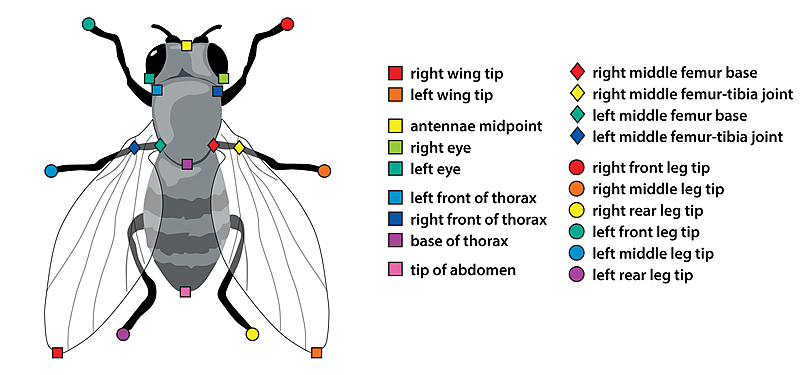}
    \caption{19 tracked landmark points on the fly body.
    }
    \label{fig:fly_landmarks}
\end{figure}

The body and wings of the flies were tracked using the FlyTracker software~\cite{eyjolfsdottir2014detecting}. 19 selected landmark points were tracked using the Animal Part Tracker (APT)~\cite{APT}, depicted in Figure~\ref{fig:fly_landmarks}. Coordinates were converted from pixels to millimeters by detecting the circular arena boundary, with $(0,0)$ corresponding to the arena center. 

\subsubsection{Fly Behavior Annotation}

Using JAABA~\cite{kabra2013jaaba}, we annotated 6 behaviors involved in fly courtship and aggression:
\begin{itemize}
\item \textbf{Aggression}: The focus fly was angled towards another fly and engaged in several touches with $\geq 2$ limbs to the head, abdomen or thorax of another fly, causing the other fly to move. This behavior included head butting, fencing, and shoving behaviors as defined~\cite{nilsen2004gender,schretter2020cell}.
\item \textbf{Chase}: The focus fly was following another moving fly, maintaining a small, somewhat constant distance to it~\cite{robie2017mapping}.
\item \textbf{Courtship}:  The focus fly was performing any stage of the courtship sequence, including orienting, following, tapping, singing, licking, attempted copulation, or copulation~\cite{sokolowski2001drosophila}.
\item \textbf{High posture fencing}: The focus fly was angled towards another fly with the mid legs of the fly angled sharply ($< 45$ degrees), and the forelegs lifted off of the bottom of the arena and touching limbs, head, abdomen or thorax of another fly~\cite{nilsen2004gender,schretter2020cell}.
\item \textbf{Wing extension}: The focus fly unilaterally rotates a wing out for an extended period of time. This behavior is likely an indication of the fly producing courtship song with the extended wing~\cite{robie2017mapping}.
\item \textbf{Wing flick}: The focus fly rapidly and symmetrically moves its wings out and back in performing a quick scissoring movement several times in a row~\cite{robie2017mapping}. 
\end{itemize}
As all of the behaviors we annotated occur rarely, we sparsely annotated the data using frames suggested using JAABA's interactive system. We only annotated frames for which we were confident of the correct class. We annotated frames across all fly types, for many different videos and flies. For all behaviors, the classifiers trained by JAABA using the annotated data looked reasonable, based on casual proofreading.

\subsubsection{Data splitting}

We split the data into 4 sets, with each set containing distinct videos and flies.
\begin{itemize}
\item User train: Data given to the competitor to learn their embedding. 
\item Evaluation train: Data used to train the linear classifier during evaluation.
\item Test 1: Data used to measure performance of the linear classifier. Performance on this dataset was presented on the leaderboard during the competition. 
\item Test 2: Final set of data used to measure performance on the linear classifier, used for determining the competition winners. 
\end{itemize}
We used simulated annealing to find a way to split the videos so that: 
\begin{itemize}
\item There were videos from each fly type in each set. 
\item There were manual labels from each fly type and each behavior category in each set. 
\item Approximately 60\% of videos were in User train, 20\% in Evaluator train, 10\% in Test 1, and 10\% in Test 2.
\item For each behavior type and fly type, approximately 40\% of manual labels for each behavior were in User train, 30\% in Test 1, and 30\% in Test 2.
\end{itemize}
We split each video into segments of length 30s (4500 frames), with gaps of a randomly selected interval between .5s (75 frames) and 2s (150 frames) between segments. Included segments were chosen such that they did not include obvious identity tracking errors (trajectory births or deaths). Flies were shuffled within each segment so that fly $i$ across segments did not correspond.

\subsection{Mice Triplets}~\label{appendix:mouse_task}

\subsubsection{Experimental Setup}

This section is adapted from \cite{beane2022video,sheppard2020gait,geuther2019robust}. Experiments were performed in the JAX Animal Behavior System (JABS), consisting of an open field arena measuring 52 cm by 52 cm, with overhead LED ring lighting on a 12:12 light-dark cycle. The arena floor is white PVC plastic covered by a layer of bedding (wood shavings and Alpha-Dri), and food and water are held in a hopper with grate access in one arena wall, and replaced when depleted. For recording videos while lights were off, additional IR LED lighting at 940 nm was added. Video was recorded at 30Hz using a Basler acA1300-75gm camera with 4-12mm lens (Tamron) and 800nm longpass filter (Hoya) to exclude visible light, using a custom recording client developed by JAX (see \href{https://github.com/KumarLabJax/JABS-data-pipeline}{https://github.com/KumarLabJax/JABS-data-pipeline}). Experimental mice were adult males between 10 and 20 weeks old, of genetic background C57Bl/6J or BTBR. Prior to testing, animals were allowed to acclimate to the behavior room for 30-60 minutes, after which three mice were introduced to the JABS arena over a period of several minutes. Behavior was recorded continuously for four days, during which time animal behavior and welfare was monitored remotely. All behavioral tests were performed in accordance with approved protocols from The Jackson Laboratory Institutional Animal Care and Use Committee guidelines.

\subsubsection{Mouse Tracking}

12 anatomical keypoints on each animal were tracked using a modified version of HRnet (provided at \href{https://github.com/KumarLabJax/deep-hrnet-mouse}{https://github.com/KumarLabJax/deep-hrnet-mouse}), with coordinates of keypoints reported in pixels \cite{sheppard2020gait}. Occurrence of each anatomically defined keypoint were grouped into up to four animal pose instances (one more than the number of mice present), using associative embedding~\cite{newell2017associative} to evaluate likelihood of keypoint pairs belonging to the same animal. The four candidate pose instances were then assigned animal identities by computing distances between all tracked pose pairs across neighboring video frames, and propagating animal IDs forward in time to the closest pose instance falling within a maximum radius. A second post-hoc pass was then applied to extracted pose tracklets, in which incomplete pose instances were merged when complementary pairs of points were found within a maximum radius, and resulting tracklets were merged based on a minimum distance criterion, to produce the final set of three pose trajectories provided in the dataset.

\begin{table}[H]
\centering
\begin{tabular}{ |c|c|c|p{8cm}| } 
 \hline
 Task Name & Type & Values & Description \\ 
 \hline
 \hline
 Experiment day & Sequence & 1-4 & Mice were filmed interacting for four days after introduction to a new arena; task is to determine which day a sequence comes from. \\ 
 \hline
 Time of day & Sequence & 0-1440 & Mice show circadian changes in their level of activity; task is to infer time of day from behavior. \\
 \hline
 Strain & Sequence & 0 or 1 & Mice are from either C57Bl/6J or BTBR genetic background. Strain field is 1 for BTBR and 0 for C57Bl/6J. \\
 \hline
 Lights & Sequence & 0 or 1 & Mice are more active when the lights are off, which occurs between 6am and 6pm; task is to infer light condition from behavior.\\
 \hline
\end{tabular}
\caption{Format of experimentally-defined tasks for mouse dataset.}\label{tab:mouse_tasks}
\end{table}

\subsubsection{Mouse Behavior Annotation}
Mouse behavioral videos were manually annotated using the VIA video annotator \cite{dutta2019via}. Each of the behaviors: huddle, chase, anal sniff, and face sniff, was annotated as an individual time series with frame-level temporal resolution. We annotated 400 clips overall (200/100/100; train/val/test), randomly selected from the full set of videos. Chase was annotated when a pair of mice moved quickly, with one mouse following close behind the other. Huddle was annotated when the bodies of the mice are in close contact and the animals are stationary for at least several seconds; it can occur between either pairs or triplets of animals. Face sniffing was annotated when a close-investigation behavior occurred in which the nose of one mouse was in close contact with the nose or face of another mouse. Anogenital sniffing was annotated for a close-investigation behavior in which one mouse is investigating the anogenital area of another, typically with its nose near the base of the tail or pushed underneath the hindquarters of the other animal.

\subsubsection{Data splitting}
Each dataset was randomly assigned into four sets; due to the relatively small number of source experiments, we did not separate sets by animal identity. The percentage of videos/trajectories assigned to each set is given in parentheses.
\begin{itemize}
    \item User train (30\%): Data given to the competitor to learn their embedding (note that competitors could also include the submission train, test 1, and test 2 video/trajectories for training, but these were not included for experiments in the main text.)
    \item Evaluation train (50\%): Data used to train the linear classifiers during evaluation.
    \item Test 1 (10\%): Data used to measure performance of the linear classifiers. Performance on this dataset was presented on the leaderboard during the competition.
    \item Test 2 10\%): Final set of data used to measure performance of the linear classifiers, and for determining the competition winners.
\end{itemize}

\subsection{Beetle Interactions}~\label{appendix:beetle_interactors}

\subsubsection{Experimental Setup}

This dataset consists of videos of paired insect interactions. One of the interactor is a symbiotic rove beetles (\textit{Sceptobius lativentris}), while the other interactor may be their host ant (\textit{Liometopum occidentale}), manipulated host ant (e.g. with pheromones stripped off), or other insects (e.g. clown or nitidulid beetles). The original video recordings consists of 8-well behavioral interaction chambers (2cm diameter circles) in the dark and illuminated with infrared lights from the side/top. A top-mounted machine vision camera sensitive to IR light monitored the two-hour behavioral trials at 60 Hz, which we downsample to 30Hz for MABe22. Individual circular wells were cropped/parsed from the multi-well video by hand and saved at 800x800 resolution.
We annotated six behaviors in whole two-hour videos, consisting of seven different types of one-on-one interactions using BORIS (14 hours total). These interactors represent a range of cue types, from the host organism with which the symbiont should interact extensively, to a neutral random other insect which the symbiont will likely ignore. Generating a meaningful representation that extracts information of interest about the different behaviors adopted by the beetle in response to these disparate cues is crucial for insight into how species interact in nature.

\subsubsection{Beetle Task Descriptions}

For the beetle dataset, identifying the sequence-level interactor apply to all frames, while the frame-level behavior tasks apply to a subset of the videos. All of these tasks are classification, except for a regression task for interaction duration, where the goal is to identify how long the two organisms have been interacting (up to 4 hours).

The following sequence-level labels describe the type of interactors present:

\begin{tabular}{|p{3cm}|p{10cm}|}
\hline
histerid & \textit{Sceptobius lativentris} interacting with a clown beetle (family \textit{Histeridae}.) \\
\hline
nitidulid & \textit{Sceptobius lativentris} interacting with a sap beetle (family \textit{Nitidulidae}.) \\
\hline
locc & \textit{Sceptobius lativentris} interacting with a live \textit{Liometopum occidentale}. \\
\hline
gasterless & \textit{Sceptobius lativentris} interacting with a live gasterless \textit{Liometopum occidentale} ant, i.e. an ant with its gaster (abomen) removed. \\
\hline
platy & \textit{Sceptobius lativentris} interacting with a live \textit{Platyusa sonomae} beetle. \\
\hline
reapplied & \textit{Sceptobius lativentris} interacting with a dead \textit{Liometopum occidentale} stripped of pheromones and then with pheromones reapplied. \\
\hline
tethered & \textit{Sceptobius lativentris} interacting with a live \textit{Liometopum occidentale} tethered to a magnet, i.e. immobilized in the center of the arena. \\
\hline
\end{tabular}

The following frame-wise labels reflect categories of behavior present in the video:

\begin{tabular}{|p{5cm}|p{10cm}|}
\hline
grooming object & \textit{Sceptobius lativentris} is grooming the interactor object/insect. \\
\hline
grooming self & \textit{Sceptobius lativentris} is grooming itself (e.g. cleaning an antenna). \\
\hline
idle alone & \textit{Sceptobius lativentris} is idle (not doing any visible behavior) by itself. \\
\hline
idle object & \textit{Sceptobius lativentris} is idle (not doing any visible behavior) by on top of the interactor/object. \\
\hline
exploring object & \textit{Sceptobius lativentris} is exploring (moving around on) atop the interactor/object. \\
\hline
exploring alone & \textit{Sceptobius lativentris} is exploring (moving around) in the arena. \\
\hline
\end{tabular}
\label{tab:beetle_behaviors}

To evaluate the performance of frame-level behavior labels, we generate two sets of evaluation conditions, same and different. For same, we create the evaluation train and test splits with the same interactor types (so the linear evaluator has access to the same interactors for behavior classification during train and test). For different, we create the evaluation train and test split with different interactor types (so the linear evaluator has access to different interactors for behavior classification during train and test). Note that this only affects the linear evaluation split, and does not affect the representation learning model.

\subsubsection{Data splitting}
Each dataset was randomly assigned into four sets; the data is split such that either the interactor type is the same across evaluation splits, or different as described above. The percentage of videos/trajectories assigned to each set is given in parentheses. Note that this percentage may vary across different conditions.
\begin{itemize}
    \item User train (25\%): Data given to the competitor to learn their embedding (note that competitors could also include the submission train, test 1, and test 2 video/trajectories for training, but these were not included for experiments in the main text.)
    \item Evaluation train (60\%): Data used to train the linear classifiers during evaluation.
    \item Test 1 (7.5\%): Data used to measure performance of the linear classifiers. Performance on this dataset was presented on the leaderboard during the competition.
    \item Test 2 7.5\%): Final set of data used to measure performance of the linear classifiers, and for determining the competition winners.
\end{itemize}

\section{Evaluation}~\label{sec:evaluation}

For all tasks, we evaluate representation learning performance using a linear evaluation protocol, by training a linear model on top of the learned representation at each frame for classification and regression on a set of downstream tasks. These downstream tasks are unseen during training of the representation learning model. We train separate linear models per task, and because of the high class imbalance of some tasks, the classes are weighted inverse to class frequencies during training. 

For training the linear models, we use three fixed random $80\%$ of the evaluation train split to train three models. All evaluations are performed on a fixed test set. For classification tasks, majority voting combines the predictions of the three classifiers. For regression tasks, the predictions are averaged. Both merging schemes are done at the frame level. The evaluation metrics are F1 score for classification and Mean Squared Error for regression computed for each sequence, then averaged over the sequences. Note that all sequences given an organism have the same number of frames. We use default hyperparameters for the Ridge classifier and do not perform hyperparameter tuning. Notably, the evaluation framework does not choose a particular feature normalization strategy, and any feature normalization should happen before input to the framework.

\paragraph{F1 score.} The F1 score is the harmonic mean of the Precision $P$ and Recall $R$:
\begin{align}
P = \frac{TP}{TP + FP} \\
R = \frac{TP}{TP + FN} \\
F1 = \frac{2 \times P \times R}{P+R}
\end{align}
Where true positives (TP) is the number of frames that a model correctly labels as positive for a class, false positives (FP) is the number of frames incorrectly labeled as positive for a class, and false negatives (FN) is the number of frames incorrectly labeled as negative for a class.

For F1 score across tasks, we take an unweighted average across classification tasks in either the mouse or fly domain. For our evaluation, the class with the highest predicted probability in each frame was used to compute F1 score, but the F1 score will likely be higher with threshold tuning.

\paragraph{Mean Squared Error.} For regression tasks, given $n$ data samples, we use the predicted values $\bar{y}$ and the real labels $y$ to compute:
\begin{align}
MSE =  \frac{1}{n}\sum_{i=1}^n (y_i - \bar{y}_i)^2
\end{align}

We normalize the label values for regression to between 0 and 1. In our dataset, the experiment day and time of day tasks are regression tasks, while all other tasks are classification tasks.

\section{Implementation Details/Hyperparameters}~\label{appendix:implementation_details}

For studying self-supervised video learning we used adapted the SlowFast~\cite{fan2020pyslowfast} implementations of SOTA methods. We list hyperparameters for each methods below.

\begin{table}[H]
    \centering
    \begin{tabular}{l|c} 
         Config & Value \\ 
         \hline
         optimizer & AdamW~\cite{loshchilov2017decoupled} \\
         optimizer momentum & $\beta_1,\beta_2$=0.9,0.95~\cite{chen2020generative} \\
         weight decay & 0.05 \\
         learning rate & 1.6e-4\\
         learning rate schedule & cosine decay~\cite{loshchilovstochastic}\\
         warmup epochs~\cite{goyal2017accurate} & 60\\
         epochs & 2000\\
         augmentation & hflip, crop [0.5, 1]\\
         batch size & 64\\
         gradient clipping & 0.02\\
    \end{tabular}
    \caption{Training parameters for MAE~\cite{he2022masked}.}
\end{table}

\begin{table}
    \centering
    \begin{tabular}{l|c} 
         Config & Value \\ 
         \hline
         optimizer & AdamW~\cite{loshchilov2017decoupled} \\
         optimizer momentum & $\beta_1,\beta_2$=0.9,0.999 \\
         weight decay & 0.05 \\
         learning rate & 0.0001\\
         learning rate schedule & cosine decay~\cite{loshchilovstochastic}\\
         warmup epochs~\cite{goyal2017accurate} & 10\\
         epochs & 800\\
         augmentation & hflip, crop [0.5, 1]\\
         batch size & 32\\
         gradient clipping & 0.02\\
    \end{tabular}
    \caption{Training parameters for MaskFeat~\cite{wei2022masked}.}
\end{table}

\begin{table}
    \centering
    \begin{tabular}{l|c} 
         Config & Value \\ 
         \hline
         optimizer & SGD \\
         optimizer momentum & $\beta_1,\beta_2$=0.9,0.999 \\
         weight decay & 1e-6 \\
         learning rate & 1.2\\
         learning rate schedule & cosine decay~\cite{loshchilovstochastic}\\
         warmup epochs~\cite{goyal2017accurate} & 35\\
         epochs & 200\\
         augmentation & hflip, crop [0.5, 1]\\
         batch size & 32\\
         gradient clipping & 0.02\\
    \end{tabular}
    \caption{Training parameters for $\rho$BYOL~\cite{feichtenhofer2021large}.}
\end{table}

\newpage

\section{Additional Trajectory Method Results}~\label{sec:addtrajectory}

We present additional results for trajectory based methods, from community-contributed solutions for the first phase of our challenge. This dataset consists of 5336 clips of mouse triplets, alongside 968 clips of fly data.

\subsection{Mouse Programmatically-Annotated Behaviors}

In addition to the experimental condition labels outlined above, the 9 behaviors were programmatically annotated using heuristics described below using the trajectory data. These programmatically-annotated behaviors were used to evaluate the mouse trajectory methods. Note that multiple behavior labels may be positive on a given frame.

\begin{itemize}
\item \textbf{Approach}: Mice move from at least 5 cm apart to less than 1 cm apart at closest point, over a period of at least 10 seconds at a maximum speed of 2 cm/sec.
 \item \textbf{Chase}:  Mice are moving above 15 cm/sec, with closest points less than 5 cm apart, and angular deviation between mice is less than 30 degrees, for at least 80\% of frames within at least one second. Merge bouts less than 0.5 seconds apart.
 \item \textbf{Close}: Closest points of mice are less than 3 cm apart. Merge bouts less than 2 seconds apart.
 \item \textbf{Contact}: Closest points of mice are less than 1 cm apart. Merge bouts less than 2 seconds apart.
 \item \textbf{Huddle}: Closest points of mice are less than 1 cm apart for at least 10 seconds, during which mice show less than 3 cm displacement. Merge bouts less than 2 seconds apart.
 \item \textbf{Oral-ear contact}:  Nose and ear of mice are less than 1.5 cm apart for at least 50\% of frames within a window of 0.25 seconds or more. Must occur less than 5 seconds after an approach. Merge bouts less than 0.5 seconds apart.
 \item \textbf{Oral-genital contact}: Nose and tail base of mice are less than 1.5 cm apart for at least 50\% of frames within a window of 0.25 seconds or more. Must occur less than 5 seconds after an approach.  Merge bouts less than 0.5 seconds apart.
 \item \textbf{Oral-oral contact}: Noses of mice are less than 1.5 cm apart for at least 50\% of frames within a window of 0.25 seconds or more. Must occur less than 5 seconds after an approach. Merge bouts less than 0.5 seconds apart.
 \item \textbf{Watching}: Mice are more than 5 cm apart but less than 20 cm apart, and gaze offset of one mouse is less than 15 degrees from body of other mouse, for a minimum duration of 3 seconds. Merge bouts less than 0.5 seconds apart.
\end{itemize}

\subsection{Results}

\begin{table}
\begin{center}
 \scalebox{0.9}{
    \begin{tabular}{l|ccccccc}
        \toprule[0.2em]
        \multirow{2}{*}{Mice Triplet} &
         Exp. & Time of & Strain & Movement  & Contact  & Watching  & Lights\\
         & Day $\downarrow$ & Day $\downarrow$ & ↑ &  Group↑ &  Group↑ & ↑ & ↑\\        
        \toprule[0.2em]
        PCA & $.0942 \pm .0000$ & $.946 \pm .0000$ & $.516 \pm .002$ & $.005 \pm .000$ & $.169 \pm .001$ & $.066 \pm .001$ & $.546 \pm .002$ \\
        TVAE & $.0940 \pm .0002$ & $.944 \pm .0001$ & $.530 \pm .001$ & $.008 \pm .000$ & $.213 \pm .001$ & $.102 \pm .002$ & $.568 \pm .005$\\
        T-Perceiver & $.0933 \pm .0005$ & $.932 \pm .0005$ & $.698 \pm .014$ & $.014 \pm .001$ & $.232 \pm .005$ & $.164 \pm .005$ & $\bm{.697 \pm .006}$\\    
        T-GPT & $.0927 \pm .0004$ & $.938 \pm .0001$ & $.645 \pm .004$ & $.012 \pm .000$ & $.252 \pm .003$ & $\bm{.179 \pm .005}$ & $.654 \pm .004$\\ 
        T-PointNet & $.0928 \pm .0001$ & $.932 \pm .0001$ & $.660 \pm .004$ & $\bm{.036 \pm .003}$ & $.256 \pm .001$ & $.156 \pm .005$ & $.672 \pm .000$\\
        T-BERT & $\bm{.0926 \pm .0004}$ & $\bm{.928 \pm .0003}$ & $\bm{.786 \pm .022}$ & $.013 \pm .000$ & $\bm{.266 \pm .003}$ & $.172 \pm .006$ & $.688 \pm .003$ \\
        \bottomrule[0.1em]
        \multirow{2}{*}{Fly Group} &
         Fly & Stimulation, & Stimulation, & Line  & Female  & Manual  & -\\
         & Type ↑ & Control ↑& Aggression ↑&  Category ↑&  vs. Male ↑ & Behaviors ↑& \\        
        \toprule[0.2em]
        PCA & $.282 \pm .017$ & $.466 \pm .002$ & $.484 \pm .001$ & $.553 \pm .006$ & $\bm{.990 \pm .000}$ & $.230 \pm .002$ & - \\
        TVAE & $.199 \pm .005$ & $.500 \pm .019$ & $.450 \pm .011$ & $.341 \pm .009$ & $.821 \pm .005$ & $.222 \pm .011$ & - \\
        T-Perceiver & $\bm{.394 \pm .018}$ & $.418 \pm .039$ & $\bm{.513 \pm .013}$ & $\bm{.573 \pm .013}$ & $.982 \pm .002$ & $.197 \pm .018$ & - \\    
        T-GPT & $.363 \pm .015$ & $\bm{.515 \pm .020}$ & $.500 \pm .009$ & $.557 \pm .019$ & $.873 \pm .001$ & $\bm{.246 \pm .014}$ & - \\ 
     \bottomrule[0.1em]
    \end{tabular}
    }
  \end{center}
  \caption{\textbf{MABe2022 Trajectory Benchmark Results}. Task-averaged MSE and F1 score are from mean and standard deviation over five runs. For mouse task groups, ``Movement" consists of approach and chase behaviors, and ``Contact" consists of close, contact, huddle, oral-ear contact, oral-genital contact, and oral-oral contact behaviors. For fly task groups, ``Fly type" corresponds to tasks 1 to 11,  ``Stimulation Control" is tasks 12 to 21, ``Stimulation Aggression" is tasks 22 to 36, ``Line Category" is tasks 37 to 43, and ``Manual Behaviors" is tasks 45 to 50 in Appendix Table~\ref{table:flytasks}. The best performing model is in bold.}
  \label{tab:res_summary}
\end{table}

First, we perform a frame-wise PCA as a simple baseline. Principal components were computed from the centered and normalized pose of each mouse, or from the centered pose of each fly and its two nearest neighbors, giving a $60$-dim representation for mouse and $253$-dim representation for fly.

Taking into account all task groups across both datasets, the current best performing models are generally based on transformer architectures (Table~\ref{tab:res_summary}). 
Interestingly, T-PointNet, which models trajectory features using point clouds, is competitive on the mouse triplet data. Further work to extend this model to account for more agents could improve its fly group performance. For many mouse and fly task groups, PCA performance was very close to the Base model. However, the top performing models show a significant improvement in performance, demonstrating that we can learn representations that improve behavior analysis performance, even without knowledge of the downstream evaluation tasks. 

In general, task categories consisting of annotated behaviors are the most challenging for existing models, likely due to the relatively rare positive behavior annotations. 
These task labels are at the frame-level, where there is a need to capture local temporal information, compared to sequence-level tasks such as ``Strain" and ``Fly Type" which does not vary over a clip. 
Representations that can further improve data efficiency of downstream classifiers or better capture local temporal information could help improve the performance of these task groups.

\end{document}